\documentclass{article}
\usepackage{amsmath,amsfonts,bm}

\def\eqref#1{equation~\ref{#1}}
\def\1{\bm{1}}

\def\va{{\bm{a}}}

\def\vd{{\bm{d}}}

\def\vh{{\bm{h}}}

\def\vx{{\bm{x}}}

\def\vz{{\bm{z}}}

\def\mA{{\bm{A}}}

\def\mD{{\bm{D}}}

\def\mW{{\bm{W}}}

\DeclareMathAlphabet{\mathsfit}{\encodingdefault}{\sfdefault}{m}{sl}
\SetMathAlphabet{\mathsfit}{bold}{\encodingdefault}{\sfdefault}{bx}{n}

\usepackage{amsmath,amsfonts,bm}
\usepackage{caption}
\usepackage{subcaption}
\usepackage{verbatim}
\usepackage[dvipsnames]{xcolor}
\definecolor{orange}{rgb}{0.93725,0.52549,0.2117647}
\definecolor{blue}{rgb}{0.23137,0.4588,0.68627}

\usepackage{dirtytalk}
\usepackage{wrapfig}
\usepackage{color, colortbl}
\usepackage{microtype}
\usepackage{stfloats}
\usepackage{graphicx}
\usepackage{booktabs} % for professional tables
\usepackage{multirow} % tables with cell spanning multiple rows
\usepackage{siunitx}
\usepackage{array}
\usepackage{amssymb}
\usepackage{caption}
\usepackage{subcaption}
\newcommand{\thickhline}{%
    \noalign {\ifnum 0=`}\fi \hrule height 1pt
    \futurelet \reserved@a \@xhline
}
\newcolumntype{"}{@{\hskip\tabcolsep\vrule width 1pt\hskip\tabcolsep}}

\usepackage{upquote,textcomp,amsmath}

\usepackage[pagebackref=false,breaklinks=true,letterpaper=true,colorlinks,bookmarks=false]{hyperref}
\hypersetup{colorlinks,breaklinks,
            urlcolor=[rgb]{0.93725,0.52549,0.2117647},
            citecolor=[rgb]{0.23137,0.4588,0.68627},
            linkcolor=[rgb]{0.93725,0.52549,0.2117647}}

\usepackage{amsthm}
\newtheorem{theorem}{Theorem}%[section]

\newtheorem{definition}{Definition}%[section]
\renewcommand{\eqref}[1]{Eq.~\ref{#1}}

\usepackage[numbers]{natbib}
\usepackage[nonatbib,final]{neurips_2024}

\usepackage[utf8]{inputenc} % allow utf-8 input
\usepackage[T1]{fontenc}    % use 8-bit T1 fonts
\usepackage{hyperref}       % hyperlinks
\usepackage{url}            % simple URL typesetting
\usepackage{booktabs}       % professional-quality tables
\usepackage{amsfonts}       % blackboard math symbols
\usepackage{nicefrac}       % compact symbols for 1/2, etc.
\usepackage{microtype}      % microtypography
\usepackage{xcolor}         % colors

\title{Almost-Linear RNNs Yield Highly Interpretable Symbolic Codes in Dynamical Systems Reconstruction}

\author{%
    Manuel Brenner\textsuperscript{1,2}\footnotemark[1] , 
    Christoph Jürgen Hemmer\textsuperscript{1,3}\footnotemark[1] ,
    Zahra Monfared\textsuperscript{2}, 
    Daniel Durstewitz\textsuperscript{1,2,3}\\
    \textsuperscript{1}Dept. of Theoretical Neuroscience, Central Institute of Mental Health, Medical Faculty, \\
    Heidelberg University, Germany \\
    \textsuperscript{2}Interdisciplinary Center for Scientific Computing (IWR), Heidelberg University, Germany \\
    \textsuperscript{3}Faculty of Physics and Astronomy, Heidelberg University, Heidelberg, Germany \\
}

\begin{document}
\renewcommand{\thefootnote}{\fnsymbol{footnote}}
\footnotetext[1]{These authors contributed equally to this work. \\ Corresponding authors: \{manuel.brenner, christoph.hemmer, daniel.durstewitz\}@zi-mannheim.de}
\renewcommand{\thefootnote}{\arabic{footnote}}

\maketitle

\begin{abstract}
Dynamical systems (DS) theory is fundamental for many areas of science and engineering. It can provide deep insights into the behavior of systems evolving in time, as typically described by differential or recursive equations. A common approach to facilitate mathematical tractability and interpretability of DS models involves decomposing nonlinear DS into multiple linear DS separated by switching manifolds, i.e. piecewise linear (PWL) systems. PWL models are popular in engineering and a frequent choice in mathematics for analyzing the topological properties of DS. However, hand-crafting such models is tedious and only possible for very low-dimensional scenarios, while inferring them from data usually gives rise to unnecessarily complex representations with very many linear subregions. Here we introduce Almost-Linear Recurrent Neural Networks (AL-RNNs) which automatically and robustly produce most parsimonious PWL representations of DS from time series data, using as few PWL nonlinearities as possible. AL-RNNs can be efficiently trained with any SOTA algorithm for dynamical systems reconstruction (DSR), and naturally give rise to a symbolic encoding of the underlying DS that provably preserves important topological properties. We show that for the Lorenz and Rössler systems, AL-RNNs discover, in a purely data-driven way, the known topologically minimal PWL representations of the corresponding chaotic attractors. We further illustrate on two challenging empirical datasets that interpretable symbolic encodings of the dynamics can be achieved, tremendously facilitating mathematical and computational analysis of the underlying systems.

\end{abstract}

\section{Introduction} \label{sec:method:hierachisation}

Dynamical systems (DS) %underpin
underlie many real-world phenomena of scientific and practical relevance. %importance. 
Complex chaotic %and multi-fractal 
DS are believed to govern market dynamics \cite{mandelbrot_misbehavior_2007}, 
the rhythms of the brain \cite{buzsaki_rhythms_2006}, climate systems \cite{tziperman-97}, or ecosystems \cite{ecology1}. A by now rapidly growing field in scientific ML is dynamical systems reconstruction (DSR), where the goal is to learn a DS model directly from data %from observations 
that constitutes a generative surrogate model of the data-generating DS. DSR increasingly relies on %ML methods for the extraction of models from data
deep learning, especially in contexts where dynamics are too complex to be captured by simple equations or where the underlying processes %principles 
are not fully understood.

%Several mathematically tractable 
One way of making DSR models mathematically accessible is piecewise linear (PWL) designs, popular among engineers for decades \cite{bemporad2000piecewise,carmona2002simplifying, juloski_bayesian_2005,  rantzer2000piecewise, sontag_nonlinear_1981}. In the mathematical theory of DS, PWL models also play a special role and simplify many types of analysis \cite{alligood_chaos_1996, lind_introduction_1995}, such as the characterization of bifurcations \citep{feigin_increasingly_1995, bernardo_local_1999, jain_border-collision_2003,
hogan_dynamics_2007, patra_multiple_2018, monfared_existence_2020}. This is because linear DS are well-understood and straightforward to analyze, while nonlinear DS lack an equally simple description \citep{perko_differential_2001, strogatz_nonlinear_2015, brunton_data-driven_2019}. RNNs based on PWL activation functions, like rectified-linear units (ReLUs), have been proposed recently for learning mathematically tractable DSR models. Such piecewise-linear RNNs (PLRNNs), combined with effective training techniques for controlling gradient flows
\cite{mikhaeil_difficulty_2022,hess_generalized_2023}, achieve state-of-the-art (SOTA) performance across a wide range of DSR tasks, including challenging (high-dimensional, noisy, chaotic, partially observed ...) empirical time series \cite{durstewitz_state_2017, brenner_tractable_2022, hess_generalized_2023, brenner_integrating_2024}.
However, while featuring a PWL design, the resulting constructions are often complex, with a large number of linear subregions required to capture the data properly, hence still impeding effective analysis. 
On the other hand, a class of switching linear DS has been proposed to decompose nonlinear DS into linear regions combined with switching states that determine the transitions between these regions \cite{ackerson_state_1970, ghahramani_variational_2000, fox_nonparametric_2008,linderman_recurrent_2016,linderman_bayesian_2017, linderman_structure-exploiting_2017, alameda-pineda_variational_2022}.
However, the underlying assumptions of these models and the complexity of the inference mechanisms these entail, often make their training challenging and impede their efficient application to DSR problems, especially when moving to real-world scenarios and higher-dimensional systems.

Here we propose almost-linear RNNs (AL-RNNs) which combine linear units and ReLUs, but can use as few of the latter as necessary to achieve a most parsimonious representation in terms of linear subregions. AL-RNNs are easy and effective to train by any SOTA algorithm for DSR. Through this, they are able to robustly identify topologically or geometrically minimal representations of well-known chaotic systems. Their structure translates naturally into a symbolic coding that preserves important topological properties. These features make AL-RNNs highly interpretable and mathematically tractable, enabling to harvest tools from symbolic dynamics \citep{osipenko_applied_1998, Lind_Marcus_2021}, including the representation of empirically observed DS via minimal topological and computational graphs.

\section{Related Work}
\paragraph{Dynamical systems reconstruction}

The field of data-driven DSR has been rapidly expanding in recent years. On the one hand there are approaches based on function libraries for approximating unknown vector fields, which have become particularly popular in some areas like physics \cite{la_cava_contemporary_2021, makke_interpretable_2024}. Among these, Sparse Identification of Nonlinear Dynamics (SINDy) and its variants \cite{brunton_discovering_2016, loiseau_constrained_2018, kaiser_sparse_2018, cortiella_sparse_2021, messenger_weak_2021,kaheman_automatic_2022} is probably the most popular. Since in these models sets of differential equations are directly formulated in terms of known, predefined function libraries, instead of using NN black-box approximators, they have some level of interpretability in the sense that they are human-readable and can easily be related to established mathematical building blocks in physical or biological theories \cite{heim_rodent_2019, rudin_stop_2019}. This does not necessarily make them mathematically tractable, however, since systems of nonlinear differential equations are in themselves usually hard to analyze (in fact, their behavior is much of the core topic of DS theory \cite{perko_differential_2001}). 
They also have other limitations, including a difficulty in capturing complex and noisy empirical data \cite{brenner_tractable_2022, hess_generalized_2023}, as they usually require considerable prior knowledge about the system's underlying structure (i.e., which terms to include in the function library). This somewhat limits their applicability for discovering novel phenomena. On the other hand, many recent powerful DSR methods rely on universal approximators, in particular the fact that sufficiently large RNNs can approximate any underlying DS \cite{funahashi_approximate_1989, kimura1998learning, pmlr-v120-hanson20a}. Such methods may be grouped into several broad classes, including reservoir computers \cite{pathak_using_2017, platt_systematic_2022,platt2023constraining}, neural ODEs/PDEs \cite{chen_neural_2018, karlsson_modelling_2019,alvarez_dynode_2020, ko_homotopy-based_2023}, neural/ Koopman operators \cite{brunton_modern_2021, lusch_deep_2018, otto_linearly-recurrent_2019, azencot_forecasting_2020, naiman_koopman_2021, geneva_transformers_2022, 
wang_koopman_2022}, and RNNs \cite{trischler_synthesis_2016, durstewitz_state_2017,vlachas_data-driven_2018, cestnik_inferring_2019, brenner_tractable_2022, rusch_long_2022, hess_generalized_2023}. The latter are commonly trained by variants of backpropagation through time (BPTT, \cite{vlachas_data-driven_2018, vlachas_backpropagation_2020, brenner_tractable_2022}), combined with specific control techniques \citep{mikhaeil_difficulty_2022} to remedy the exploding/ vanishing gradients problem \citep{bengio_learning_1994, mikhaeil_difficulty_2022, brenner_tractable_2022, hess_generalized_2023}. While DSR algorithms based on universal approximators achieve SOTA performance on DSR tasks, and often work particularly well on empirical time series \cite{brenner_tractable_2022, hess_generalized_2023}, they commonly deliver a complex model structure that is difficult to interpret and parse mathematically.

\paragraph{Nonlinear dynamics via linear DS}
The idea of approaching nonlinear DS through our good grasp of linear DS has been around for quite a long time, reflected in important theoretical results like the Hartman-Grobman theorem \cite{hartman_lemma_1960}\footnote{The Hartman-Grobman theorem states that nonlinear hyperbolic DS are topologically conjugate to a linear DS in some neighborhood of the system's equilibrium points.} or Koopman operator theory \cite{koopman_dynamical_1932, brunton_modern_2021}. While linear DS are easy to analyze and well understood \citep{perko_differential_2001,strogatz_nonlinear_2015}, however, they cannot properly capture most real-world systems, as they cannot produce many important DS phenomena such as limit cycles, chaos, or multistability. This has motivated the modeling of complex dynamics in terms of compositions of locally linear dynamics, as the next best %thing so-to-speak
alternative, i.e. piecewise-linear (PWL) maps or ODE systems \citep{casdagli_nonlinear_1989, ives_detecting_2012, costa_adaptive_2019}. PWL models have been popular in engineering and mathematics for several decades for these reasons, including earlier attempts for learning such models directly from data \cite{storace_piecewise-linear_2004, de_feo_piecewise-linear_2007}. Switching linear DS are one particular brand of PWL models with a long tradition in DS and control theory \cite{daafouz_stability_2002, sun_switched_2006,ackerson_state_1970, ghahramani_variational_2000, fox_nonparametric_2008}. These systems model nonlinear dynamics through a set of linear (or affine) DS combined with a switching rule which decides which linear DS is currently active. Likewise, in mathematics PWL models served well for investigating generic properties of nonlinear systems, e.g. the tent map which is topologically conjugate to the logistic map \cite{alligood_chaos_1996}. PWL models often lend themselves to particularly convenient symbolic representations \cite{lind_introduction_1995,alligood_chaos_1996}, based on which important topological properties of the underlying system, e.g. the nature and number of unstable periodic orbits embedded within a chaotic attractor, can be analyzed \cite{mischaikow_construction_1999,osipenko_applied_1998,tomasrodriguez_linear_2003}.

More recently, various modern approaches for inferring PWL models from data have been formulated.
For instance, switching state space models combine hidden Markov models with linear DS, jointly inferring the state of a switching (random) variable with the linear DS parameters conditioned on these states \cite{ghahramani_variational_2000}. Various extensions of this basic setup like recurrent and hierarchical switching linear DS and fully Bayesian inference methods have been advanced in recent years \cite{stanculescu_hierarchical_2014,linderman_recurrent_2016, linderman_bayesian_2017}. However, inference in these models is often complex and not necessarily optimized for DSR, limiting their applicability to mainly low-dimensional scenarios with comparatively simple dynamics.
Most of these models are also discontinuous in their dynamics across switches, while most commonly we would require the state variables to evolve continuously across the whole of state space. 
Piecewise-linear RNNs (PLRNNs), and, relatedly, threshold-linear networks \citep{hahnloser_permitted_2000, wersing_dynamical_2001,yi_multistability_2003}, on the other hand, are based on familiar ReLUs and hence change continuously across their switching manifolds \cite{durstewitz_state_2017,koppe_identifying_2019,schmidt_identifying_2021}. They also have some biological justification \citep{hahnloser_permitted_2000,durstewitz_state_2017}. Commonly they are trained by variants of BPTT backed up by specific control-theoretic approaches like sparse \cite{mikhaeil_difficulty_2022} or generalized \cite{hess_generalized_2023} teacher forcing which make them SOTA on many DSR tasks. Different PLRNN architectures have been proposed to enhance expressivity or reduce the dimensionality of trained models \cite{brenner_tractable_2022,hess_generalized_2023}. Yet, while these advances may yield comparatively low-dimensional state spaces, the number of different linear subregions that need to be allocated usually remains very high, hampering efficient mathematical analysis.

\section{Methodological and Theoretical Prerequisites}

\subsection{AL-RNN Model}
Consider a piecewise linear recurrent neural network (PLRNN, \cite{durstewitz_state_2017}):

\begin{align}\label{eq:plrnn_lat}
	\bm{z}_t = F_{\bm{\theta}}(\bm{z}_{t-1})=\bm{A} \bm{z}_{t-1} + \bm{W}  \phi(\bm{z}_{t-1}) + \bm{h},% + \bm{C s}_t, %+ \bm{\epsilon}_t, \quad \bm{\epsilon}_t \sim\mathcal{N}(\bm{0},\bm{\Sigma}),
	\end{align}
where diagonal $\bm{A} \in \mathbb{R}^{M \times M}$ contains linear self-connections, %and the entries of 
$\bm{W} \in \mathbb{R}^{M \times M}$ are nonlinear connections between units, $\bm{h} \in \mathbb{R}^{M}$ is a bias term, and $\phi(\bm{z})=\max[0,\bm{z}]$ is an element-wise ReLU nonlinearity. To expose the piecewise linear structure of this model more clearly, by noting that the slope of the ReLU is either $0$ or $1$ depending on the sign of $z_{m,t}$, one can reformulate this as
\begin{align}\label{eq-th-1}
 \vz_t &= (\mA + \mW \mD_{\Omega(t-1)}) \vz_{t-1} + \vh := \mW_{\Omega(t-1)} \, \vz_{t-1} + \vh,
\end{align}
where $\mD_{\Omega(t)} := diag(\vd_{\Omega(t)})$ is a diagonal matrix and $\vd_{\Omega(t)}=\left(d_1, d_2, \cdots, d_M \right)$ an indicator vector with $d_m(z_{m,t})=1$ whenever $z_{m,t}>0$ and zero otherwise \citep{eisenmann2023bifurcations}. For the $2^M$ different configurations of $\mD_{\Omega(t)}$, $\mD_{\Omega^k}$, $k \in \lbrace 1,2, \cdots, 2^{M} \rbrace$, the phase space of system eq. \ref{eq-th-1} is divided into $2^M$ subregions with linear dynamics
\begin{align} \label{eq-subreg}
 \vz_{t+1}\, = \, \mW_{\Omega^k} \, \vz_{t} + \, \vh, \hspace{1cm} \mW_{\Omega^k}:= \mA +\mW \mD_{\Omega^k}.
\end{align}

Empirically, $M$ often needs to be quite large (at least on the order of the number of observations) for achieving good reconstructions of observed DS. Since the number of subregions grows as $2^M$, analyzing inferred models in terms of the subregions can thus become very challenging. We therefore introduce a novel variant of the PLRNN in which only a subset of $P << M$ units are equipped with a ReLU nonlinearity, yielding
%applies the piecewise nonlinearity on $P$ neurons, defining: 
\begin{align}\label{eq:alrnn}
	\bm{z}_t = \bm{A} \bm{z}_{t-1} + \bm{W}  \Phi^*(\bm{z}_{t-1}) + \bm{h}
\end{align}
where
\begin{equation}
\Phi^*(\bm{z}_{t}) = \left[ z_{1,t}, \cdots, z_{M-P,t}, \max(0, z_{M-P+1,t}), \cdots, \max(0, z_{M,t}) \right]^T.
\end{equation}

In this formulation, we thus only have $2^P$ different linear subregions, while still accommodating a sufficiently large number of latent states for capturing unobserved dimensions in the data and disentangling trajectories sufficiently \cite{takens_detecting_1981, sauer_embedology_1991}.\footnote{Strictly, in this formulation, for the linear units the diagonal entries in $\bm{W}$ are redundant to those in $\bm{A}$ and could be omitted, but we found in practice this hardly makes a difference.}

The model is trained on the $N$-dimensional observations $\{\bm{x}_t\}, t=1 \dots T, \bm{x}_t \in \mathbb{R}^N$, by a variant of sparse teacher forcing called identity teacher forcing \cite{mikhaeil_difficulty_2022, brenner_tractable_2022}. In identity teacher forcing, the first $N$ latent states (`readout neurons') are replaced by the $N$-dimensional observations every $\tau$ time steps, where $\tau$ is chosen such as to optimally control trajectory and gradient flows, avoiding exploding gradients while providing the model sufficient opportunity to unroll into the future to capture the underlying DS' long-term behavior (see \cite{mikhaeil_difficulty_2022,hess_generalized_2023} for details); see Appx. \ref{appx:stf_training} for details on training. We emphasize that sparse teacher forcing is \textit{only used for training} the model, and is turned off at test time where the model generates new trajectories completely independent from the data.
\begin{figure*}[!htb]
    \centering
	\includegraphics[width=0.99\linewidth]{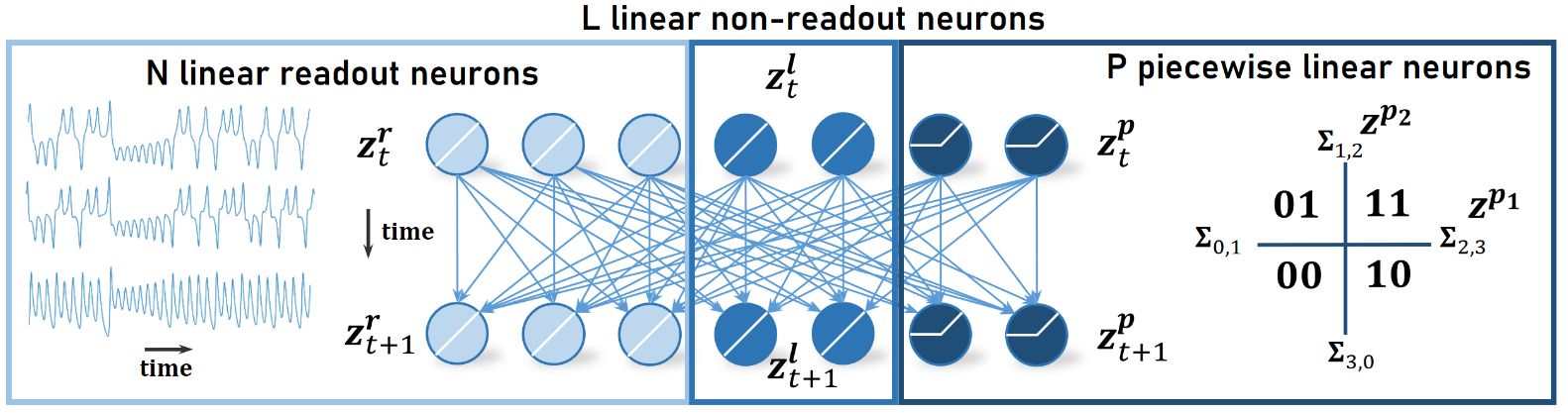}
	\caption{Illustration of the AL-RNN architecture.}
	\label{fig:rnn_architecture}
\end{figure*}

\subsection{Theoretical Background: Symbolic Dynamics and Symbolic Coding of AL-RNN} \label{sec:methods_symbolic_approach}
The mathematical field of symbolic dynamics formulates conditions under which a DS has a unique symbolic representation and discusses how to harvest this symbolic representation to prove certain properties of the underlying system, which otherwise may be more difficult to address \cite{osipenko_applied_1998,Lind_Marcus_2021}. In fact, symbolic dynamics has led to many powerful insights and formal results in DS theory, e.g. about the properties of chaos or type and number of periodic orbits \citep{guckenheimer_nonlinear_1983, wiggins_global_1988}. An appealing feature of symbolic dynamics for the field of ML/AI is that it links concepts in DS theory to computational concepts like finite state automata or formal languages, as well as graph theory \cite{Lind_Marcus_2021,hao1998applied}. It can thus facilitate the computational interpretation of natural or trained dynamical systems, like RNNs. 

Assume we have an alphabet of $n$ symbols $\mathcal{A}=\{0, \dots, n-1\}$, from which we form infinite sequences (bidirectionally or only in forward-time) $\bm{a}=\dots a_{-2} a_{-1}. a_{0} a_{1} a_{2} \dots$ with $a_k \in \mathcal{A}$, and the dot separating past from future (i.e., indices $k<0$ indicate backward time, and $k \geq 0$ present and forward time). Then the space of all possible sequences, together with the so-called (left) shift operator given by
\begin{align}\label{eq:shift}
    \sigma(\bm{a})=\sigma(\dots a_{-2} a_{-1}. a_{0} a_{1} a_{2} \dots)=\dots a_{-1} a_{0}. a_{1} a_{2} a_{3} \dots
\end{align}
defines the full shift space $\mathcal{A}^{\mathbb{Z}}$. We denote by $\sigma^k = \sigma \circ \sigma \circ \dots \circ \sigma$ the $k$-times iteration of the shift. Now consider a DS $(S,\phi)$ consisting of a %compact 
metric (state) space $S$ and a recursive (flow) map $\phi$. The flow map $\phi_{\Delta t}(\bm{x})$ advances the system's current state $\bm{x}$ by $\Delta t$ and may be thought of as the solution operator of the underlying DS $\dot{\bm{x}}=f(\bm{x})$ \cite{perko_differential_2001}. When training RNNs $\bm{z}_t=F_{\bm{\theta}}(\bm{z}_{t-1})$ on time series of observations $\{g(\bm{x}_{k \Delta t})\}, k=1 \dots T$, from the underlying DS, where $g$ is the observation function, we are %essentially 
trying to approximate this flow map. Assume the whole state space $S$ can be partitioned into a finite set $\mathcal{U}=\{U_0 \dots U_{n-1}\}$ of disjoint open sets $U_e$, such that $S = \bigcup_{e=0}^{n-1} \overline{U_e}$, i.e. $S$ is covered by the union of the closures of these sets. We call this a \textit{topological partition} of $S$ \cite{Lind_Marcus_2021}. 

\begin{figure*}[!htb]
    \centering
	\includegraphics[width=0.99\linewidth]{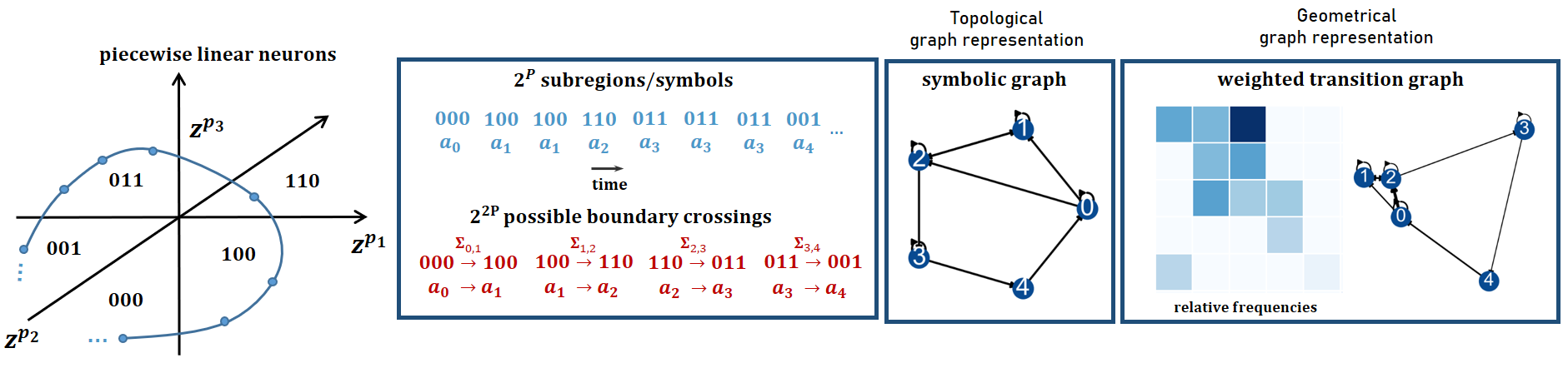}
	\caption{Illustration of symbolic approach (3 panels on the left) and geometrical graphs (right).}
	\label{fig:symbolic_approach}
\end{figure*}

The central idea now is to assign a unique symbol $a_e \in \mathcal{A}$ to each set $U_e \in \mathcal{U}$, with $n=|\mathcal{U}|=|\mathcal{A}|$. As a trajectory $\bm{x}(t)$ of the underlying DS travels through the system's state space $S$, observed at times $k \Delta t$ as it passes through different subregions $U_e$, %on its journey, 
it gives rise to a specific symbolic sequence $\bm{a}_{\bm{x},\phi}$ (with a unique symbol assigned at each time step via $h: S \rightarrow \mathcal{A}, \bm{x}_{k \Delta t} \mapsto a_k$). We may thus think of the shift operator $\sigma$ as moving along a trajectory in correspondence with the flow map $\phi_{\Delta t}(\bm{x})$. If the symbolic coding of each trajectory is unique, $\mathcal{U}$ may constitute a \textit{Markov partition} (see Appx. \ref{proofs} for a formal definition). We denote by $(A_{S,\phi},\sigma)$ the \textit{shift of finite type} induced by the flow $\phi$ which picks out from the full shift space $\mathcal{A}^{\mathbb{Z}}$ only those \textit{admissible} symbolic sequences that correspond to valid trajectories of $(S,\phi)$ (we will use the term `induced by' to refer to this property). The set of admissible blocks constitutes the \textit{language} of $(A_{S,\phi},\sigma)$. Every shift of finite type has a graph representation $\mathcal{G}=\{\mathcal{V},\mathcal{E}\}$ (Fig. \ref{fig:symbolic_approach}), with either the edges $e_{ij} \in \mathcal{E}$ or vertices $v_i \in \mathcal{V}$ of the graph encoding the permitted transitions among symbols $a_k \in \mathcal{A}$ of admissible sequences $\bm{a} \in A$ \cite{Lind_Marcus_2021}. 

The finite collection $\mathcal{U}=\{U_0 \dots U_{n-1}\}, \; n = 2^P$, of \textit{linear subregions} of an AL-RNN defined in eq. \ref{eq:alrnn}, separated by the switching manifolds $\Sigma_{i,j} = \overline{U_i} \cap \overline{U_j}$ between every pair of neighboring subregions $U_i$ and $U_j$, forms a topological partition.
Here we use this partition as the basis of our symbolic coding and the respective symbolic dynamics $(A_{\mathcal{U},F_{\bm{\theta}}},\sigma)$ induced by the AL-RNN, where we assign to each state $\bm{z}_t \in S$ (i.e., at each time point) the unique symbol $a_t \in \mathcal{A}$ such that $a_t=a_i$ iff $\bm{z}_t \in U_i$ (Fig. \ref{fig:symbolic_approach}).\footnote{For simplicity we will ignore here and in the following the borders between subregions, on which the coding is ambiguous.} In the corresponding symbolic graphs, we identify vertices $v_i$ with symbols $a_i \in \mathcal{A}$ and draw a directed edge $e_{ij}$ from $v_i$ to $v_j$ whenever $F_{\bm{\theta}}(U_i \cap B) \cap U_j \neq \varnothing$, where $B$ is the attracting set of interest (see Appx. \ref{appx:graph_repr} for details).
As we will show further below, this particular partition has useful theoretical properties that makes the symbolic coding topologically interpretable w.r.t. the AL-RNN map $F_{\bm{\theta}}$. In fact, a large literature in symbolic dynamics has dealt with the relation between the dynamics in a finite shift space and that of a PWL map, like, e.g., the tent map, with a partition of $S$ into the map's different linear subregions as we use here for the AL-RNN \cite{Lind_Marcus_2021, alligood_chaos_1996, milnor_concept_1985}.

\section{Theoretical Results}
Recall that within each subregion $U_e$ the map $F_{\bm{\theta}}$ is monotonic and the dynamics are linear (ruling out certain possibilities, like chaos or isolated cycles occurring within just one subregion). We furthermore assume that the dynamics are globally non-diverging (this could be strictly enforced through `state-clipping' and constraints on matrix $\bm{A}$ in eq. \ref{eq:alrnn}, see \citet{hess_generalized_2023}, but will also be the case for a well-trained AL-RNN). %, but empirically rarely occurred in well-trained AL-RNNs anyway). 
Here we claim that for %uniformly 
hyperbolic AL-RNNs $F_{\bm{\theta}}$\footnote{By %\textit{uniformly hyperbolic} 
\textit{hyperbolic AL-RNN} we mean the AL-RNN is hyperbolic in each of its linear subregions, i.e. none of its Jacobians $\mA + \mW \mD_i$ has eigenvalues of absolute magnitude $1$. The chances that this condition is \textit{not} met in practice in trained AL-RNNs, i.e. the non-hyperbolic case, are close to zero numerically.}, we have 1:1 relations between important topological objects in the AL-RNN's state space and those of the symbolic coding formed from the linear subregions $U_e$ of the AL-RNN, as expressed in the following theoretical results.

Consider a %uniformly 
hyperbolic, non-globally-diverging AL-RNN $F_{\bm{\theta}}$, eq. \ref{eq:alrnn}, and a topological partition $\mathcal{U}$ of the state space into its linear subregions $U_e \subseteq S, e=0 \dots 2^P-1$. Denote by $(A_{\mathcal{U},F_{\bm{\theta}}},\sigma)$ the finite shift induced by $(S,F_{\bm{\theta}})$, with each $a_e \in \mathcal{A}$ of its alphabet $\mathcal{A}$ associated with exactly one linear subregion $U_e \in \mathcal{U}$, and let us consider the system's evolution only in forward time. Then the following holds:

%---------------------------------
\begin{theorem}\label{FPsymbPLRNN}
An orbit $\Omega_S=\{\bm{z}_1, \ldots, \bm{z}_n, \ldots\}$ of the AL-RNN $F_{\bm{\theta}}$ is asymptotically fixed (i.e., converges to a fixed point) if and only if the corresponding symbolic sequence $\bm{a} \, = \, (a_{1} a_{2} a_{3}\dots a_{N-1})(a^*)^{\infty} \in A_{\mathcal{U},F_{\bm{\theta}}}$ is an eventually fixed point of the shift map $\sigma$ (where by `eventually' we mean it exactly lands on the point in the limit, see Appx. \ref{proofs} for a precise definition).
\end{theorem}
\begin{proof}
See Appx. \ref{proofs}.
\end{proof}

\begin{theorem}\label{CyclesymbPLRNN}
An orbit $\Omega_S=\{\bm{z}_1, \ldots, \bm{z}_n, \ldots\}$ of the AL-RNN $F_{\theta}$ is asymptotically $p$-periodic if and only if the corresponding symbolic sequence 
$$\bm{a} \, = \, (a_{1} a_{2} \ldots a_{N-1})(a^*_1 a^*_2 \ldots a^*_{p})^{\infty} \in A_{\mathcal{U},F_{\bm{\theta}}}$$ is an eventually $p$-periodic orbit of the shift map $\sigma$.
\end{theorem}
\begin{proof}
See Appx. \ref{proofs}.
\end{proof}

\begin{theorem}\label{ChaossymbPLRNN}
 An orbit $\Omega_S=\{\bm{z}_1, \ldots, \bm{z}_n, \ldots\}$ is an asymptotically \textbf{aperiodic} (irregular) orbit of the AL-RNN $F_{\bm{\theta}}$   
if and only if the corresponding symbolic sequence $(a_1, \ldots, a_n, \ldots)$ is aperiodic.
\end{theorem}
\begin{proof}
See Appx. \ref{proofs}.
\end{proof}

Loosely speaking, these results confirm that fixed points of our symbolic coding correspond to fixed points of the AL-RNN, cycles to cycles, and chaos to chaos, %preserving crucial topological properties.
thus preserving important topological properties in the symbolic representation.

\section{Experimental Results}

To assess the quality of DSR, we employed established performance criteria based on long-term, invariant topological, geometrical, and temporal features of %the reconstructed 
DS \cite{koppe_identifying_2019,brenner_tractable_2022,hess_generalized_2023}. %, to assess the quality of DSR of trained models. 
Due to exponential trajectory divergence %present 
in chaotic systems, mean-squared prediction errors rapidly %diverge
grow even for well-trained systems, and hence are only of limited suitability for evaluating DSR quality \cite{wood_statistical_2010, mikhaeil_difficulty_2022}. Thus, we prioritize the geometric agreement between true and reconstructed attractors, quantified by a Kullback-Leibler divergence ($D_{\text{stsp}}$, Appx. \ref{appx:geometric_agreement}) \cite{koppe_identifying_2019}. Additionally, we examine the long-term temporal agreement between true and reconstructed time series by evaluating the average dimension-wise Hellinger distance ($D_{\text{H}}$) between their power spectra (Appx. \ref{appx:temporal_agreement}). We first confirmed that the AL-RNN is at least on par with other SOTA methods for DSR. We then tested AL-RNNs on two commonly employed benchmark DS for which minimal PWL representations are known, the famous Lorenz-63 model of atmospheric convection \cite{lorenz_deterministic_1963} and the chaotic Rössler system \cite{rossler_equation_1976}. We finally explored the suitability of our approach on two real-world examples, human electrocardiogram (ECG) and human functional magnetic resonance imaging (fMRI) data.

\subsection{SOTA performance}

While our goal here is a technique that constructs topologically minimal, interpretable DS representations, at the same time we do not want to compromise on DSR performance which should still be within the same ballpark as existing SOTA methods. We checked this on the Lorenz-63, Rössler and ECG data noted above, and in addition on the higher-dimensional chaotic Lorenz-96 system \cite{lorenz_predictability_1996} and on human electroencephalogram (EEG) data. Table \ref{tab:gtf_benchmarks} confirms that the AL-RNN is not only on par with, but indeed outperforms most other techniques when trained with sparse teacher forcing (which may be rooted in its simple and parsimonious design).

\subsection{Reconstructed Systems Occupy a Small Number of Subregions}

\begin{figure*}[!htb]
    \centering
	\includegraphics[width=0.99\linewidth]{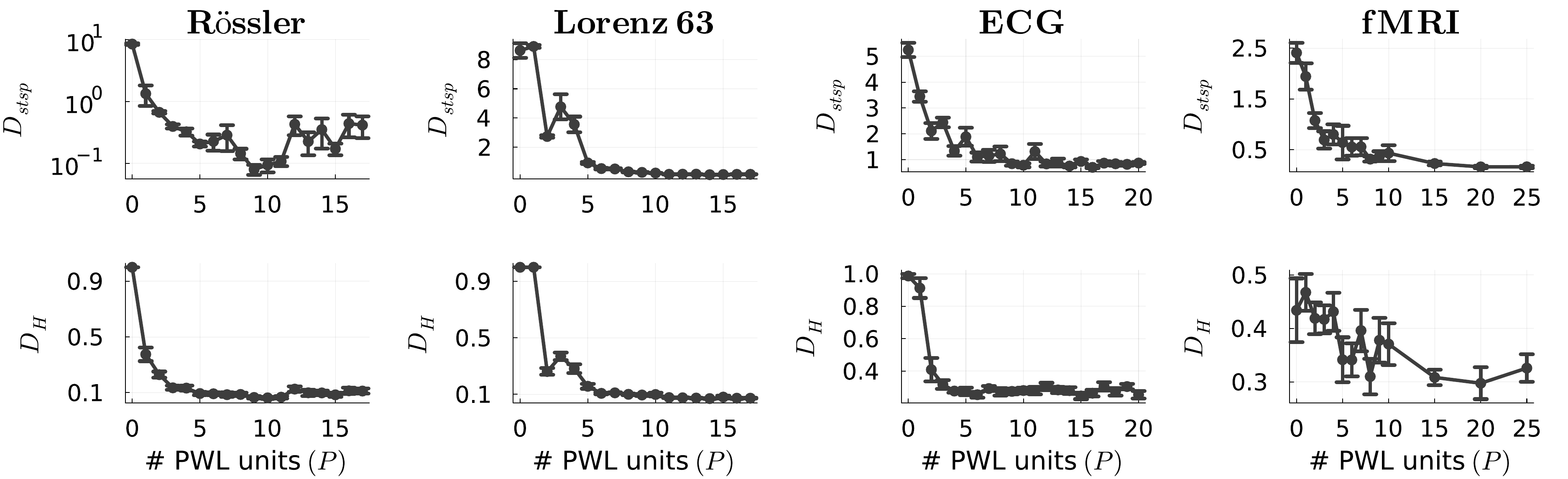}
	\caption{Quantification of DSR quality in terms of attractor geometry disagreement ($D_{\text{stsp}}$, top row) and disagreement in temporal structure ($D_{\text{H}}$, bottom row) as a function of the number of ReLUs ($P$) in the AL-RNN (%with the total number of units constant, 
    Rössler: $M=20$, Lorenz-63: $M=20$, ECG: $M=100$, fMRI: $M=50$). The little humps at $P=3$ for the Lorenz-63 indicate that performance may sometimes first degrade again when passing the number of minimally necessary PWL units (see also Fig. \ref{fig:regularization}). Error bars = SEM. %Reconstruction performance plateaus at a sufficiently large number of PWL units.
    }
	\label{fig:performance_relu}
\end{figure*}
\begin{figure*}[!htb]
    \centering
    \begin{subfigure}[b]{0.49\textwidth}
         \includegraphics[width=\textwidth]{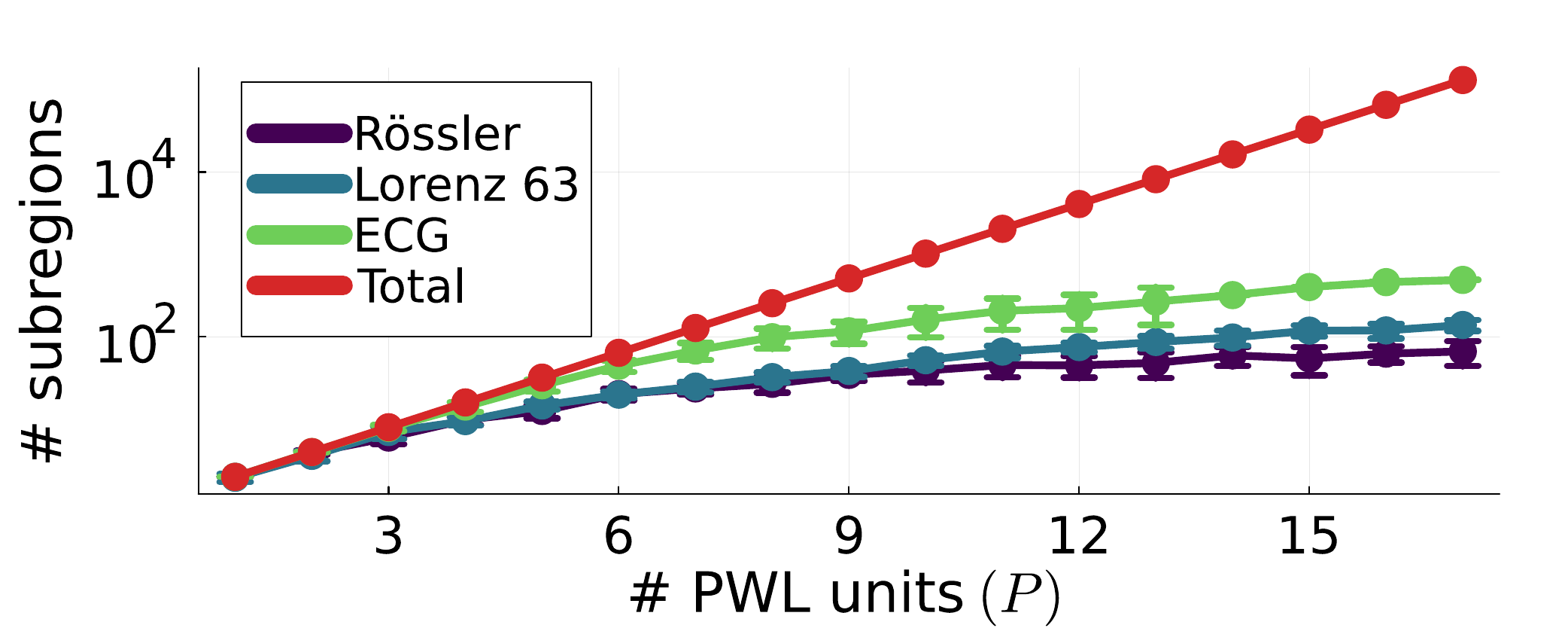}
    \end{subfigure}
    \hfill
    \begin{subfigure}[b]{0.49\textwidth}
         \includegraphics[width=\textwidth]{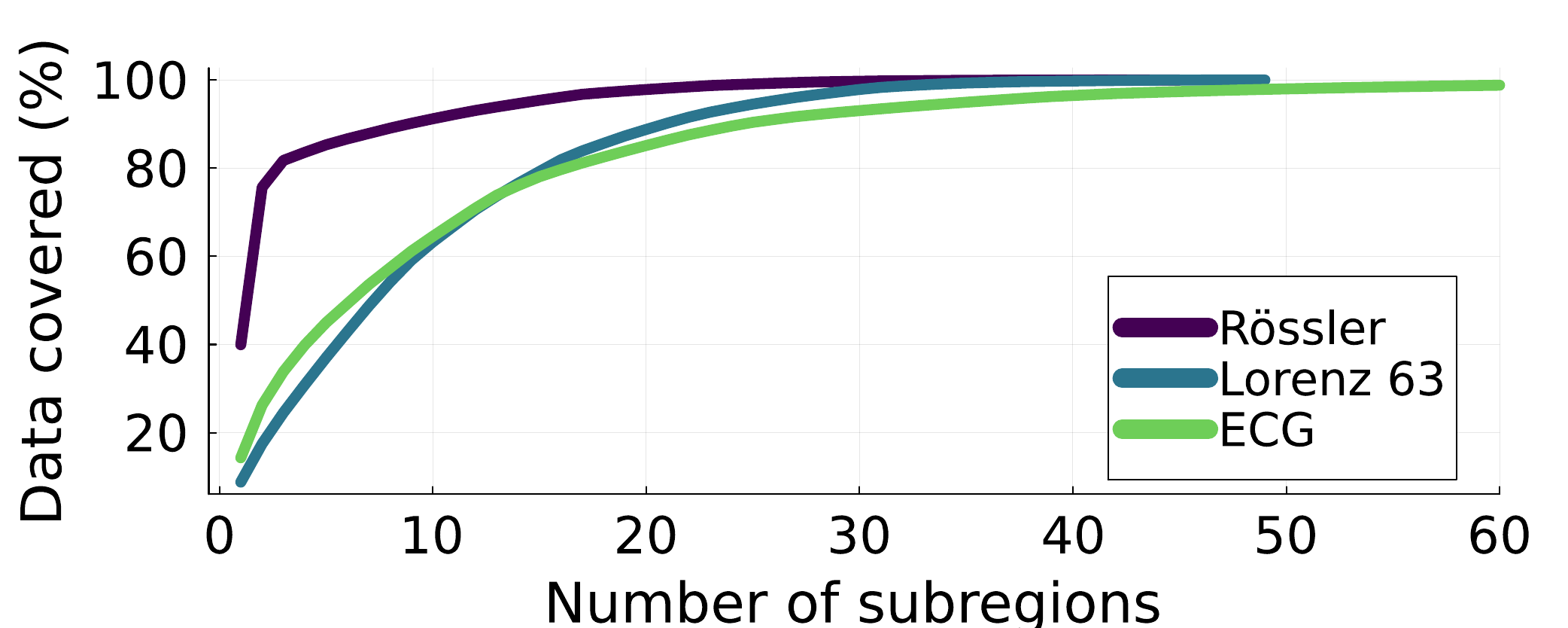}
    \end{subfigure}
    \caption{Left: Number of linear subregions traversed by trained AL-RNNs as a function of the number $P$ of ReLUs. Theoretical limit ($2^P$) in red. Right: Cumulative number of data (trajectory) points covered by linear subregions in trained AL-RNNs (Rössler: $M=20, P=10$, Lorenz-63: $M=20, P=10$, ECG: $M=100, P=10$), illustrating that trajectories on an attractor live in a relatively small subset of subregions.}
    \label{fig:subregions_covered}
\end{figure*}

Fig. \ref{fig:performance_relu} illustrates reconstruction performance for varying numbers of ReLU nonlinearities at constant network size $M$. We found that a small number of PWL units already significantly improves performance, especially for the Lorenz-63 and Rössler systems, and that beyond that number performance starts to plateau (or even briefly decrease again). 
Additionally, some linear units are necessary to sufficiently expand the space, but they cannot compensate for an insufficient number of PWL units (Fig. \ref{fig:performance_M_Lorenz63}).\footnote{Note that the number of linear units does not affect the interpretability or symbolic coding of the model.}
Moreover, as shown in Fig. \ref{fig:subregions_covered} (left), the number of linear subregions explored by the trained dynamics saturates well below the theoretical limit of $2^P$ once this performance threshold is reached. Within this already small subset of explored subregions, generated network activity %of trained models 
is furthermore concentrated within an even smaller number of dominant subregions: For instance, for the Rössler system $4$ out of $45$ subregions used cover $80\%$ of the data (Fig. \ref{fig:subregions_covered}, right).
This substantial reduction of necessary linear subregions strongly facilitates the analysis of trained models with respect to fixed points and $k$-cycles, which naively would require examining $2^P$ and $2^{kP}$ combinations of subregions, respectively.
To select the optimal number of PWL units, the point where performance starts plateauing (as in Fig. \ref{fig:performance_relu}) may be chosen. Alternatively, one may restrict the number of linear subregions employed through regularization, adding a penalty for ReLU nonlinearities. %(Appx. ***).
This approach, as Fig. \ref{fig:regularization} illustrates, results in the same number of selected PWL units.

%\subsection{Topologically minimal reconstructions}
\subsection{Minimal PWL Reconstructions of Chaotic attractors}\label{sec:topologically_minimal}
\paragraph{Topologically minimal reconstructions}
 
\begin{figure*}[!htb]
    \centering
	\includegraphics[width=0.99\linewidth]{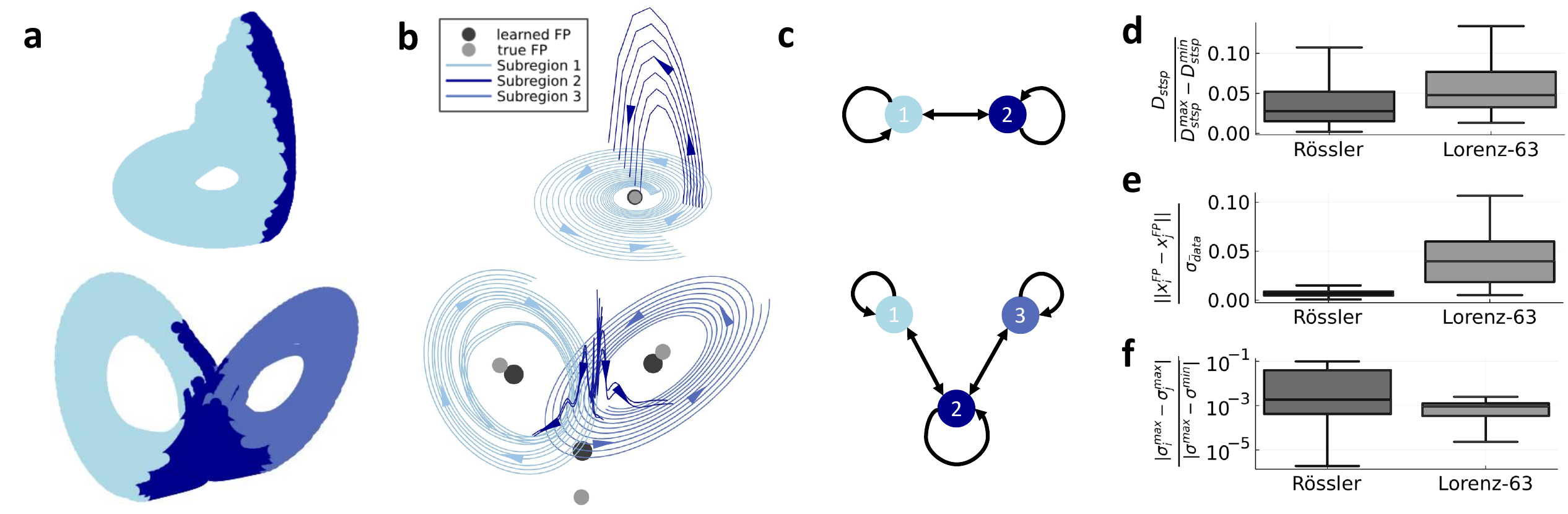}
	\caption{\textbf{a}: Color-coded linear subregions %(mapped to observation space) 
    of minimal AL-RNNs representing the Rössler (top) and Lorenz-63 (bottom) chaotic attractor. \textbf{b}: Illustration of how the AL-RNN creates the chaotic dynamics. For the Rössler, trajectories diverge from an unstable spiral point (true position in gray, learned position in black) into the second subregion, where
    after about half a cycle 
    they are propelled back into the first. For the Lorenz-63, two unstable spiral points (true: gray; learned: black) create the diverging spiraling dynamics in the two lobes, separated by the 
    %stable manifold of a 
    saddle node in the center. \textbf{c}: Topological graphs of the symbolic coding. While for the Rössler it is fully connected, for the Lorenz-63 the crucial role of the center saddle region in distributing trajectories onto the two lobes is apparent. \textbf{d}: Geometrical divergence ($D_{\text{stsp}}$) among repeated trainings of AL-RNNs ($n=20$), separately evaluated within each subregion, shows close agreement among different training runs. Likewise, low \textbf{e}: normalized distances between fixed point locations and \textbf{f}: relative differences in maximum absolute eigenvalues $\sigma^{\text{max}}$ across $20$ trained models indicate that these topologically minimal representations are robustly identified.}
	\label{fig:minimal_reconstructions}
\end{figure*}

Investigating reconstructions with the minimal number of PWL units needed for close-to-optimal performance (Fig. \ref{fig:performance_relu}), we found that the AL-RNN would deliver reconstructions capturing the overall structure of the attractor using only three (Lorenz-63 system) or two (Rössler system) linear subregions (Fig. \ref{fig:minimal_reconstructions}\textbf{a}), explaining the strong performance gains in Fig. \ref{fig:performance_relu} for $2$ and $1$ PWL units, respectively. These representations, and their symbolic coding (Fig. \ref{fig:minimal_reconstructions}\textbf{c}), expose the mechanisms of chaotic dynamics (Fig. \ref{fig:minimal_reconstructions}\textbf{b}). Notably, these closely agree with %correspond to 
the minimal topologically equivalent PWL representations of the two chaotic DS as described in %the literature 
\citet{amaral_piecewise_2006}: The Lorenz-63 system has at its core two unstable spiral points in the two lobes, separated by the %manifold of a 
saddle node in the center (Fig. \ref{fig:minimal_reconstructions}\textbf{b}).
For the Rössler system, the topologically minimal PWL representation indeed consists of just two subregions \cite{amaral_piecewise_2006}, one containing an unstable spiral in the $x$-$y$ plane and the other a `half-spiral' almost orthogonal to that plane %unstable spiral in the z-plane 
(Fig. \ref{fig:minimal_reconstructions}\textbf{b}). The AL-RNN automatically and robustly discovers these representations from data: across multiple training runs, performance values are very similar (Figs. \ref{fig:chi2_comparison}, \ref{fig:minimal_subregions_Lorenz63}), the assignment of subregions to different parts of the attractor remains almost the same (Figs. \ref{fig:minimal_subregions_Rossler} $\&$ \ref{fig:minimal_subregions_Lorenz63}), and the regions with linear dynamics closely agree both in terms of their topology and geometry (in fact, the topological graphs remained identical). This is in contrast to the standard PLRNN, where assignments strongly varied among multiple training runs (Figs. \ref{fig:chi2_comparison}, \ref{fig:minimal_subregions_Lorenz63}). We quantified this further by computing across training runs separately for each subregion $D_{\text{stsp}}$ (Fig. \ref{fig:minimal_reconstructions}\textbf{d}), the normalized distances between fixed points (Fig. \ref{fig:minimal_reconstructions}\textbf{e}), and the normalized differences between the maximum absolute eigenvalues $\sigma_{\text{max}}$ of the AL-RNN's Jacobians (Fig. \ref{fig:minimal_reconstructions}\textbf{f}), obtaining values close to zero in all three cases (see Fig. %\ref{fig:minimal_reconstructions}\textbf{d-f}) and in absolute (Fig. 
\ref{fig:absolute_robustness} for absolute values). %(see Fig. A\ref{fig:minimal_subregions_Rossler} \& A\ref{fig:minimal_subregions_Lorenz63} for further examples).
While \citet{amaral_piecewise_2006} explicitly handcrafted such minimal PWL representations, the AL-RNN extracts them automatically without the provision of any prior knowledge about the system.

\paragraph{Geometrically minimal reconstructions}

\begin{figure*}[!htb]
    \centering
	\includegraphics[width=0.9\linewidth]{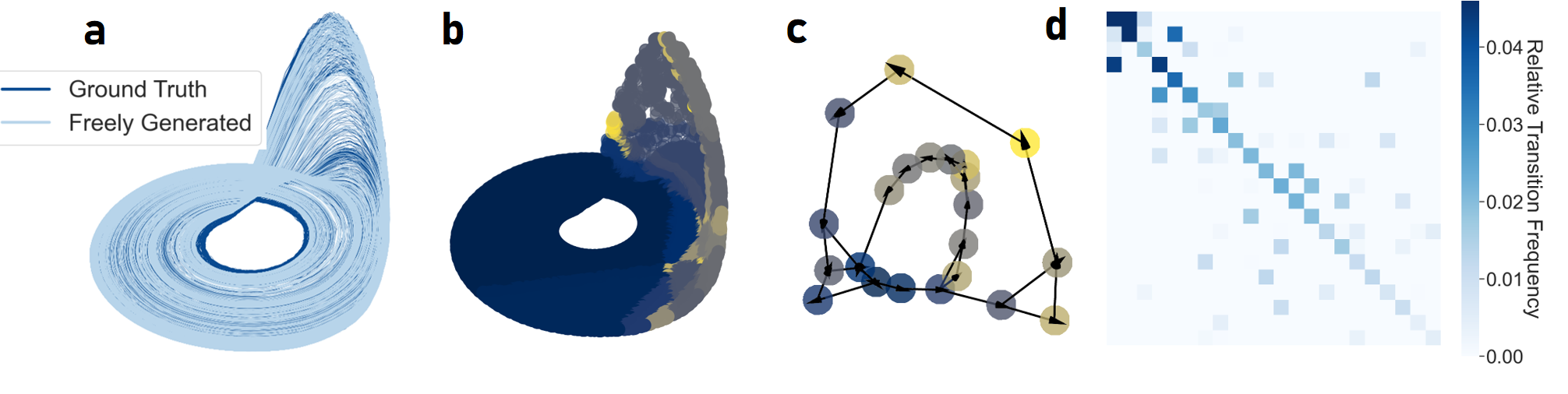}
	\caption{Geometrically minimal reconstruction and graph representation of the Rössler attractor ($M=30, P=10, D_{stsp}=0.08,  D_{H}=0.06$). \textbf{a}: Provided a sufficient number of linear subregions, the geometry of the attractor is almost perfectly captured. \textbf{b}: Reconstruction with %corresponding 
    linear subregions color-coded by frequency of visits (dark: most frequently visited regions, yellow: least frequent regions). \textbf{c}: Corresponding geometrical graph, which contains information about transition frequencies via node distances, %structure 
    visualized using the spectral layout in \texttt{networkx}. Note that self-connections were omitted in this representation. \textbf{d}: Connectome of relative transition frequencies between subregions.}
	\label{fig:graphs_roessler}
\end{figure*}

While these reconstructions capture the topology of the underlying DS, they do not yet %perfectly 
capture the full geometry and temporal structure of the attractor (Fig. \ref{fig:performance_relu}). 
%Increasing the number of PWL units to the threshold in Fig. \ref{fig:performance_relu} allows the model to capture finer details of attractor geometry. 
Fig. \ref{fig:graphs_roessler} illustrates for the Rössler system that as the number of PWL units is further increased to $P=10$, the geometrical agreement becomes almost perfect. Although the mapping from latent to observation space is not 1:1 (since $M>N$), points close in observation space still tended to fall into the same latent subregion, such that the observed system's attractor still decomposed into distinct subregions, as %assessed
confirmed by proximity matching (see Appx. \ref{appx:proximity_matching}). For the Rössler system there is just one nonlinearity, the $x\cdot z$ term in the temporal derivative of $z$ (eq. \ref{eq:roessler}). Accordingly, the AL-RNN devotes most of its subregions to the lobe along the $z$ coordinate, while dynamics in the $(x,y)$ plane is geometrically faithfully represented by only $4$ subregions. Hence, the AL-RNN utilizes additional subregions to express finer geometric details where dynamics are more nonlinear. This is apparent from a more geometrical graph representation (see Fig. \ref{fig:symbolic_approach}, right), where -- in addition to topological information -- transition probabilities among subregions are being used to construct node distances via the graph Laplacian (see Appx. \ref{laplacian_matrix}), see Fig. \ref{fig:graphs_roessler} for the Rössler and Fig. \ref{fig:graphs_examples} for the Lorenz-63.

\subsection{PWL Reconstructions of Real-World Systems}

\paragraph{Topologically minimal reconstructions}

We next considered two experimental datasets, human ECG data (with $1d$ membrane potential recordings delay-embedded into $N=5$, see Appx. \ref{appx:datasets}) and fMRI recordings (with $N=20$ time series extracted, cf. Appx. \ref{appx:datasets}) from human subjects performing three different types of cognitive task \cite{koppe_temporal_2014, kramer22a}. %We used $P=2$ for the ECG data and $P=3$ for the fMRI data.

\begin{figure*}[!htb]
    \centering
	\includegraphics[width=0.9\linewidth]{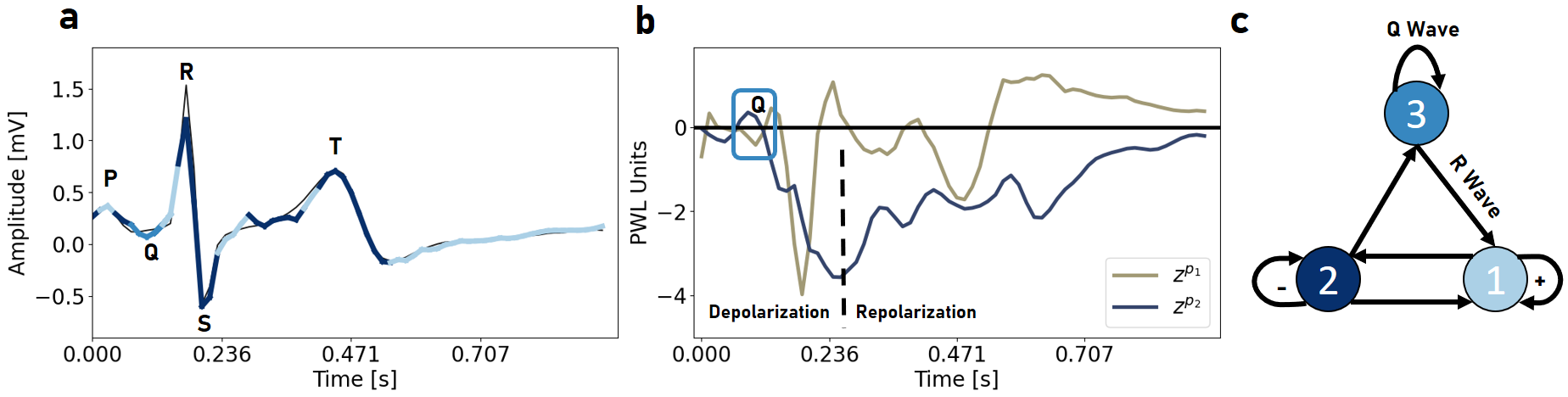}
	\caption{\textbf{a}: Freely generated ECG activity using an AL-RNN with 3 linear subregions (color-coded according to subregion) and ground truth time series in black. % aligning with ECG activation patterns. 
    \textbf{b}: After activation of the Q wave in the third subregion, the 
    %plunges the second PWL unit far beneath the subregion boundary, 
    second PWL unit is driven far below 0, 
    whose activity, consistent with the known physiology \cite{sodi-pallares_activation_1951}, mimics the latent de- and re-polarization process of the interventricular septum.
    \textbf{c}: Symbolic graph representation of the trained AL-RNN.}
	\label{fig:ECG_minimal}
\end{figure*}

As for the Lorenz-63, %system, 
for the ECG data we observed a strong performance gain for just $P=2$ PWL units, see Fig. \ref{fig:performance_relu}. Indeed, Fig. \ref{fig:ECG_minimal}\textbf{a} confirms that the complex activity pattern and positive %maximum 
max. Lyapunov exponent ($\lambda_{\text{max}}=1.96\;s^{-1}$, ground truth: $\lambda^{\text{true}}_{\text{max}}=2.19\;s^{-1}$ \cite{hess_generalized_2023}) of the ECG time series could be achieved with $P=2$ in only $3$ linear subregions.
These subregions corresponded to distinct parts of the ECG activity: ramping-up phases (light blue, node $\#1$), declining activity (dark blue, node $\#2$), represented by two unstable spirals with shifted phases (Fig. \ref{fig:ecg_linear_activities}), and the Q wave (medium blue, node $\#3$).
The activity in the third region ($\sigma_{max}\approx1.34$, other two: $\sigma_{max}\approx 1.02$, Fig. \ref{fig:ecg_linear_activities}) caused a strong inhibition in the second PWL unit (Fig. \ref{fig:ECG_minimal}\textbf{b}) that captures the critical transition initiated by the Q wave. The Q wave triggers the depolarization of the interventricular septum during the QRS complex \cite{sodi-pallares_activation_1951}, indicating that this latent depolarization process is captured by the model. These results suggest that the AL-RNN cannot only learn dynamically but also biologically interpretable latent representations.
The core aspects of this representation were furthermore consistent across successful reconstructions (Fig. \ref{fig:ECG_minimal_further}). The symbolic sequences corresponding to the graph in Fig. \ref{fig:ECG_minimal} reveal the nearly periodic yet chaotic dynamics of the ECG (Fig. \ref{fig:ECG_symbolic}).\footnote{In general, we also found that quantities computed from symbolic sequences like the topological entropy correlated highly with the maximum Lyapunov exponent, Fig. \ref{fig:entropy_lyaps}.}

For the short ($T=360$) fMRI time series, $P=3$ often resulted in reconstructions matching the complex activity patterns reasonably well (cf. Fig. \ref{fig:performance_relu}, right). Fig. \ref{fig:fMRI_freely_generated} illustrates results for an example subject using $8$ linear subregions. The second most visited subregion implemented an unstable spiral, while the most visited region had a stable \textit{virtual} fixed point also located in the second subregion. These two regions covered over $50\%$ of the data and were strongly connected (Fig. \ref{fig:fMRI_freely_generated}\textbf{b}). This balance between stable and unstable activity %highlights a clear 
suggests a mechanism through which the network implements chaotic dynamics, with the stable virtual fixed point pulling activity into the second subregion from which it then diverges again (Fig. \ref{fig:fMRI_freely_generated}\textbf{d}).

\begin{figure*}[!htb]
    \centering
	\includegraphics[width=0.8\linewidth]{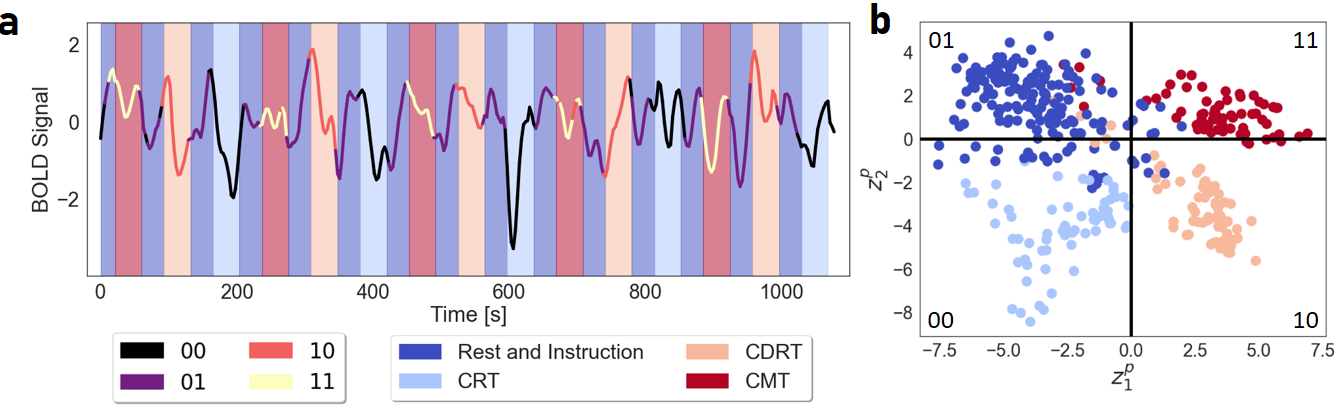}
	\caption{Reconstructions from human fMRI data using an AL-RNN with $M=100$ total units and $P=2$ PWL units. \textbf{a}: Mean generated BOLD activity color-coded according to the linear subregion. Background color shadings indicate the task stage. \textbf{b}: Generated activity (trajectory points) in the latent space of PWL units with color-coding indicating task stage as in \textbf{a}. %The quadrants encoding the four linear subregions closely align with the different task stages.
 }% \textbf{c}: Generated AL-RNN activity for $10$ example units when replacing the readout neurons every $\tau=7$ time steps.
	\label{fig:fMRI_linear_regions}
\end{figure*}

\paragraph{Task stages align with linear subregions}

To integrate the fMRI signal with cognitive (task-phase) information, in addition to the linear decoder for the BOLD signal, we coupled the PWL units $\bm{z}^p_t$ to a categorical decoder model (eq. \ref{eq:supp:cat_obs}) which predicts the three cognitive task stages and the 'Rest and Instruction' period, as in \cite{kramer22a, brenner_integrating_2024}. Using only $P = 2$ PWL units on the fMRI data made it challenging to capture the complex, chaotic long-term activity patterns in the freely generated activity of the AL-RNN. However, the dynamics were still well-approximated locally (Fig. \ref{fig:fMRI_generated_linear_tf}). To maintain temporal alignment with the task stages when sampling from the AL-RNN, the AL-RNN's readout units were reset to the observations every $7$ time steps (Fig. \ref{fig:fMRI_linear_regions}\textbf{a}). Fig. \ref{fig:fMRI_linear_regions}\textbf{a}-\textbf{b} show that in this setup, the four linear subregions of the AL-RNN often closely aligned with the different task stages of the experimental time series. Similar results were obtained across different subjects, with an average classification accuracy of $p = 0.78 \pm 0.05$ (mean $\pm$ SEM) (see Appx. \ref{alignment_score} for details). While the categorical decoder aids in separating latent states according to task stage, there is nothing that would bias this separation to align with the linear subregions. The observed alignment therefore suggests that the AL-RNN learns to leverage distinct linear dynamics in each subregion to represent differences across cognitive tasks. Furthermore, as shown in Fig. \ref{fig:fMRI_weights}, the network weights of the PWL units were significantly larger than those of other units, indicating their critical role in modulating the dynamics and representing task-related variations in brain activity. 
This approach also demonstrates how local context-aligned linear approximations can be achieved using the AL-RNN, which is useful in areas such as model-predictive control \cite{muske_model_1993, daafouz_stability_2002, linderman_recurrent_2016}.

\section{Conclusion}
Here we introduced a novel variant of a PLRNN, the AL-RNN, which learns to represent nonlinear DS with as few PWL nonlinearities as possible. Despite its simple design and the minimal hyperparameter tuning required, the AL-RNN robustly and automatically identifies highly interpretable, topologically minimal representations of complex nonlinear DS, reproducing known minimal PWL forms of chaotic attractors \cite{amaral_piecewise_2006}. Such minimal PWL forms that allow for an interpretable symbolic and graph-theoretical representation were discovered even from challenging physiological and neuroscientific data. They also profoundly ease subsequent model analysis. For instance, with only a few linear subregions to consider, the search for fixed points or cycles becomes very fast and efficient \citep{eisenmann2023bifurcations}.

\paragraph{Limitations} 
While this seems promising, how to determine whether a topologically minimal and valid reconstruction from empirical data has truly been achieved remains an open topic. Performance curves as in Fig. \ref{fig:performance_relu} or Fig. \ref{fig:regularization} give an indication of how many PWL units may be required to yield an optimal minimal representation, but whether there is a more principled way of automatically inferring the optimal number $P$ of PWL nonlinearities from data may be an interesting future direction. Finally, the current finding that even for empirical ECG and fMRI data a few linear subregions ($\leq 8$) suffice for faithful reconstructions is encouraging. Whether this more generally will be the case in empirical scenarios is another interesting and open question. Not all types of (empirically observed) dynamical systems may easily allow for such topologically minimal representations.

All code created is available at \url{https://github.com/DurstewitzLab/ALRNN-DSR}.

\section*{Acknowledgements}
This work was funded %by the European Union’s Horizon 2020 Programme under grant agreement 945263 (IMMERSE), 
by the Federal Ministry of Science, Education, and Culture (MWK) of the state of Baden-Württemberg within the AI Health Innovation Cluster Initiative, by the German Research Foundation (DFG) within Germany’s Excellence Strategy EXC 2181/1 – 390900948 (STRUCTURES), and through DFG individual grant Du 354/15-1 to DD. ZM was funded by the Federal Ministry of Education and Research (BMBF) through project OIDLITDSM, 01IS24061.

\bibliography{bibliography.bib}

\begin{thebibliography}{109}
\providecommand{\natexlab}[1]{#1}
\providecommand{\url}[1]{\texttt{#1}}
\expandafter\ifx\csname urlstyle\endcsname\relax
  \providecommand{\doi}[1]{doi: #1}\else
  \providecommand{\doi}{doi: \begingroup \urlstyle{rm}\Url}\fi

\bibitem[Ackerson and Fu(1970)]{ackerson_state_1970}
G.~Ackerson and K.~Fu.
\newblock On state estimation in switching environments.
\newblock \emph{IEEE Transactions on Automatic Control}, 15\penalty0 (1):\penalty0 10--17, February 1970.
\newblock ISSN 1558-2523.
\newblock \doi{10.1109/TAC.1970.1099359}.
\newblock URL \url{https://ieeexplore.ieee.org/document/1099359}.
\newblock Conference Name: IEEE Transactions on Automatic Control.

\bibitem[Alameda-Pineda et~al.(2022)Alameda-Pineda, Drouard, and Horaud]{alameda-pineda_variational_2022}
Xavier Alameda-Pineda, Vincent Drouard, and Radu~Patrice Horaud.
\newblock Variational {Inference} and {Learning} of {Piecewise} {Linear} {Dynamical} {Systems}.
\newblock \emph{IEEE Transactions on Neural Networks and Learning Systems}, 33\penalty0 (8):\penalty0 3753--3764, August 2022.
\newblock ISSN 2162-2388.
\newblock \doi{10.1109/TNNLS.2021.3054407}.
\newblock URL \url{https://ieeexplore.ieee.org/document/9353398}.
\newblock Conference Name: IEEE Transactions on Neural Networks and Learning Systems.

\bibitem[Alligood et~al.(1996)Alligood, Sauer, and Yorke]{alligood_chaos_1996}
Kathleen~T. Alligood, Tim~D. Sauer, and James~A. Yorke.
\newblock \emph{Chaos: An Introduction to Dynamical Systems}.
\newblock Textbooks in Mathematical Sciences. Springer, 1996.
\newblock ISBN 978-0-387-94677-1 978-0-387-22492-3.
\newblock \doi{10.1007/b97589}.

\bibitem[Alvarez et~al.(2020)Alvarez, Roşca, and Fălcuţescu]{alvarez_dynode_2020}
Victor M.~Martinez Alvarez, Rareş Roşca, and Cristian~G. Fălcuţescu.
\newblock {DyNODE}: {Neural} {Ordinary} {Differential} {Equations} for {Dynamics} {Modeling} in {Continuous} {Control}, September 2020.

\bibitem[Amaral et~al.(2006)Amaral, Letellier, and Aguirre]{amaral_piecewise_2006}
Gleison F.~V. Amaral, Christophe Letellier, and Luis~Antonio Aguirre.
\newblock Piecewise affine models of chaotic attractors: the {Rossler} and {Lorenz} systems.
\newblock \emph{Chaos (Woodbury, N.Y.)}, 16\penalty0 (1):\penalty0 013115, March 2006.
\newblock ISSN 1054-1500.
\newblock \doi{10.1063/1.2149527}.

\bibitem[Azencot et~al.(2020)Azencot, Erichson, Lin, and Mahoney]{azencot_forecasting_2020}
Omri Azencot, N.~Benjamin Erichson, Vanessa Lin, and Michael~W. Mahoney.
\newblock Forecasting {Sequential} {Data} using {Consistent} {Koopman} {Autoencoders}.
\newblock In \emph{Proceedings of the 37th {International} {Conference} on {Machine} {Learning}}, 2020.
\newblock URL \url{http://arxiv.org/abs/2003.02236}.

\bibitem[Belkin and Niyogi(2001)]{belkin_laplacian_2001}
Mikhail Belkin and Partha Niyogi.
\newblock Laplacian {Eigenmaps} and {Spectral} {Techniques} for {Embedding} and {Clustering}.
\newblock In \emph{Advances in {Neural} {Information} {Processing} {Systems}}, volume~14. MIT Press, 2001.
\newblock URL \url{https://proceedings.neurips.cc/paper_files/paper/2001/hash/f106b7f99d2cb30c3db1c3cc0fde9ccb-Abstract.html}.

\bibitem[Bemporad et~al.(2000)Bemporad, Borrelli, and Morari]{bemporad2000piecewise}
Alberto Bemporad, Francesco Borrelli, and Manfred Morari.
\newblock Piecewise linear optimal controllers for hybrid systems.
\newblock In \emph{Proceedings of the 2000 American Control Conference. ACC (IEEE Cat. No. 00CH36334)}, volume~2, pages 1190--1194. IEEE, 2000.

\bibitem[Bengio et~al.(1994)Bengio, Simard, and Frasconi]{bengio_learning_1994}
Y.~Bengio, P.~Simard, and P.~Frasconi.
\newblock Learning long-term dependencies with gradient descent is difficult.
\newblock \emph{{IEEE} transactions on neural networks}, 5\penalty0 (2):\penalty0 157--166, 1994.
\newblock ISSN 1045-9227.
\newblock \doi{10.1109/72.279181}.

\bibitem[Bernardo et~al.(1999)Bernardo, Feigin, Hogan, and Homer]{bernardo_local_1999}
Mario~Di Bernardo, Mark~I. Feigin, Stephen~J. Hogan, and Martin~E. Homer.
\newblock Local {Analysis} of {C}-{Bifurcations} in n-{Dimensional} {Piecewise}-{Smooth} {Dynamical} {Systems}.
\newblock \emph{Chaos, Solitons and Fractals: the interdisciplinary journal of Nonlinear Science, and Nonequilibrium and Complex Phenomena}, 11\penalty0 (10):\penalty0 1881--1908, 1999.
\newblock ISSN 0960-0779.
\newblock URL \url{https://www.infona.pl//resource/bwmeta1.element.elsevier-b61cfd87-9650-310f-bd8f-4bd7e7174946}.

\bibitem[Brenner et~al.(2022)Brenner, Hess, Mikhaeil, Bereska, Monfared, Kuo, and Durstewitz]{brenner_tractable_2022}
Manuel Brenner, Florian Hess, Jonas~M. Mikhaeil, Leonard~F. Bereska, Zahra Monfared, Po-Chen Kuo, and Daniel Durstewitz.
\newblock Tractable {Dendritic} {RNNs} for {Reconstructing} {Nonlinear} {Dynamical} {Systems}.
\newblock In \emph{Proceedings of the 39th {International} {Conference} on {Machine} {Learning}}, pages 2292--2320. PMLR, June 2022.
\newblock URL \url{https://proceedings.mlr.press/v162/brenner22a.html}.
\newblock ISSN: 2640-3498.

\bibitem[Brenner et~al.(2024)Brenner, Hess, Koppe, and Durstewitz]{brenner_integrating_2024}
Manuel Brenner, Florian Hess, Georgia Koppe, and Daniel Durstewitz.
\newblock Integrating {Multimodal} {Data} for {Joint} {Generative} {Modeling} of {Complex} {Dynamics}.
\newblock In \emph{Proceedings of the 41st {International} {Conference} on {Machine} {Learning}}, pages 4482--4516. PMLR, July 2024.
\newblock URL \url{https://proceedings.mlr.press/v235/brenner24a.html}.
\newblock ISSN: 2640-3498.

\bibitem[Brunton and Kutz(2019)]{brunton_data-driven_2019}
Steven~L Brunton and J~Nathan Kutz.
\newblock \emph{Data-driven science and engineering: {Machine} learning, dynamical systems, and control}.
\newblock Cambridge University Press, 2019.

\bibitem[Brunton et~al.(2016)Brunton, Proctor, and Kutz]{brunton_discovering_2016}
Steven~L. Brunton, Joshua~L. Proctor, and J.~Nathan Kutz.
\newblock Discovering governing equations from data by sparse identification of nonlinear dynamical systems.
\newblock \emph{Proceedings of the National Academy of Sciences USA}, 113\penalty0 (15):\penalty0 3932--3937, 2016.
\newblock ISSN 0027-8424.
\newblock \doi{10.1073/pnas.1517384113}.

\bibitem[Brunton et~al.(2021)Brunton, Budišić, Kaiser, and Kutz]{brunton_modern_2021}
Steven~L. Brunton, Marko Budišić, Eurika Kaiser, and J.~Nathan Kutz.
\newblock Modern {Koopman} {Theory} for {Dynamical} {Systems}, October 2021.
\newblock arXiv:2102.12086 [cs, eess, math].

\bibitem[Buzsaki(2006)]{buzsaki_rhythms_2006}
Gyorgy Buzsaki.
\newblock \emph{Rhythms of the {Brain}}.
\newblock Oxford University Press, August 2006.
\newblock ISBN 978-0-19-804125-2.
\newblock Google-Books-ID: ldz58irprjYC.

\bibitem[Carmona et~al.(2002)Carmona, Freire, Ponce, and Torres]{carmona2002simplifying}
Victoriano Carmona, Emilio Freire, Enrique Ponce, and Francisco Torres.
\newblock On simplifying and classifying piecewise-linear systems.
\newblock \emph{IEEE Transactions on Circuits and Systems I: Fundamental Theory and Applications}, 49\penalty0 (5):\penalty0 609--620, 2002.

\bibitem[Casdagli(1989)]{casdagli_nonlinear_1989}
Martin Casdagli.
\newblock Nonlinear prediction of chaotic time series.
\newblock \emph{Physica D: Nonlinear Phenomena}, 35\penalty0 (3):\penalty0 335--356, May 1989.
\newblock ISSN 0167-2789.
\newblock \doi{10.1016/0167-2789(89)90074-2}.

\bibitem[Cestnik and Abel(2019)]{cestnik_inferring_2019}
Rok Cestnik and Markus Abel.
\newblock Inferring the dynamics of oscillatory systems using recurrent neural networks.
\newblock \emph{Chaos: An Interdisciplinary Journal of Nonlinear Science}, 29\penalty0 (6):\penalty0 063128, June 2019.
\newblock ISSN 1054-1500.
\newblock \doi{10.1063/1.5096918}.
\newblock URL \url{https://doi.org/10.1063/1.5096918}.

\bibitem[Chen et~al.(2018)Chen, Rubanova, Bettencourt, and Duvenaud]{chen_neural_2018}
Ricky T.~Q. Chen, Yulia Rubanova, Jesse Bettencourt, and David Duvenaud.
\newblock Neural {Ordinary} {Differential} {Equations}.
\newblock In \emph{Advances in {Neural} {Information} {Processing} {Systems} 31}, 2018.
\newblock URL \url{http://arxiv.org/abs/1806.07366}.

\bibitem[Cortiella et~al.(2021)Cortiella, Park, and Doostan]{cortiella_sparse_2021}
Alexandre Cortiella, Kwang-Chun Park, and Alireza Doostan.
\newblock Sparse identification of nonlinear dynamical systems via reweighted l1-regularized least squares.
\newblock \emph{Computer Methods in Applied Mechanics and Engineering}, 376:\penalty0 113620, April 2021.
\newblock ISSN 0045-7825.
\newblock \doi{10.1016/j.cma.2020.113620}.

\bibitem[Costa et~al.(2019)Costa, Ahamed, and Stephens]{costa_adaptive_2019}
Antonio~C. Costa, Tosif Ahamed, and Greg~J. Stephens.
\newblock Adaptive, locally linear models of complex dynamics.
\newblock \emph{Proceedings of the National Academy of Sciences}, 116\penalty0 (5):\penalty0 1501--1510, January 2019.
\newblock \doi{10.1073/pnas.1813476116}.
\newblock Publisher: Proceedings of the National Academy of Sciences.

\bibitem[Daafouz et~al.(2002)Daafouz, Riedinger, and Iung]{daafouz_stability_2002}
J.~Daafouz, P.~Riedinger, and C.~Iung.
\newblock Stability analysis and control synthesis for switched systems: a switched {Lyapunov} function approach.
\newblock \emph{IEEE Transactions on Automatic Control}, 47\penalty0 (11):\penalty0 1883--1887, November 2002.
\newblock ISSN 1558-2523.
\newblock \doi{10.1109/TAC.2002.804474}.
\newblock URL \url{https://ieeexplore.ieee.org/document/1047016}.
\newblock Conference Name: IEEE Transactions on Automatic Control.

\bibitem[De~Feo and Storace(2007)]{de_feo_piecewise-linear_2007}
Oscar De~Feo and Marco Storace.
\newblock Piecewise-{Linear} {Identification} of {Nonlinear} {Dynamical} {Systems} in {View} of {Their} {Circuit} {Implementations}.
\newblock \emph{IEEE Transactions on Circuits and Systems I: Regular Papers}, 54\penalty0 (7):\penalty0 1542--1554, July 2007.
\newblock ISSN 1558-0806.
\newblock \doi{10.1109/TCSI.2007.899613}.
\newblock URL \url{https://ieeexplore.ieee.org/document/4268404}.
\newblock Conference Name: IEEE Transactions on Circuits and Systems I: Regular Papers.

\bibitem[Durstewitz(2017)]{durstewitz_state_2017}
Daniel Durstewitz.
\newblock A state space approach for piecewise-linear recurrent neural networks for identifying computational dynamics from neural measurements.
\newblock \emph{PLoS Comput. Biol.}, 13\penalty0 (6):\penalty0 e1005542, 2017.
\newblock ISSN 1553-7358.
\newblock \doi{10.1371/journal.pcbi.1005542}.

\bibitem[Eisenmann et~al.(2024)Eisenmann, Monfared, Göring, and Durstewitz]{eisenmann2023bifurcations}
Lukas Eisenmann, Zahra Monfared, Niclas Göring, and Daniel Durstewitz.
\newblock Bifurcations and loss jumps in {RNN} training.
\newblock \emph{Advances in Neural Information Processing Systems}, 36, 2024.

\bibitem[Feigin(1995)]{feigin_increasingly_1995}
Mark~I Feigin.
\newblock The increasingly complex structure of the bifurcation tree of a piecewise-smooth system.
\newblock \emph{Journal of Applied Mathematics and Mechanics}, 59\penalty0 (6):\penalty0 853--863, 1995.
\newblock ISSN 0021-8928.
\newblock \doi{10.1016/0021-8928(95)00118-2}.
\newblock URL \url{https://www.sciencedirect.com/science/article/pii/0021892895001182}.

\bibitem[Fox et~al.(2008)Fox, Sudderth, Jordan, and Willsky]{fox_nonparametric_2008}
Emily Fox, Erik Sudderth, Michael Jordan, and Alan Willsky.
\newblock Nonparametric {Bayesian} {Learning} of {Switching} {Linear} {Dynamical} {Systems}.
\newblock In \emph{Advances in {Neural} {Information} {Processing} {Systems}}, volume~21. Curran Associates, Inc., 2008.
\newblock URL \url{https://papers.nips.cc/paper_files/paper/2008/hash/950a4152c2b4aa3ad78bdd6b366cc179-Abstract.html}.

\bibitem[Funahashi(1989)]{funahashi_approximate_1989}
Ken-ichi Funahashi.
\newblock On the approximate realization of continuous mappings by neural networks.
\newblock \emph{Neural Networks}, 1989.
\newblock \doi{10.1016/0893-6080(89)90003-8}.

\bibitem[Geneva and Zabaras(2022)]{geneva_transformers_2022}
Nicholas Geneva and Nicholas Zabaras.
\newblock Transformers for modeling physical systems.
\newblock \emph{Neural Networks}, 146:\penalty0 272--289, February 2022.
\newblock ISSN 0893-6080.
\newblock \doi{10.1016/j.neunet.2021.11.022}.

\bibitem[Ghahramani and Hinton(2000)]{ghahramani_variational_2000}
Z.~Ghahramani and G.~E. Hinton.
\newblock Variational learning for switching state-space models.
\newblock \emph{Neural Computation}, 12\penalty0 (4):\penalty0 831--864, April 2000.
\newblock ISSN 0899-7667.
\newblock \doi{10.1162/089976600300015619}.

\bibitem[Guckenheimer and Holmes(1983)]{guckenheimer_nonlinear_1983}
John Guckenheimer and Philip Holmes.
\newblock \emph{Nonlinear {Oscillations}, {Dynamical} {Systems}, and {Bifurcations} of {Vector} {Fields}}, volume~42 of \emph{Applied {Mathematical} {Sciences}}.
\newblock Springer, New York, NY, 1983.
\newblock ISBN 978-1-4612-7020-1 978-1-4612-1140-2.
\newblock \doi{10.1007/978-1-4612-1140-2}.
\newblock URL \url{http://link.springer.com/10.1007/978-1-4612-1140-2}.

\bibitem[Hahnloser and Seung(2000)]{hahnloser_permitted_2000}
Richard Hahnloser and H.~Sebastian Seung.
\newblock Permitted and {Forbidden} {Sets} in {Symmetric} {Threshold}-{Linear} {Networks}.
\newblock In \emph{Advances in {Neural} {Information} {Processing} {Systems}}, volume~13. MIT Press, 2000.

\bibitem[Hanson and Raginsky(2020)]{pmlr-v120-hanson20a}
Joshua Hanson and Maxim Raginsky.
\newblock Universal simulation of stable dynamical systems by recurrent neural nets.
\newblock In \emph{Proceedings of the 2nd Conference on Learning for Dynamics and Control}, volume 120 of \emph{Proceedings of Machine Learning Research}, pages 384--392. PMLR, 10--11 Jun 2020.
\newblock URL \url{https://proceedings.mlr.press/v120/hanson20a.html}.

\bibitem[Hao and Zheng(1998)]{hao1998applied}
Bailin Hao and Weimou Zheng.
\newblock \emph{Applied symbolic dynamics and chaos}.
\newblock World Scientific, 1998.

\bibitem[Hartman(1960)]{hartman_lemma_1960}
Philip Hartman.
\newblock A lemma in the theory of structural stability of differential equations.
\newblock \emph{Proceedings of the American Mathematical Society}, 11\penalty0 (4):\penalty0 610--620, 1960.
\newblock ISSN 0002-9939, 1088-6826.
\newblock \doi{10.1090/S0002-9939-1960-0121542-7}.
\newblock URL \url{https://www.ams.org/proc/1960-011-04/S0002-9939-1960-0121542-7/}.

\bibitem[Heim et~al.(2019)Heim, Šmídl, and Pevný]{heim_rodent_2019}
Niklas Heim, Václav Šmídl, and Tomáš Pevný.
\newblock Rodent: {Relevance} determination in differential equations.
\newblock \emph{arXiv preprint arXiv:1912.00656}, 2019.
\newblock URL \url{http://arxiv.org/abs/1912.00656}.

\bibitem[Hemmer et~al.(2024)Hemmer, Brenner, Hess, and Durstewitz]{hemmer_optimal_2024}
Christoph~Jürgen Hemmer, Manuel Brenner, Florian Hess, and Daniel Durstewitz.
\newblock Optimal {Recurrent} {Network} {Topologies} for {Dynamical} {Systems} {Reconstruction}.
\newblock In \emph{Proceedings of the 41st {International} {Conference} on {Machine} {Learning}}, pages 18174--18204. PMLR, July 2024.
\newblock URL \url{https://proceedings.mlr.press/v235/hemmer24a.html}.
\newblock ISSN: 2640-3498.

\bibitem[Hess et~al.(2023)Hess, Monfared, Brenner, and Durstewitz]{hess_generalized_2023}
Florian Hess, Zahra Monfared, Manuel Brenner, and Daniel Durstewitz.
\newblock Generalized {Teacher} {Forcing} for {Learning} {Chaotic} {Dynamics}.
\newblock In \emph{Proceedings of the 40th {International} {Conference} on {Machine} {Learning}}, pages 13017--13049. PMLR, July 2023.
\newblock URL \url{https://proceedings.mlr.press/v202/hess23a.html}.
\newblock ISSN: 2640-3498.

\bibitem[Hogan et~al.(2007)Hogan, Higham, and Griffin]{hogan_dynamics_2007}
Stephen~J. Hogan, L.~Higham, and T.~C.~L. Griffin.
\newblock Dynamics of a piecewise linear map with a gap.
\newblock \emph{Proceedings of the Royal Society A: Mathematical, Physical and Engineering Sciences}, 463\penalty0 (2077):\penalty0 49--65, 2007.
\newblock \doi{10.1098/rspa.2006.1735}.
\newblock URL \url{https://royalsocietypublishing.org/doi/abs/10.1098/rspa.2006.1735}.

\bibitem[Ives and Dakos(2012)]{ives_detecting_2012}
Anthony~R. Ives and Vasilis Dakos.
\newblock Detecting dynamical changes in nonlinear time series using locally linear state-space models.
\newblock \emph{Ecosphere}, 3\penalty0 (6):\penalty0 art58, 2012.
\newblock ISSN 2150-8925.
\newblock \doi{10.1890/ES11-00347.1}.
\newblock \_eprint: https://onlinelibrary.wiley.com/doi/pdf/10.1890/ES11-00347.1.

\bibitem[Jain and Banerjee(2003)]{jain_border-collision_2003}
Parag Jain and Soumitro Banerjee.
\newblock Border-collision bifurcations in one-dimensional discontinuous maps.
\newblock \emph{Int. J. Bifurcation Chaos}, 13\penalty0 (11):\penalty0 3341--3351, 2003.
\newblock ISSN 0218-1274.
\newblock \doi{10.1142/S0218127403008533}.
\newblock URL \url{https://www.worldscientific.com/doi/abs/10.1142/S0218127403008533}.

\bibitem[Juloski et~al.(2005)Juloski, Weiland, and Heemels]{juloski_bayesian_2005}
A.L. Juloski, S.~Weiland, and W.P.M.H. Heemels.
\newblock A {Bayesian} approach to identification of hybrid systems.
\newblock \emph{IEEE Transactions on Automatic Control}, 50\penalty0 (10):\penalty0 1520--1533, October 2005.
\newblock ISSN 1558-2523.
\newblock \doi{10.1109/TAC.2005.856649}.
\newblock URL \url{https://ieeexplore.ieee.org/document/1516255}.
\newblock Conference Name: IEEE Transactions on Automatic Control.

\bibitem[Kaheman et~al.(2022)Kaheman, Brunton, and Kutz]{kaheman_automatic_2022}
Kadierdan Kaheman, Steven~L. Brunton, and J.~Nathan Kutz.
\newblock Automatic differentiation to simultaneously identify nonlinear dynamics and extract noise probability distributions from data.
\newblock \emph{Machine Learning: Science and Technology}, 3\penalty0 (1):\penalty0 015031, March 2022.
\newblock ISSN 2632-2153.
\newblock \doi{10.1088/2632-2153/ac567a}.
\newblock Publisher: IOP Publishing.

\bibitem[Kaiser et~al.(2018)Kaiser, Kutz, and Brunton]{kaiser_sparse_2018}
E.~Kaiser, J.~N. Kutz, and S.~L. Brunton.
\newblock Sparse identification of nonlinear dynamics for model predictive control in the low-data limit.
\newblock \emph{Proceedings of the Royal Society A: Mathematical, Physical and Engineering Sciences}, 474\penalty0 (2219):\penalty0 20180335, November 2018.
\newblock \doi{10.1098/rspa.2018.0335}.
\newblock Publisher: Royal Society.

\bibitem[Karlsson and Svanström(2019)]{karlsson_modelling_2019}
Daniel Karlsson and Olle Svanström.
\newblock Modelling {Dynamical} {Systems} {Using} {Neural} {Ordinary} {Differential} {Equations}.
\newblock 2019.
\newblock URL \url{https://hdl.handle.net/20.500.12380/256887}.

\bibitem[Kimura and Nakano(1998)]{kimura1998learning}
Masahiro Kimura and Ryohei Nakano.
\newblock Learning dynamical systems by recurrent neural networks from orbits.
\newblock \emph{Neural Networks}, 11\penalty0 (9):\penalty0 1589--1599, 1998.

\bibitem[Ko et~al.(2023)Ko, Koh, Park, and Jhe]{ko_homotopy-based_2023}
Joon-Hyuk Ko, Hankyul Koh, Nojun Park, and Wonho Jhe.
\newblock Homotopy-based training of {NeuralODEs} for accurate dynamics discovery, May 2023.
\newblock URL \url{http://arxiv.org/abs/2210.01407}.
\newblock arXiv:2210.01407 [physics].

\bibitem[Koopman and Neumann(1932)]{koopman_dynamical_1932}
B.~O. Koopman and J.~v. Neumann.
\newblock Dynamical {Systems} of {Continuous} {Spectra}.
\newblock \emph{Proceedings of the National Academy of Sciences}, 18\penalty0 (3):\penalty0 255--263, March 1932.
\newblock \doi{10.1073/pnas.18.3.255}.
\newblock Publisher: Proceedings of the National Academy of Sciences.

\bibitem[Koppe et~al.(2014)Koppe, Gruppe, Sammer, Gallhofer, Kirsch, and Lis]{koppe_temporal_2014}
Georgia Koppe, Harald Gruppe, Gebhard Sammer, Bernd Gallhofer, Peter Kirsch, and Stefanie Lis.
\newblock Temporal unpredictability of a stimulus sequence affects brain activation differently depending on cognitive task demands.
\newblock \emph{{NeuroImage}}, 101:\penalty0 236--244, 2014.
\newblock ISSN 1095-9572.
\newblock \doi{10.1016/j.neuroimage.2014.07.008}.

\bibitem[Koppe et~al.(2019)Koppe, Toutounji, Kirsch, Lis, and Durstewitz]{koppe_identifying_2019}
Georgia Koppe, Hazem Toutounji, Peter Kirsch, Stefanie Lis, and Daniel Durstewitz.
\newblock Identifying nonlinear dynamical systems via generative recurrent neural networks with applications to {fMRI}.
\newblock \emph{PLOS Computational Biology}, 15\penalty0 (8):\penalty0 e1007263, 2019.
\newblock ISSN 1553-7358.
\newblock \doi{10.1371/journal.pcbi.1007263}.

\bibitem[Kramer et~al.(2022)Kramer, Bommer, Tombolini, Koppe, and Durstewitz]{kramer22a}
Daniel Kramer, Philine~L Bommer, Carlo Tombolini, Georgia Koppe, and Daniel Durstewitz.
\newblock Reconstructing nonlinear dynamical systems from multi-modal time series.
\newblock In \emph{Proceedings of the 39th International Conference on Machine Learning}, volume 162 of \emph{Proceedings of Machine Learning Research}, pages 11613--11633. PMLR, 17--23 Jul 2022.
\newblock URL \url{https://proceedings.mlr.press/v162/kramer22a.html}.

\bibitem[La~Cava et~al.(2021)La~Cava, Orzechowski, Burlacu, de~França, Virgolin, Jin, Kommenda, and Moore]{la_cava_contemporary_2021}
William La~Cava, Patryk Orzechowski, Bogdan Burlacu, Fabrício~Olivetti de~França, Marco Virgolin, Ying Jin, Michael Kommenda, and Jason~H. Moore.
\newblock Contemporary {Symbolic} {Regression} {Methods} and their {Relative} {Performance}, July 2021.
\newblock URL \url{http://arxiv.org/abs/2107.14351}.
\newblock arXiv:2107.14351 [cs].

\bibitem[Lind and Marcus(1995)]{lind_introduction_1995}
Douglas Lind and Brian Marcus.
\newblock \emph{An {Introduction} to {Symbolic} {Dynamics} and {Coding}}.
\newblock Cambridge University Press, Cambridge, 1995.
\newblock \doi{10.1017/CBO9780511626302}.
\newblock URL \url{https://www.cambridge.org/core/books/an-introduction-to-symbolic-dynamics-and-coding/331DF5E9DC464B340DED80431BD6D186}.

\bibitem[Lind and Marcus(2021)]{Lind_Marcus_2021}
Douglas Lind and Brian Marcus.
\newblock \emph{An Introduction to Symbolic Dynamics and Coding}.
\newblock Cambridge Mathematical Library. Cambridge University Press, 2 edition, 2021.

\bibitem[Linderman et~al.(2017)Linderman, Johnson, Miller, Adams, Blei, and Paninski]{linderman_bayesian_2017}
Scott Linderman, Matthew Johnson, Andrew Miller, Ryan Adams, David Blei, and Liam Paninski.
\newblock Bayesian {Learning} and {Inference} in {Recurrent} {Switching} {Linear} {Dynamical} {Systems}.
\newblock In \emph{Proceedings of the 20th {International} {Conference} on {Artificial} {Intelligence} and {Statistics}}, pages 914--922. PMLR, April 2017.
\newblock URL \url{https://proceedings.mlr.press/v54/linderman17a.html}.
\newblock ISSN: 2640-3498.

\bibitem[Linderman and Johnson(2017)]{linderman_structure-exploiting_2017}
Scott~W. Linderman and Matthew~J. Johnson.
\newblock Structure-{Exploiting} variational inference for recurrent switching linear dynamical systems.
\newblock In \emph{2017 {IEEE} 7th {International} {Workshop} on {Computational} {Advances} in {Multi}-{Sensor} {Adaptive} {Processing} ({CAMSAP})}, pages 1--5, December 2017.
\newblock \doi{10.1109/CAMSAP.2017.8313132}.
\newblock URL \url{https://ieeexplore.ieee.org/document/8313132}.

\bibitem[Linderman et~al.(2016)Linderman, Miller, Adams, Blei, Paninski, and Johnson]{linderman_recurrent_2016}
Scott~W. Linderman, Andrew~C. Miller, Ryan~P. Adams, David~M. Blei, Liam Paninski, and Matthew~J. Johnson.
\newblock Recurrent switching linear dynamical systems.
\newblock \emph{arXiv:1610.08466 [stat]}, October 2016.
\newblock URL \url{http://arxiv.org/abs/1610.08466}.
\newblock arXiv: 1610.08466.

\bibitem[Liu et~al.(2020)Liu, Jiang, He, Chen, Liu, Gao, and Han]{Liu2020On}
Liyuan Liu, Haoming Jiang, Pengcheng He, Weizhu Chen, Xiaodong Liu, Jianfeng Gao, and Jiawei Han.
\newblock On the variance of the adaptive learning rate and beyond.
\newblock In \emph{International Conference on Learning Representations}, 2020.
\newblock URL \url{https://openreview.net/forum?id=rkgz2aEKDr}.

\bibitem[Loiseau and Brunton(2018)]{loiseau_constrained_2018}
Jean-Christophe Loiseau and Steven~L. Brunton.
\newblock Constrained sparse {Galerkin} regression.
\newblock \emph{Journal of Fluid Mechanics}, 838:\penalty0 42--67, March 2018.
\newblock ISSN 0022-1120, 1469-7645.
\newblock \doi{10.1017/jfm.2017.823}.
\newblock Publisher: Cambridge University Press.

\bibitem[Lorenz(1963)]{lorenz_deterministic_1963}
Edward~N Lorenz.
\newblock Deterministic nonperiodic flow.
\newblock \emph{Journal of atmospheric sciences}, 20\penalty0 (2):\penalty0 130--141, 1963.

\bibitem[Lorenz(1996)]{lorenz_predictability_1996}
Edward~N Lorenz.
\newblock Predictability: {A} problem partly solved.
\newblock In \emph{Proc. {Seminar} on predictability}, volume~1, 1996.

\bibitem[Lusch et~al.(2018)Lusch, Kutz, and Brunton]{lusch_deep_2018}
Bethany Lusch, J.~Nathan Kutz, and Steven~L. Brunton.
\newblock Deep learning for universal linear embeddings of nonlinear dynamics.
\newblock \emph{Nat Commun}, 9\penalty0 (1):\penalty0 4950, December 2018.
\newblock ISSN 2041-1723.
\newblock \doi{10.1038/s41467-018-07210-0}.
\newblock URL \url{http://arxiv.org/abs/1712.09707}.
\newblock arXiv: 1712.09707.

\bibitem[Makke and Chawla(2024)]{makke_interpretable_2024}
Nour Makke and Sanjay Chawla.
\newblock Interpretable scientific discovery with symbolic regression: a review.
\newblock \emph{Artificial Intelligence Review}, 57\penalty0 (1):\penalty0 2, January 2024.
\newblock ISSN 1573-7462.
\newblock \doi{10.1007/s10462-023-10622-0}.
\newblock URL \url{https://doi.org/10.1007/s10462-023-10622-0}.

\bibitem[Mandelbrot and Hudson(2007)]{mandelbrot_misbehavior_2007}
Benoit Mandelbrot and Richard~L. Hudson.
\newblock \emph{The {Misbehavior} of {Markets}: {A} {Fractal} {View} of {Financial} {Turbulence}}.
\newblock Basic Books, March 2007.
\newblock ISBN 978-0-465-00468-3.
\newblock Google-Books-ID: GMKeUqufPQ0C.

\bibitem[Messenger and Bortz(2021)]{messenger_weak_2021}
Daniel~A. Messenger and David~M. Bortz.
\newblock Weak {SINDy}: {Galerkin}-{Based} {Data}-{Driven} {Model} {Selection}.
\newblock \emph{Multiscale Modeling \& Simulation}, 19\penalty0 (3):\penalty0 1474--1497, January 2021.
\newblock ISSN 1540-3459.
\newblock \doi{10.1137/20M1343166}.
\newblock URL \url{https://epubs.siam.org/doi/10.1137/20M1343166}.
\newblock Publisher: Society for Industrial and Applied Mathematics.

\bibitem[Mikhaeil et~al.(2022)Mikhaeil, Monfared, and Durstewitz]{mikhaeil_difficulty_2022}
Jonas Mikhaeil, Zahra Monfared, and Daniel Durstewitz.
\newblock On the difficulty of learning chaotic dynamics with {RNNs}.
\newblock \emph{Advances in Neural Information Processing Systems}, 35:\penalty0 11297--11312, December 2022.
\newblock URL \url{https://proceedings.neurips.cc/paper_files/paper/2022/hash/495e55f361708bedbab5d81f92048dcd-Abstract-Conference.html}.

\bibitem[Milnor(1985)]{milnor_concept_1985}
John Milnor.
\newblock On the concept of attractor.
\newblock \emph{Communications in Mathematical Physics}, 99\penalty0 (2):\penalty0 177--195, June 1985.
\newblock ISSN 1432-0916.
\newblock \doi{10.1007/BF01212280}.
\newblock URL \url{https://doi.org/10.1007/BF01212280}.

\bibitem[Mischaikow et~al.(1999)Mischaikow, Mrozek, Reiss, and Szymczak]{mischaikow_construction_1999}
K.~Mischaikow, M.~Mrozek, J.~Reiss, and A.~Szymczak.
\newblock Construction of {Symbolic} {Dynamics} from {Experimental} {Time} {Series}.
\newblock \emph{Physical Review Letters}, 82\penalty0 (6):\penalty0 1144--1147, February 1999.
\newblock \doi{10.1103/PhysRevLett.82.1144}.
\newblock URL \url{https://link.aps.org/doi/10.1103/PhysRevLett.82.1144}.
\newblock Publisher: American Physical Society.

\bibitem[Monfared and Durstewitz(2020)]{monfared_existence_2020}
Zahra Monfared and Daniel Durstewitz.
\newblock Existence of n-cycles and border-collision bifurcations in piecewise-linear continuous maps with applications to recurrent neural networks.
\newblock \emph{Nonlinear Dyn}, 101\penalty0 (2):\penalty0 1037--1052, 2020.
\newblock ISSN 1573-269X.
\newblock \doi{10.1007/s11071-020-05841-x}.
\newblock URL \url{https://doi.org/10.1007/s11071-020-05841-x}.

\bibitem[Mumby et~al.(2007)Mumby, Hastings, and Edwards]{ecology1}
Peter~J. Mumby, Alan Hastings, and Helen~J. Edwards.
\newblock Thresholds and the resilience of caribbean coral reefs.
\newblock \emph{Nature}, 450\penalty0 (7166):\penalty0 98--101, 2007.
\newblock \doi{10.1038/nature06252}.
\newblock URL \url{https://doi.org/10.1038/nature06252}.

\bibitem[Muske and Rawlings(1993)]{muske_model_1993}
Kenneth~R. Muske and James~B. Rawlings.
\newblock Model predictive control with linear models.
\newblock \emph{AIChE Journal}, 39\penalty0 (2):\penalty0 262--287, 1993.
\newblock ISSN 1547-5905.
\newblock \doi{10.1002/aic.690390208}.
\newblock URL \url{https://onlinelibrary.wiley.com/doi/abs/10.1002/aic.690390208}.
\newblock \_eprint: https://onlinelibrary.wiley.com/doi/pdf/10.1002/aic.690390208.

\bibitem[Naiman and Azencot(2021)]{naiman_koopman_2021}
Ilan Naiman and Omri Azencot.
\newblock A {Koopman} {Approach} to {Understanding} {Sequence} {Neural} {Models}.
\newblock \emph{arXiv:2102.07824 [cs, math]}, October 2021.
\newblock URL \url{http://arxiv.org/abs/2102.07824}.
\newblock arXiv: 2102.07824.

\bibitem[Osipenko and Campbell(1998)]{osipenko_applied_1998}
George Osipenko and Stephen Campbell.
\newblock Applied symbolic dynamics: attractors and filtrations.
\newblock \emph{Discrete and Continuous Dynamical Systems}, 5\penalty0 (1):\penalty0 43--60, September 1998.
\newblock ISSN 1078-0947.
\newblock \doi{10.3934/dcds.1999.5.43}.
\newblock URL \url{https://www.aimsciences.org/en/article/doi/10.3934/dcds.1999.5.43}.
\newblock Publisher: Discrete and Continuous Dynamical Systems.

\bibitem[Otto and Rowley(2019)]{otto_linearly-recurrent_2019}
Samuel~E. Otto and Clarence~W. Rowley.
\newblock Linearly-{Recurrent} {Autoencoder} {Networks} for {Learning} {Dynamics}, January 2019.
\newblock URL \url{http://arxiv.org/abs/1712.01378}.
\newblock arXiv:1712.01378 [cs, math, stat].

\bibitem[Pathak et~al.(2017)Pathak, Lu, Hunt, Girvan, and Ott]{pathak_using_2017}
Jaideep Pathak, Zhixin Lu, Brian~R. Hunt, Michelle Girvan, and Edward Ott.
\newblock Using {Machine} {Learning} to {Replicate} {Chaotic} {Attractors} and {Calculate} {Lyapunov} {Exponents} from {Data}.
\newblock \emph{Chaos: An Interdisciplinary Journal of Nonlinear Science}, 27\penalty0 (12):\penalty0 121102, December 2017.
\newblock ISSN 1054-1500, 1089-7682.
\newblock \doi{10.1063/1.5010300}.
\newblock URL \url{http://arxiv.org/abs/1710.07313}.
\newblock arXiv: 1710.07313.

\bibitem[Patra(2018)]{patra_multiple_2018}
Mahashweta Patra.
\newblock Multiple {Attractor} {Bifurcation} in {Three}-{Dimensional} {Piecewise} {Linear} {Maps}.
\newblock \emph{Int. J. Bifurcation Chaos}, 28\penalty0 (10):\penalty0 1830032, 2018.
\newblock ISSN 0218-1274.
\newblock \doi{10.1142/S021812741830032X}.
\newblock URL \url{https://www.worldscientific.com/doi/abs/10.1142/S021812741830032X}.

\bibitem[Perko(2001)]{perko_differential_2001}
Lawrence Perko.
\newblock \emph{Differential equations and dynamical systems}.
\newblock Number~7 in Texts in applied mathematics. Springer, New York, 3rd ed edition, 2001.
\newblock ISBN 978-0-387-95116-4.

\bibitem[Platt et~al.(2022)Platt, Penny, Smith, Chen, and Abarbanel]{platt_systematic_2022}
Jason~A. Platt, Stephen~G. Penny, Timothy~A. Smith, Tse-Chun Chen, and Henry D.~I. Abarbanel.
\newblock A {Systematic} {Exploration} of {Reservoir} {Computing} for {Forecasting} {Complex} {Spatiotemporal} {Dynamics}, January 2022.
\newblock URL \url{http://arxiv.org/abs/2201.08910}.
\newblock arXiv:2201.08910 [cs].

\bibitem[Platt et~al.(2023)Platt, Penny, Smith, Chen, and Abarbanel]{platt2023constraining}
Jason~A Platt, Stephen~G Penny, Timothy~A Smith, Tse-Chun Chen, and Henry~DI Abarbanel.
\newblock Constraining chaos: Enforcing dynamical invariants in the training of reservoir computers.
\newblock \emph{Chaos: An Interdisciplinary Journal of Nonlinear Science}, 33\penalty0 (10), 2023.

\bibitem[Rantzer and Johansson(2000)]{rantzer2000piecewise}
Anders Rantzer and Mikael Johansson.
\newblock Piecewise linear quadratic optimal control.
\newblock \emph{IEEE transactions on automatic control}, 45\penalty0 (4):\penalty0 629--637, 2000.

\bibitem[Reiss et~al.(2019)Reiss, Indlekofer, Schmidt, and Van~Laerhoven]{reiss2019deep}
Attila Reiss, Ina Indlekofer, Philip Schmidt, and Kristof Van~Laerhoven.
\newblock Deep ppg: Large-scale heart rate estimation with convolutional neural networks.
\newblock \emph{Sensors}, 19\penalty0 (14):\penalty0 3079, 2019.

\bibitem[Rudin(2019)]{rudin_stop_2019}
Cynthia Rudin.
\newblock Stop explaining black box machine learning models for high stakes decisions and use interpretable models instead.
\newblock \emph{Nature Machine Intelligence}, 1\penalty0 (5):\penalty0 206--215, May 2019.
\newblock ISSN 2522-5839.
\newblock \doi{10.1038/s42256-019-0048-x}.
\newblock URL \url{https://www.nature.com/articles/s42256-019-0048-x}.
\newblock Publisher: Nature Publishing Group.

\bibitem[Rusch et~al.(2022)Rusch, Mishra, Erichson, and Mahoney]{rusch_long_2022}
T~Konstantin Rusch, Siddhartha Mishra, N~Benjamin Erichson, and Michael~W Mahoney.
\newblock Long expressive memory for sequence modeling.
\newblock In \emph{International Conference on Learning Representations}, 2022.

\bibitem[Rössler(1976)]{rossler_equation_1976}
O.~E. Rössler.
\newblock An equation for continuous chaos.
\newblock \emph{Physics Letters A}, 57\penalty0 (5):\penalty0 397--398, 1976.
\newblock ISSN 0375-9601.
\newblock \doi{10.1016/0375-9601(76)90101-8}.
\newblock URL \url{https://www.sciencedirect.com/science/article/pii/0375960176901018}.

\bibitem[Sauer et~al.(1991)Sauer, Yorke, and Casdagli]{sauer_embedology_1991}
Tim Sauer, James~A Yorke, and Martin Casdagli.
\newblock Embedology.
\newblock \emph{Journal of statistical Physics}, 65\penalty0 (3):\penalty0 579--616, 1991.

\bibitem[Schalk et~al.(2000)Schalk, {McFarland}, Hinterberger, Birbaumer, and Wolpaw]{schalk_bci2000_2004}
Gerwin Schalk, Dennis~J. {McFarland}, Thilo Hinterberger, Niels Birbaumer, and Jonathan~R. Wolpaw.
\newblock {BCI}2000: a general-purpose brain-computer interface ({BCI}) system.
\newblock \emph{{IEEE} transactions on bio-medical engineering}, 51\penalty0 (6):\penalty0 1034--1043, 2000.
\newblock ISSN 0018-9294.
\newblock \doi{10.1109/TBME.2004.827072}.

\bibitem[Schmidt et~al.(2021)Schmidt, Koppe, Monfared, Beutelspacher, and Durstewitz]{schmidt_identifying_2021}
Dominik Schmidt, Georgia Koppe, Zahra Monfared, Max Beutelspacher, and Daniel Durstewitz.
\newblock Identifying nonlinear dynamical systems with multiple time scales and long-range dependencies.
\newblock In \emph{Proceedings of the 9th {International} {Conference} on {Learning} {Representations}}, 2021.
\newblock URL \url{http://arxiv.org/abs/1910.03471}.

\bibitem[Sodi-Pallares et~al.(1951)Sodi-Pallares, Rodriguez, Chait, and Zuckermann]{sodi-pallares_activation_1951}
Demetrio Sodi-Pallares, María~Isabel Rodriguez, Leonardo~O. Chait, and Rudolf Zuckermann.
\newblock The activation of the interventricular septum.
\newblock \emph{American Heart Journal}, 41\penalty0 (4):\penalty0 569--608, April 1951.
\newblock ISSN 0002-8703.
\newblock \doi{10.1016/0002-8703(51)90024-5}.
\newblock URL \url{https://www.sciencedirect.com/science/article/pii/0002870351900245}.

\bibitem[Solomon(2015)]{solomon_pde_2015}
Justin Solomon.
\newblock {PDE} {Approaches} to {Graph} {Analysis}, April 2015.
\newblock URL \url{http://arxiv.org/abs/1505.00185}.
\newblock arXiv:1505.00185 [cs, math].

\bibitem[Sontag(1981)]{sontag_nonlinear_1981}
E.~Sontag.
\newblock Nonlinear regulation: {The} piecewise linear approach.
\newblock \emph{IEEE Transactions on Automatic Control}, 26\penalty0 (2):\penalty0 346--358, April 1981.
\newblock ISSN 1558-2523.
\newblock \doi{10.1109/TAC.1981.1102596}.
\newblock URL \url{https://ieeexplore.ieee.org/document/1102596}.
\newblock Conference Name: IEEE Transactions on Automatic Control.

\bibitem[Stanculescu et~al.(2014)Stanculescu, Williams, and Freer]{stanculescu_hierarchical_2014}
Ioan Stanculescu, Christopher K.~I. Williams, and Yvonne Freer.
\newblock A {Hierarchical} {Switching} {Linear} {Dynamical} {System} {Applied} to the {Detection} of {Sepsis} in {Neonatal} {Condition} {Monitoring}.
\newblock In \emph{Proceedings of the 30th {Conference} on {Uncertainty} in {Artificial} {Intelligence} ({UAI} 2014)}. 2014.
\newblock URL \url{https://www.research.ed.ac.uk/en/publications/a-hierarchical-switching-linear-dynamical-system-applied-to-the-d}.

\bibitem[Storace and De~Feo(2004)]{storace_piecewise-linear_2004}
M.~Storace and O.~De~Feo.
\newblock Piecewise-linear approximation of nonlinear dynamical systems.
\newblock \emph{IEEE Transactions on Circuits and Systems I: Regular Papers}, 51\penalty0 (4):\penalty0 830--842, April 2004.
\newblock ISSN 1558-0806.
\newblock \doi{10.1109/TCSI.2004.823664}.
\newblock Conference Name: IEEE Transactions on Circuits and Systems I: Regular Papers.

\bibitem[Strogatz(2015)]{strogatz_nonlinear_2015}
Steven~H. Strogatz.
\newblock \emph{Nonlinear Dynamics and Chaos}.
\newblock {CRC} Press, 1 edition, 2015.
\newblock ISBN 978-0-429-96111-3.
\newblock \doi{10.1201/9780429492563}.
\newblock URL \url{https://www.taylorfrancis.com/books/9780429961113}.

\bibitem[Sun(2006)]{sun_switched_2006}
Zhendong Sun.
\newblock \emph{Switched {Linear} {Systems}: {Control} and {Design}}.
\newblock Springer Science \& Business Media, March 2006.
\newblock ISBN 978-1-84628-131-0.
\newblock Google-Books-ID: u4GArZN1bmsC.

\bibitem[Takens(1981)]{takens_detecting_1981}
Floris Takens.
\newblock Detecting strange attractors in turbulence.
\newblock In \emph{Dynamical {Systems} and {Turbulence}, {Warwick} 1980}, volume 898, pages 366--381. Springer, 1981.
\newblock ISBN 978-3-540-11171-9 978-3-540-38945-3.
\newblock URL \url{http://link.springer.com/10.1007/BFb0091924}.

\bibitem[Talathi and Vartak(2016)]{talathi_improving_2016}
Sachin~S. Talathi and Aniket Vartak.
\newblock Improving performance of recurrent neural network with relu nonlinearity.
\newblock In \emph{Proceedings of the 4th {International} {Conference} on {Learning} {Representations}}, 2016.
\newblock URL \url{http://arxiv.org/abs/1511.03771}.

\bibitem[Tomas‐Rodriguez and Banks(2003)]{tomasrodriguez_linear_2003}
M.~Tomas‐Rodriguez and S.~P. Banks.
\newblock Linear approximations to nonlinear dynamical systems with applications to stability and spectral theory.
\newblock \emph{IMA Journal of Mathematical Control and Information}, 20\penalty0 (1):\penalty0 89--103, March 2003.
\newblock ISSN 0265-0754.
\newblock \doi{10.1093/imamci/20.1.89}.
\newblock URL \url{https://doi.org/10.1093/imamci/20.1.89}.

\bibitem[Trischler and D’Eleuterio(2016)]{trischler_synthesis_2016}
Adam~P. Trischler and Gabriele~M.T. D’Eleuterio.
\newblock Synthesis of recurrent neural networks for dynamical system simulation.
\newblock \emph{Neural Networks}, 80:\penalty0 67--78, 2016.
\newblock ISSN 08936080.
\newblock \doi{10.1016/j.neunet.2016.04.001}.
\newblock URL \url{https://linkinghub.elsevier.com/retrieve/pii/S0893608016300314}.

\bibitem[Tziperman et~al.(1997)Tziperman, Scher, Zebiak, and Cane]{tziperman-97}
Eli Tziperman, Harvey Scher, Stephen~E. Zebiak, and Mark~A. Cane.
\newblock Controlling spatiotemporal chaos in a realistic el ni\~no prediction model.
\newblock \emph{Phys. Rev. Lett.}, 79:\penalty0 1034--1037, Aug 1997.
\newblock \doi{10.1103/PhysRevLett.79.1034}.
\newblock URL \url{https://link.aps.org/doi/10.1103/PhysRevLett.79.1034}.

\bibitem[Vlachas et~al.(2018)Vlachas, Byeon, Wan, Sapsis, and Koumoutsakos]{vlachas_data-driven_2018}
Pantelis~R. Vlachas, Wonmin Byeon, Zhong~Y. Wan, Themistoklis~P. Sapsis, and Petros Koumoutsakos.
\newblock Data-driven forecasting of high-dimensional chaotic systems with long short-term memory networks.
\newblock \emph{Proc. R. Soc. A.}, 474\penalty0 (2213):\penalty0 20170844, 2018.
\newblock ISSN 1364-5021, 1471-2946.
\newblock \doi{10.1098/rspa.2017.0844}.
\newblock URL \url{https://royalsocietypublishing.org/doi/10.1098/rspa.2017.0844}.

\bibitem[Vlachas et~al.(2020)Vlachas, Pathak, Hunt, Sapsis, Girvan, Ott, and Koumoutsakos]{vlachas_backpropagation_2020}
Pantelis~R. Vlachas, Jaideep Pathak, Brian~R. Hunt, Themistoklis~P. Sapsis, Michelle Girvan, Edward Ott, and Petros Koumoutsakos.
\newblock Backpropagation {Algorithms} and {Reservoir} {Computing} in {Recurrent} {Neural} {Networks} for the {Forecasting} of {Complex} {Spatiotemporal} {Dynamics}.
\newblock \emph{arXiv:1910.05266 [physics]}, February 2020.
\newblock URL \url{http://arxiv.org/abs/1910.05266}.
\newblock arXiv: 1910.05266.

\bibitem[Vogt et~al.(2022)Vogt, Puelma~Touzel, Shlizerman, and Lajoie]{vogt_lyapunov_2022}
Ryan Vogt, Maximilian Puelma~Touzel, Eli Shlizerman, and Guillaume Lajoie.
\newblock On lyapunov exponents for {RNNs}: Understanding information propagation using dynamical systems tools.
\newblock \emph{Frontiers in Applied Mathematics and Statistics}, 8, 2022.
\newblock ISSN 2297-4687.
\newblock URL \url{https://www.frontiersin.org/articles/10.3389/fams.2022.818799}.

\bibitem[Wang et~al.(2022)Wang, Dong, Arik, and Yu]{wang_koopman_2022}
Rui Wang, Yihe Dong, Sercan~Ö Arik, and Rose Yu.
\newblock Koopman {Neural} {Forecaster} for {Time} {Series} with {Temporal} {Distribution} {Shifts}, October 2022.
\newblock URL \url{http://arxiv.org/abs/2210.03675}.
\newblock arXiv:2210.03675 [cs, stat].

\bibitem[Wersing et~al.(2001)Wersing, Beyn, and Ritter]{wersing_dynamical_2001}
H.~Wersing, W.~J. Beyn, and H.~Ritter.
\newblock Dynamical stability conditions for recurrent neural networks with unsaturating piecewise linear transfer functions.
\newblock \emph{Neural Computation}, 13\penalty0 (8):\penalty0 1811--1825, August 2001.
\newblock ISSN 0899-7667.
\newblock \doi{10.1162/08997660152469350}.

\bibitem[Wiggins(1988)]{wiggins_global_1988}
Stephen Wiggins.
\newblock \emph{Global {Bifurcations} and {Chaos}}, volume~73 of \emph{Applied {Mathematical} {Sciences}}.
\newblock Springer, New York, NY, 1988.
\newblock ISBN 978-1-4612-1041-2 978-1-4612-1042-9.
\newblock \doi{10.1007/978-1-4612-1042-9}.
\newblock URL \url{http://link.springer.com/10.1007/978-1-4612-1042-9}.

\bibitem[Wolf et~al.(1985)Wolf, Swift, Swinney, and Vastano]{wolf_determining_1985}
Alan Wolf, Jack~B. Swift, Harry~L. Swinney, and John~A. Vastano.
\newblock Determining lyapunov exponents from a time series.
\newblock \emph{Physica D: Nonlinear Phenomena}, 16\penalty0 (3):\penalty0 285--317, 1985.
\newblock ISSN 0167-2789.
\newblock \doi{10.1016/0167-2789(85)90011-9}.
\newblock URL \url{https://www.sciencedirect.com/science/article/pii/0167278985900119}.

\bibitem[Wood(2010)]{wood_statistical_2010}
Simon~N. Wood.
\newblock Statistical inference for noisy nonlinear ecological dynamic systems.
\newblock \emph{Nature}, 466\penalty0 (7310):\penalty0 1102--1104, August 2010.
\newblock ISSN 1476-4687.
\newblock \doi{10.1038/nature09319}.
\newblock URL \url{https://www.nature.com/articles/nature09319}.
\newblock Number: 7310 Publisher: Nature Publishing Group.

\bibitem[Yi et~al.(2003)Yi, Tan, and Lee]{yi_multistability_2003}
Zhang Yi, K.~K. Tan, and T.~H. Lee.
\newblock Multistability {Analysis} for {Recurrent} {Neural} {Networks} with {Unsaturating} {Piecewise} {Linear} {Transfer} {Functions}.
\newblock \emph{Neural Computation}, 15\penalty0 (3):\penalty0 639--662, March 2003.
\newblock ISSN 0899-7667.
\newblock \doi{10.1162/089976603321192112}.
\newblock URL \url{https://doi.org/10.1162/089976603321192112}.

\end{thebibliography}
\bibliographystyle{plainnat}

\newpage
\appendix

\section{Appendix}

\subsection{Graph Representation}\label{appx:graph_repr}
\begin{figure*}[!htb]
    \centering
	\includegraphics[width=0.99\linewidth]{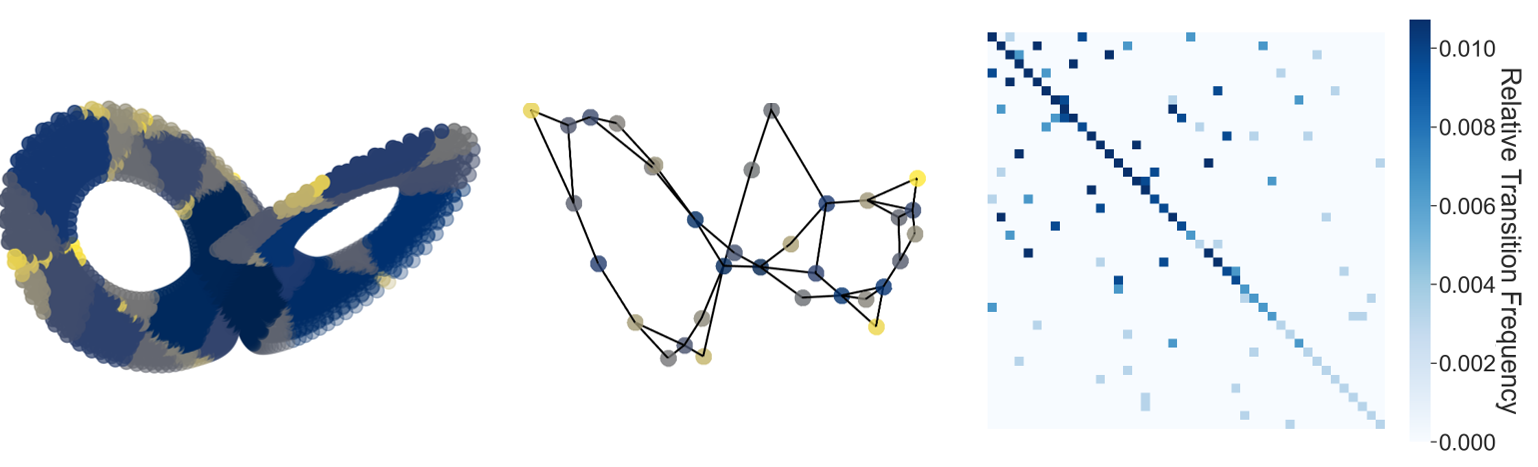}
	\caption{Optimal geometric reconstruction of a Lorenz-63 using the AL-RNN with $P=8$ PWL units. Left: reconstruction with subregions color-coded by frequency of trajectory visits (dark: most frequently visited regions, yellow: least frequent regions). Center: Resulting geometrical graph structure (using transition probabilities for placing the nodes) visualized using the spectral layout in \texttt{networkx}. Note that self-connections and directedness of edges were omitted in this representation. The resulting graph shadows the layout of the reconstructed system. Right: Connectome of transitions between subregions.}
	\label{fig:graphs_examples}
\end{figure*}

\begin{figure*}[!htb]
    \centering
	\includegraphics[width=0.99\linewidth]{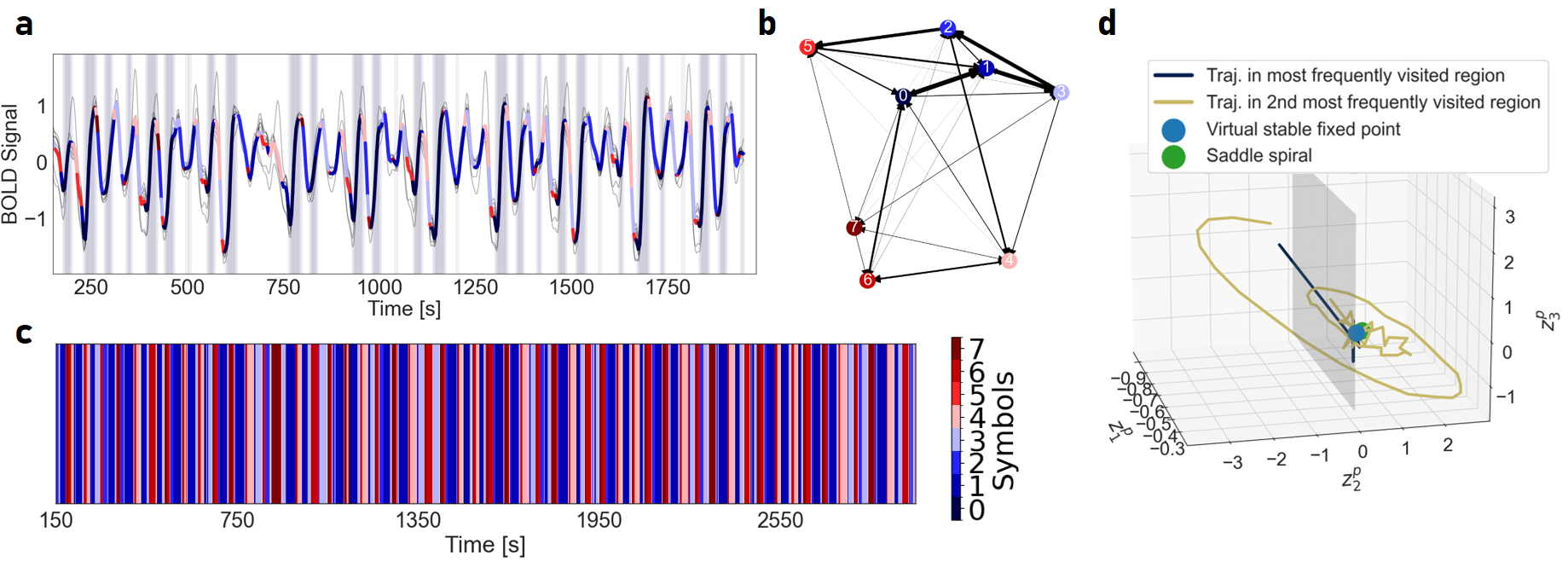}
	\caption{Freely generated fMRI activity using an AL-RNN with $M=100$ total units and $P=3$ PWL units. \textbf{a}: Mean generated activity color-coded according to linear subregions, with background shading highlighting the most frequently visited subregion. \textbf{b}: Geometrical graph representation of connections between linear subregions, with edge weights representing relative transition frequencies (self-connections omitted). \textbf{c}: Time series of the symbolic coding of dynamics according to linear subregions. \textbf{d}: Dynamics in the two most frequently visited linear subregions in the subspace of the three PWL units, with the boundary between subregions in gray. The dark blue trajectory bit in the first subregion moves towards a virtual stable fixed point located near the center of the saddle spiral in the second subregion. The yellow trajectory illustrates an orbit cycling away from this spiral point and eventually crossing into the first subregion. From there, trajectories are pulled back into the second subregion through the virtual stable fixed point located close to the saddle spiral (see also activity with background shading in \textbf{a}). This dynamical behavior is similar to the one observed in the chaotic benchmark systems, where locally divergent activity of the AL-RNN is propelled back into the center of an unstable manifold within another subregion.}
	\label{fig:fMRI_freely_generated}
\end{figure*}

The symbolic graph construction followed the rules given in Sect. \ref{sec:methods_symbolic_approach}: Most generally, each linear subregion $U_i$ is assigned a node (symbol), and a directed edge is drawn between nodes $i,j, i \rightarrow j$, whenever $F_{\bm{\theta}}(U_i) \cap U_j \neq \varnothing$. Here, however, we are mostly interested in the topological graphs representing particular chaotic attractors (like the Lorenz-63 or Rössler attractor), and hence restrict the graph representation to the nodes corresponding to subregions visited by trajectories on the attractor $B$, and edges drawn when $F_{\bm{\theta}}(U_i \cap B) \cap U_j \neq \varnothing$. More specifically, we first sampled a long trajectory $\bm{{Z}}=\{\bm{{z}}_1,\cdots,\bm{{z}}_T\}$ with $T=100,000$ time steps, removed the first $1000$ time steps as transients, and counted all transitions between any two subregions $U_i, U_j$. To obtain a more nuanced geometrical/ statistical picture, we also evaluated the relative number of time steps the trajectory spent in each subregion $U_i$ (i.e., an estimate of the occupation measure), %$|\{F_{\bm{\theta}}(\bm{z}_{t}) \in U_i\}|/(100,000-1000)$, 
$|\{\bm{z}_{t} \in U_i, t>1000\}|/(100,000-1000)$, 
as well as the relative frequency of transitions between any two subregions $U_i, U_j$, $|\{\bm{z}_{t},\bm{z}_{t+1}| t>1000, \bm{z}_{t} \in U_i,F_{\bm{\theta}}(\bm{z}_{t}) \in U_j\}|/(100,000-1001)$, yielding a weight for each edge, or a distance between nodes through the graph's Laplacian (see below). To enhance the readability of the larger graphs in Figs. \ref{fig:graphs_roessler}, \ref{fig:graphs_examples}, \ref{fig:fMRI_freely_generated} and \ref{fig:fMRI_pca}, we removed self-connections (time steps where the trajectory remained within a single subregion).% and directedness information (arrows).

\paragraph{Laplacian matrix}
\label{laplacian_matrix}

The Laplacian matrix of a graph is defined as $\bm{L}=\bm{D}-\bm{A}$
where $\bm{A}$ is the adjacency matrix of the graph containing the weights (transition probabilities), and $\bm{D}$ is the out-degree matrix, which is a diagonal matrix where each element is the sum of the outgoing edge weights of node $i$, $D_{ii} = \sum_j A_{ij}$. The spectral layout in \texttt{networkx} uses the eigenvectors of the Laplacian matrix corresponding to the smallest non-zero eigenvalues as positions for the nodes. This tends to group more tightly connected nodes closer together. The Laplacian is more widely used as a dimensionality reduction technique in ML, for example in Laplacian eigenmaps \citep{belkin_laplacian_2001}, and has also been used to represent discretizations of PDEs as graphs \citep{solomon_pde_2015}.

\paragraph{Proximity matching} \label{appx:proximity_matching}

The mapping from latent space to observation space is not unique because $M>N$, and hence the nullspace of the linear observation model is non-empty (does not contain just the $\bm{0}$ vector). For the AL-RNN this is evident from the fact that the non-readout neurons, particularly the PWL neurons, do not contribute to the observations. However, in practice, for the freely generated activity of trained AL-RNNs, points that fall into the same linear subregion in latent space also were close in observation space. This implies that attractors are segmented into different subregions in latent space in accordance with the observable dynamics. To numerically confirm this, we found that proximal points in observation space were typically related to the same linear subregion when generating activity from trained AL-RNNs: We conducted proximity matching by defining a threshold distance (e.g. $d=0.05$, corresponding to $5 \%$ of the data variance) and assessing whether generated latent trajectory points proximal in observation space fall into the same or different subregions. We found that for the geometrically minimal reconstruction of the Rössler system (Fig. \ref{fig:graphs_roessler}), only $6\%$ of proximal data points (within $d$) belonged to different subregions, % primarily on the lobe in the z plane,
while for the Lorenz-63 attractor (Fig. \ref{fig:graphs_examples}) $4\%$ of proximal data points (within $d$) belonged to different subregions, confirming that the attractors were segmented into relatively distinct patches.

\subsection{Methodological Details} \label{appx:method_details}

\paragraph{Training method} \label{appx:stf_training}

Our training method rests on a variant of sparse teacher forcing. This approach has recently been proven to effectively tackle gradient divergence when training on observations from chaotic DS \cite{mikhaeil_difficulty_2022} and has shown SOTA performance on benchmark and real-world systems in DSR \cite{brenner_tractable_2022, hess_generalized_2023}. In sparse teacher forcing, the idea is to directly replace latent states (or a subset of them) with states inferred from observations at intervals $\tau$ while leaving the network to evolve freely otherwise. To obtain forced states, the observation model needs to be `pseudo-inverted'. Here we employ a specific variant of sparse teacher forcing called identity teacher forcing \cite{mikhaeil_difficulty_2022, brenner_tractable_2022}, where this pseudo-inversion becomes trivial by adopting an identity mapping as the observation model:
\begin{equation}\label{eq:obs:identity}
    \hat\vx_t = \mathcal{I}\bm{z}_t,
\end{equation}
with $\mathcal{I} \in \mathbb{R}^{N \times M}$, and $\mathcal{I}_{rr}=1$ for the $N$ read-out neurons, $r\leq N$, and zeroes elsewhere. During training, the read-out states are replaced with observations every $\tau$ time steps:
\begin{align}\label{eq:forcing}
\bm{z}_{t+1} =
  \begin{cases}
       F_{\bm{\theta}}(\tilde{\bm{z}}_{t}) & \text{if $t\in \mathcal{T}$} \\
       F_{\bm{\theta}}(\bm{z}_{t}) & \text{else} 
  \end{cases}
\end{align}
with $\mathcal{T}=\{l\tau+1\}_{l \in \mathbb{N}_0}$, and $\tilde{\bm{z}}_{t}=(\vx_t,\bm{z}_{N+1:M,t})^T$. Employing identity teacher forcing splits the AL-RNN into essentially three types of units, the $N$ linear readout-neurons $\bm{z}^r_t$ which are equivalent to the predicted observations and teacher-forced during training, the remaining $M-P-N$ linear neurons $\bm{z}^l_t$, and the $P$ nonlinear neurons $\bm{z}^p_t$. The overall model and architecture are illustrated in Fig. \ref{fig:rnn_architecture}. The AL-RNN can be trained using Mean Squared Error (MSE) loss over model predictions and observations:
\begin{equation}\label{eq:mse_loss}
\ell_{MSE}(\hat{\bm{X}}, \bm{X}) = \frac{1}{N \cdot T}\sum_{t=1}^{T} \left\lVert \hat{\bm{x}}_t - \bm{x}_t \right\rVert_2^2,
\end{equation}
where $\hat{\bm{X}}$ are the model predictions and $\bm{X}$ denotes the training sequence of length $T$. We found that performance was better when the read-out neurons were linear rather than ReLU-based. Note that the $M-N$ non-readout neurons, including the PWL neurons, do not explicitly contribute to the loss function. We used rectified adaptive moment estimation (RADAM) \cite{Liu2020On} as the optimizer, with $L=50$ batches with $S=16$ sequences per epoch. Further, we chose $M=\{20,20,100,100,100,130\}$, $\tau=\{16,8,10,7,20,10\}$, $T=\{200,300,50,72,50,100\}$, initial learning rates $\eta_{\text{start}}=\{10^{-3},5\cdot10^{-3},2\cdot10^{-3},5\cdot10^{-3},10^{-3},10^{-3}\}$, $\eta_{\text{end}}=10^{-5}$ and $epochs=\{2000,3000,4000,2000,3000,2000\}$ for the \{Lorenz-63, Rössler, ECG, fMRI,Lorenz-96,EEG\} dataset, respectively. Parameters in $\bm{W}$ were initialized using a Gaussian initialization with $\sigma=0.01$, $\bm{h}$ as a vector of zeros, and $\bm{A}$ as the diagonal of a normalized positive-definite random matrix \cite{brenner_tractable_2022,talathi_improving_2016}. The initial latent state $\bm{z}_1 = [\bm{x}_1, \bm{Lx}_1]^T$
is estimated from $\bm{x}_1$ using a matrix $\bm{L}\in\mathbb{R}^{(M-N)\times N}$ which is jointly learned with the other model parameters.
Additionally, for the Rössler and Lorenz systems, we added $5\%$ observation noise during training. Across all training epochs of a given run, we consistently selected the model with the lowest $D_{\text{stsp}}$.  Each individual training run of the AL-RNN was performed on a single CPU. Depending on the training sequence length, a single epoch took between 0.5 to 3 seconds.

\paragraph{Geometric agreement} \label{appx:geometric_agreement}

For evaluating attractor geometries, we use a state space measure $D_{\textrm{stsp}}$ based on the Kullback-Leibler (KL) divergence, which assesses the (mis)match between the ground truth spatial distribution of data points, $p_{\text {true}}(\vx)$, and the distribution $p_{\text {gen}}(\vx|\vz)$ of trajectory points freely generated by the inferred DSR model. These probability distributions can be approximated in different ways from the observed/ generated trajectories. Here, we usually sampled long trajectories corresponding to the test set length (usually $T=100,000$ time steps, but sometimes shorter for the empirical time series) from trained systems, removing transients to ensure that the system has reached a limit set. For low-dimensional systems, the KL divergence can be straightforwardly calculated through a discrete binning approximation \cite{brenner_tractable_2022}:
\begin{align}\label{eq:D_stsp}
D_{\mathrm{stsp}}\left(p_{\mathrm {true }}(\vx), p_{\mathrm {gen }}(\vx \mid \vz)\right) \approx \sum_{k=1}^{K} \hat{p}_{\mathrm {true }}^{(k)}(\vx) \log \left(\frac{\hat{p}_{\mathrm{true }}^{(k)}(\vx)}{\hat{p}_{\mathrm {gen }}^{(k)}(\vx \mid \vz)}\right),
\end{align}
where $K = m^{N}$ is the total number of bins, with $m$ bins per dimension and $N$ being the dimension of the ground truth system. A bin number of $m=30$ per dimension was chosen as a good compromise for distinguishing between successful and bad reconstructions for 3d systems. Since the number of data required to fill the bins scales exponentially with $N$, for the ECG time series ($N=5$) we reduced the number of bins to $m=8$, as suggested in \cite{hemmer_optimal_2024}.

\paragraph{Temporal agreement} \label{appx:temporal_agreement}

To assess temporal agreement, we computed Hellinger distances $D_{H}$ between power spectra \cite{mikhaeil_difficulty_2022, hess_generalized_2023}. We first simulated long time series $T=100,000$ (as with $D_{stsp}$ above). After standardization, we computed dimension-wise Fast Fourier Transforms (FFT). The power spectra were smoothened using a Gaussian kernel and normalized, and the extended, high-frequency tails, which predominantly contained noise, were truncated. The Hellinger distance between smoothed ground-truth spectra $F(\omega)$ and generated spectra $G(\omega)$ is given by:
\begin{equation}
H(F(\omega), G(\omega))=\sqrt{1-\int_{-\infty}^{\infty} \sqrt{F(\omega) G(\omega)} d \omega} \in[0,1]
\end{equation}

\paragraph{Maximum Lyapunov exponent} \label{appx:maximum_lyap_exponent}

The maximum Lyapunov exponent of a system quantifies how fast nearby trajectories diverge, and for a flow map can be computed in the limit $T \rightarrow \infty$ from the system's product of Jacobians. To approximate the maximum exponent numerically, we first iterated a trained model forward from some randomly drawn initial condition and discarded transients. Given that for chaotic systems the product of Jacobians itself explodes \cite{mikhaeil_difficulty_2022}, we employed a numerical algorithm from \cite{wolf_determining_1985, vogt_lyapunov_2022} that re-orthogonalizes the series of Jacobian products at regular intervals using a QR decomposition.

\paragraph{Categorical decoder}

We coupled categorical observations to the $P$ PWL neurons via a link function given by
\begin{align} \label{eq:supp:cat_obs}
&\pi_i=\frac{\exp \left(\boldsymbol{\beta}_i^T \mathbf{z}^p_t\right)}{1+\sum_{j=1}^{K-1} \exp \left(\boldsymbol{\beta}_j^T \mathbf{z}^p_t\right)} \quad \forall i \in\{1 \ldots K-1\}  \\
&\pi_K=\frac{1}{1+\sum_{j=1}^{K-1} \exp \left(\boldsymbol{\beta}_j^T \mathbf{z}^p_t\right)} \nonumber.
\end{align}

The regression weights $\boldsymbol{\beta}_i \in \mathbb{R}^{P \times 1}$ are learned for each category $i=1 \ldots K-1$, while the normalization ensures that the total probability over all categories sums to one, $\sum_{i=1}^K \pi_i=1$.

\subsection{Benchmark datasets} \label{appx:datasets}

\paragraph{Lorenz-63}

The Lorenz-63 system, formulated by Edward Lorenz in 1963 \cite{lorenz_deterministic_1963} to model atmospheric convection, is defined by
\begin{align}
   % \begin{gathered}
        \frac{\text{d}x}{\text{d}t} & = \sigma(y-x)\\ \nonumber
        \frac{\text{d}y}{\text{d}t} & = x(\rho-z)-y\\ \nonumber
        \frac{\text{d}z}{\text{d}t} & = xy-\beta z,
   % \end{gathered}
\end{align}
where %$x,y,z$ represent the three dimensions and 
$\sigma, \rho, \beta$, are parameters that control the dynamics of the system (here set to $\sigma=10$, $\beta=\frac{8}{3}$, and $\rho=28$, in the chaotic regime). The system was solved numerically with integration time step $\Delta t = 0.01$ using \texttt{scipy.integrate} with the default \texttt{RK45} solver.

\paragraph{Rössler}

The Rössler system, intended by Otto Rössler in 1976 \cite{rossler_equation_1976} as a further simplification of the Lorenz model, produces chaotic dynamics using a single nonlinearity in one state variable:
\begin{align} \label{eq:roessler}
   % \begin{gathered}
        \frac{\text{d}x}{\text{d}t} & = -y-z\\ \nonumber
        \frac{\text{d}y}{\text{d}t} & = x+ay\\ \nonumber
        \frac{\text{d}z}{\text{d}t} & = b+z(x-c),
   % \end{gathered}
\end{align}
where $a$, $b$, $c$, are parameters controlling the dynamics of the system (here set to $a = 0.2$, $b = 0.2$, and $c = 5.7$, in the chaotic regime). The system was solved with integration time step $\Delta t = 0.08$ using \texttt{scipy.integrate} with the default \texttt{RK45} solver.

\paragraph{Human electrocardiogram}

The electrocardiogram (ECG) time series was taken from the PPG-DaLiA dataset \cite{reiss2019deep}. With a sampling frequency of $700Hz$, the recording duration spanned 600 seconds, yielding a time series of $T=419,973$ time points. Initially, a Gaussian smoothing filter with $\sigma=6$ was applied, resulting in a filter length of $l=8\sigma+1=49$. We standardized the time series and applied temporal delay embedding using the \texttt{DynamicalSystems.jl} Julia library, resulting in an embedding dimension of $m=5$. For model training, the first $T=100,000$ time steps (approximately 143 seconds) were used, downsampled to every 10th datapoint.

\paragraph{Human fMRI data}

The functional magnetic resonance imaging (fMRI) data from human subjects performing three cognitive tasks is publicly available on GitHub \cite{kramer22a}. We followed \citet{kramer22a} and selected the first principal component of BOLD activity in each of the 20 brain regions. Subjects underwent five rounds of three cognitive tasks (`Choice Reaction Task [CRT]’, `Continuous Delayed Response Task [CDRT]’ and `Continuous Matching Task [CMT]’), together with a `Rest’ and `Instruction’ period. The time series per subject were short ($T=360$) and reconstructions in \cite{kramer22a} exhibited a positive maximum Lyapunov exponent, indicating the chaotic nature of the underlying system.

\paragraph{Lorenz-96}
The Lorenz-96 system, formulated by Edward Lorenz in 1996 \cite{lorenz_predictability_1996}, is defined by 
\begin{equation}
    \frac{\text{d}x_i}{\text{d}t}=(x_{i+1}-x_{i-2})x_{i-1}-x_i+F,
\end{equation}
with system variables $x_i$, $i=1,...,N$, and forcing term $F$ (here, $F=8$, in the chaotic regime). Furthermore, cyclic boundary conditions are assumed with $x_{-1}=x_{N-1}$, $x_0=x_N$, $x_{N+1}=x_1$, and the system was solved with integration step $\Delta t = 0.04$ using \texttt{scipy.integrate} with the default \texttt{RK45} solver.

\paragraph{Human EEG data}
Electroencephalogram (EEG) data were taken from a study by \citet{schalk_bci2000_2004}. These are 64-channel EEG recordings obtained from human subjects during different motor and imagery tasks. The signal was standardized and smoothed using a Hann function and a window length of 15, as in \cite{brenner_tractable_2022}.

\subsection{Further Results}

\begin{table}[!ht]
\definecolor{Gray}{gray}{0.825}
\caption{Comparison of AL-RNN to different SOTA models for dynamical systems reconstruction. Comparison values, datasets and evaluation measures as in \cite{hess_generalized_2023}, based on code provided on GitHub by the authors. id-TF: identity teacher forcing, GTF: generalized teacher forcing, $D_{stsp}$: state space divergence, $D_H$: Hellinger distance across power spectra. Reported values are median $\pm$ median absolute deviation. \\}
\centering
\scalebox{0.75}{
\begin{tabular}{l l c c c}
\toprule
Dataset & Method & $D_{\textrm{stsp}}$ $\downarrow$ & $D_H$ $\downarrow$ & $\lvert \bm{\theta} \rvert$ \\
\midrule
\rowcolor{Gray}\multirow{7}{4em}{Lorenz-63 (3d)}
& \cellcolor{Gray}{AL-RNN + id-TF} & $0.34\pm0.05$ & $0.081\pm0.012$ & $360$ \\
& shPLRNN + GTF & $0.26\pm0.03$ & $0.090\pm0.007$ & $365$ \\
& dend-RNN + id-TF & $0.9\pm0.2$ & $0.15\pm0.03$ & $361$ \\
& {Reservoir Computer} & $0.52\pm0.12$ & $0.19\pm0.04$ & $603$ \\
& {LSTM-TBPTT} & $0.46\pm0.22$ & $0.11\pm0.03$ & $1188$ \\
& {SINDy} & $0.24\pm0.00$ & $0.091\pm0.000$ & $30$ \\
& {Neural-ODE} & $0.63\pm0.2$ & $0.15\pm0.05$ & $353$ \\
& {Long Expressive Memory} & $0.39\pm0.24$ & $0.12\pm0.05$ & $360$ \\
\midrule
\rowcolor{Gray}\multirow{8}{4em}{ECG}
& \cellcolor{Gray}{AL-RNN + id-TF} & $3.0\pm0.7$ & $0.29\pm0.04$ & $2808$ \\
& shPLRNN + GTF & $4.3\pm0.6$ & $0.34\pm0.02$ & $2785$ \\
& dendPLRNN + id-TF & $5.8\pm0.6$ & $0.37\pm0.06$ & $3245$ \\
& {Reservoir Computer} & $5.3\pm1.7$ & $0.39\pm0.05$ & $5000$ \\
& {LSTM-TBPTT} & $15.2\pm0.5$ & $0.73\pm0.02$ & $5920$ \\
& {SINDy} & diverging & diverging & $3960$ \\
& {Neural-ODE} & $12.2\pm0.7$ & $0.7\pm0.03$ & $4955$ \\
& {Long Expressive Memory} & $16.3\pm0.2$ & $0.56\pm0.04$ & $4872$ \\
\midrule
\rowcolor{Gray}\multirow{8}{4em}{Lorenz-96 (20d)}
& \cellcolor{Gray}{AL-RNN + id-TF} & $1.64\pm0.03$ & $0.089\pm0.001$ & $4623$ \\
& shPLRNN + GTF & $1.68\pm0.06$ & $0.072\pm0.001$ & $4540$ \\
& dendPLRNN + id-TF & $1.65\pm0.05$ & $0.083\pm0.005$ & $5740$ \\
& {Reservoir Computer} & $2.40\pm0.15$ & $0.14\pm0.02$ & $12000$ \\
& {LSTM-TBPTT} & $5\pm1$ & $0.31\pm0.04$ & $10580$ \\
& {SINDy} & $1.59 \pm 0.00$ & $0.06\pm0.00$ & $4620$ \\
& {Neural-ODE} & $1.77\pm0.07$ & $0.076\pm0.01$ & $4530$ \\
& {Long Expressive Memory} & $7.2\pm2.3$ & $0.54\pm0.13$ & $4620$ \\

\midrule
\rowcolor{Gray}\multirow{8}{4em}{EEG (64d)}
& \cellcolor{Gray}{AL-RNN + id-TF } & $2.6\pm0.3$ & $0.14\pm0.03$ & $17955$ \\
& shPLRNN + GTF & $2.1\pm0.2$ & $0.11\pm0.01$ & $17952$ \\
& dendPLRNN + id-TF & $3\pm1$ & $0.13\pm0.04$ & $18099$ \\
& {Reservoir Computer} & $14\pm7$ & $0.54\pm0.15$ & $28672$ \\
& {LSTM-TBPTT} & $30\pm21$ & $0.2\pm0.1$ & $51584$ \\
& {SINDy} & diverging & diverging & $133120$ \\
& {Neural-ODE} & $20\pm0.5$ & $0.47\pm0.01$ & $17995$ \\
& {Long Expressive Memory} & $10.2\pm1.5$ & $0.38\pm0.06$ & $18304$ \\

\bottomrule
\end{tabular}

}
\label{tab:gtf_benchmarks}
\end{table}

\begin{figure}[!htb]
    \centering
	\includegraphics[width=0.75\linewidth]{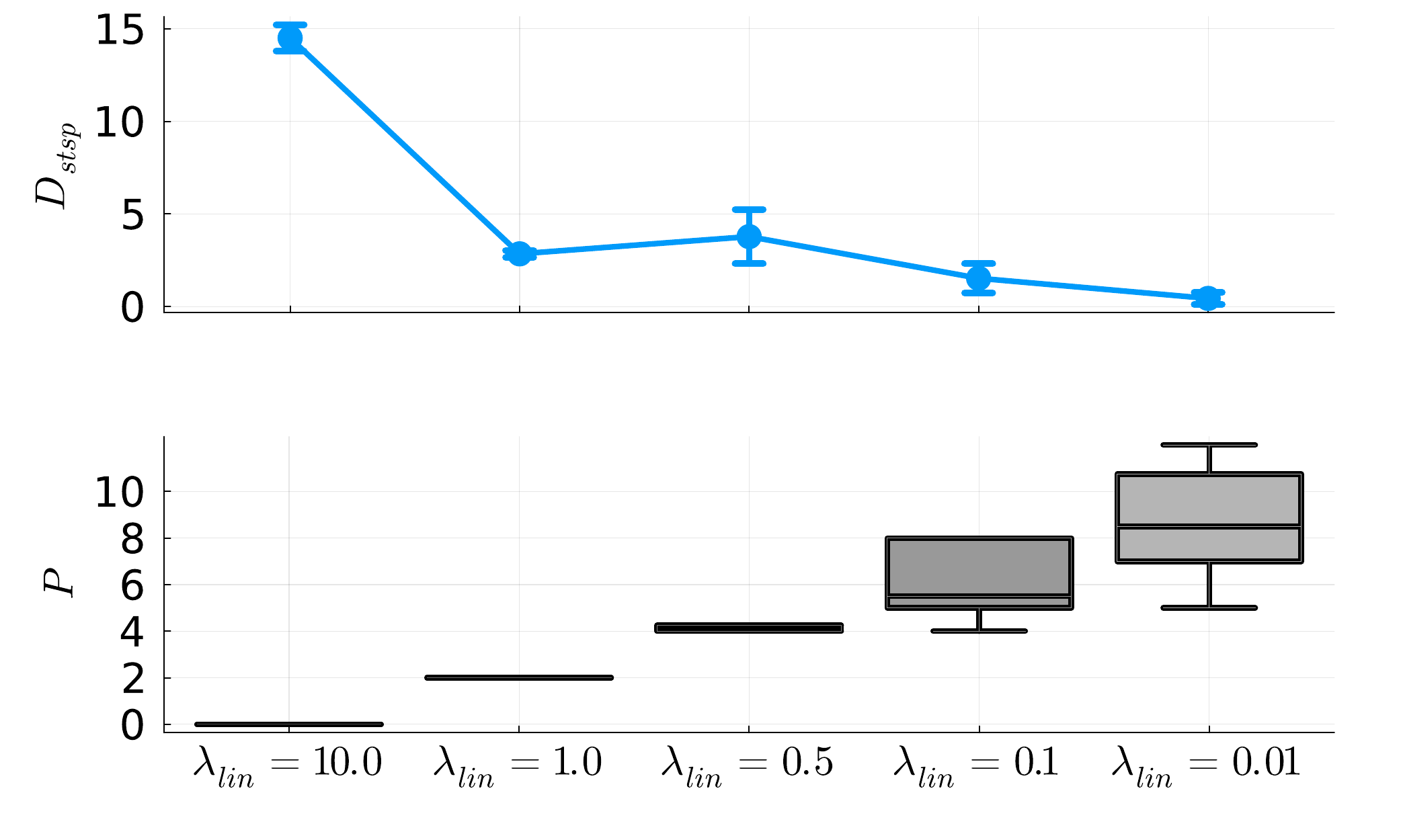}
	\caption{Top: DSR quality (assessed by $D_{stsp}$) as a function of strength of regularization on the number of nonlinearities for the AL-RNN trained on Lorenz-63. Bottom: Number of %identified linear units dependent on regularization strength. 
 selected piecewise-linear units $P$ as a function of regularization strength. As in Fig. \ref{fig:performance_relu}, a first optimum consistently occurs for $P=2$. 
 To select the number of nonlinear units through regularization, we replaced the standard ReLU by a leaky ReLU $\max(\alpha_i z_i,z_i), \alpha_i \in (0,1)$, for each of the $i= 1 \dots M$ units. The slope $\alpha_i=\sigma(\gamma_i)$ is determined through a steep sigmoid, $\sigma(\gamma_i)=1/(1+\exp(-500(\gamma_i-0.5)))$, via trainable parameter $\gamma_i$, ensuring that it is either close to $0$ or close to $1$. To encourage linearity, we include a loss term $\mathcal{L}_{\text{lin}} = \lambda_{\text{lin}} \sum_{i=1}^{M} |\alpha_i - 1|$, pushing slopes towards $1$. After training, units with $\alpha_i \approx 1$ are classified as linear, while all remaining units were considered nonlinear to provide an estimate for $P$.}
	\label{fig:regularization}
\end{figure}

\paragraph{Task stages align with subregions} \label{alignment_score}
To test the alignment of the reconstructed AL-RNN activity in the subspace of the $P=2$ PWL units with the $4$ task stages, we trained $10$ models on each of the $10$ subjects without visible movement artifacts (yielding $100$ trained models). We then checked which assignment of the four subregions (00, 01, 10, and 11) to the four task stages (Rest and Instruction, CRT, CDRT, and CMT) gave the highest alignment (Fig. \ref{fig:fMRI_linear_regions}), and used this to determine the average classification accuracy as the percentage of time points correctly assigned to their respective task stage based on the four subregions.

\paragraph{Geometrically minimal reconstructions of empirical time series}

\begin{figure*}[!htb]
    \centering
	\includegraphics[width=0.99\linewidth]{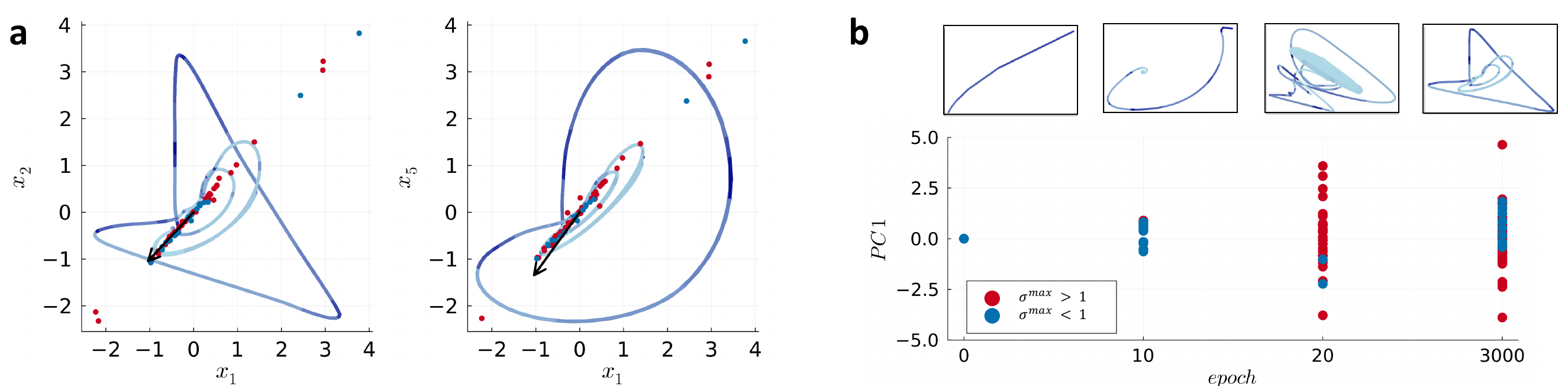}
	\caption{\textbf{a}: Variation in the location of the analytically computed real and virtual fixed points of the linear subregions aligns with the first PC of the data (generated trajectories in bluish, color-coded according to linear subregion). \textbf{b}: Fixed point location along the first principal component (with corresponding dynamics within $(x_1$,$x_2)$-plane of observation space on top) at different characteristic stages of training. At the early stages of training, fixed points of the linear subregions are distributed along the data manifold within the subspace of readout units. Around epoch $20$, the maximum absolute eigenvalues $\sigma^{\text{max}}$ of the Jacobians in many subregions become larger than one, inducing local divergence necessary for producing the observed chaotic dynamics.}
	\label{fig:ECG_pca}
\end{figure*}

\begin{figure*}[!htb]
    \centering
	\includegraphics[width=0.99\linewidth]{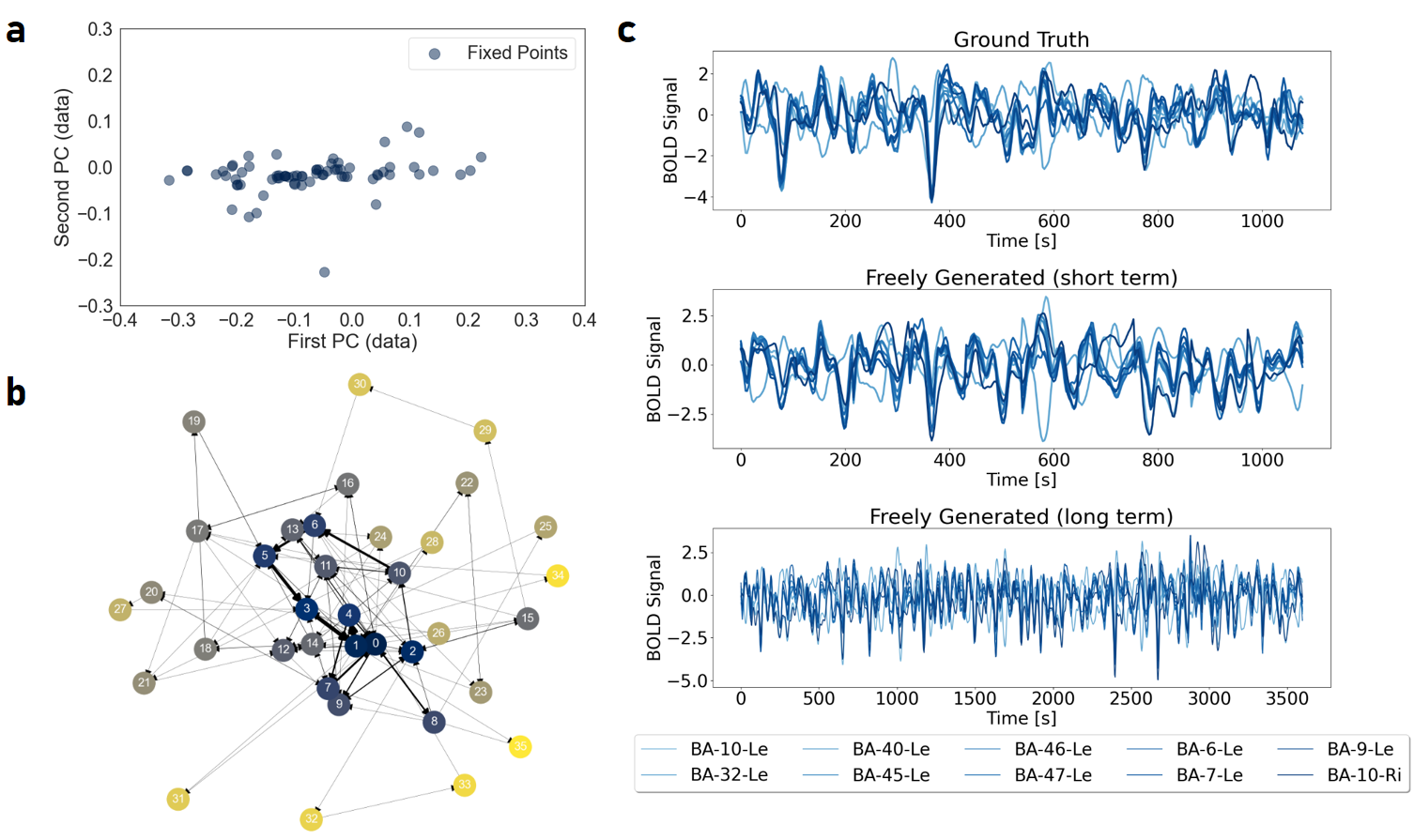}
	\caption{\textbf{a}: Analytically computed real and virtual fixed points of the linear subregions of a geometrically minimal AL-RNN ($M=100, P=10$) align with the first PC of the data within the subspace of readout units. The BOLD time series for different brain regions were highly correlated, so PC1 accounted for approximately $80\%$ of the data variance. \textbf{b}: Geometrical graph representation with relative frequency of transitions between linear subregions indicated by line thickness of edges, showing a central highly connected subgraph of frequently visited (bluish) dominant subregions, as in Fig. \ref{fig:subregions_covered}. \textbf{c}: Example freely generated activity from ten simulated brain regions.}
	\label{fig:fMRI_pca}
\end{figure*}

Increasing the number of PWL units also improved geometric agreement for the empirical time series (Fig. \ref{fig:performance_relu}, $P=10$ for the ECG data), while dynamics remained confined to a relatively small subset of linear subregions (Fig. \ref{fig:subregions_covered}). This confinement within only a few subregions allows for the efficient computation of dynamical objects like fixed points. For the ECG data, real and virtual fixed points in the linear subregions were primarily located within a 3d hyperplane of the 5d data. Principal component analysis showed that this hyperplane harboring the fixed points aligned with the first principal component of the data, explaining approximately $40\%$ of the data's variance (Fig. \ref{fig:ECG_pca}\textbf{a}/\textbf{b}). A similar pattern was observed in the fMRI data, where fixed point locations often aligned with PC1 of the data (Fig. \ref{fig:fMRI_pca}).

\begin{figure*}[!htb]
    \begin{subfigure}[b]{0.99\textwidth}
         \includegraphics[width=\textwidth]{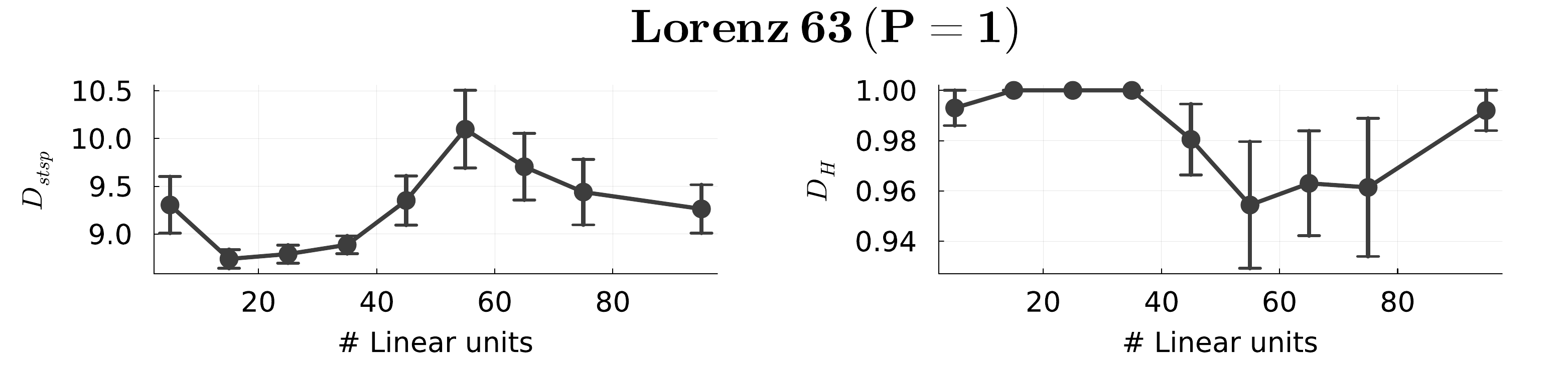}
    \end{subfigure}
    \hfill
    \begin{subfigure}[b]{0.99\textwidth}
         \includegraphics[width=\textwidth]{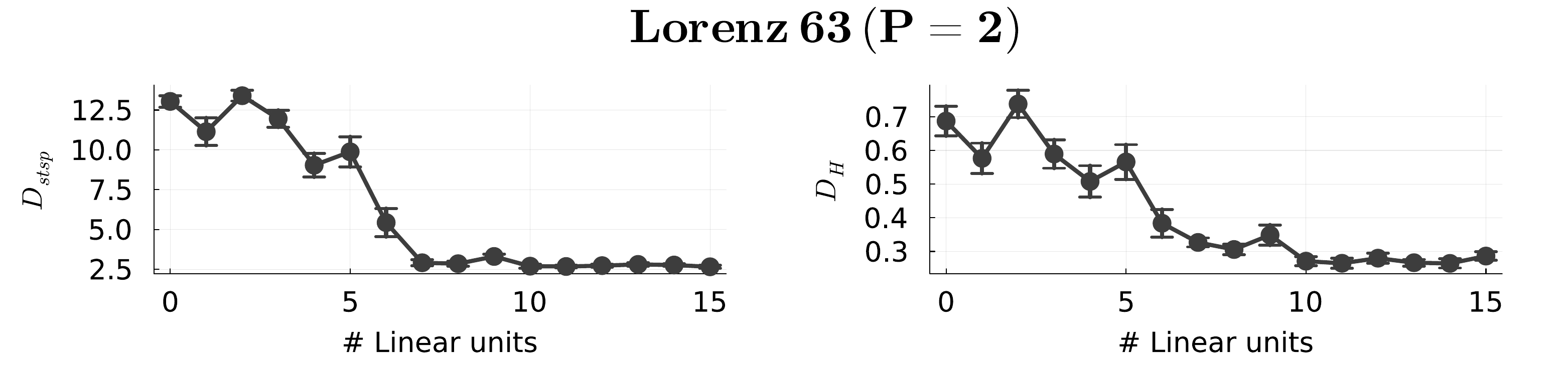}
    \end{subfigure}
	\caption{Reconstruction performance on Lorenz-63 system for an AL-RNN ($M=20$) as a function of the number of linear units, once for the case where the number of PWL units was insufficient for a topologically accurate reconstruction ($P=1$, top), and once for the case where it was sufficient ($P=2$, bottom). Results indicate performance cannot be improved by adding more linear units if $P$ is too small, but can be -- up to some saturation level -- when $P$ is sufficiently large. Error bars = SEM.}
	\label{fig:performance_M_Lorenz63}
\end{figure*}

\begin{figure*}[!htb]
    \centering
	\includegraphics[width=0.99\linewidth]{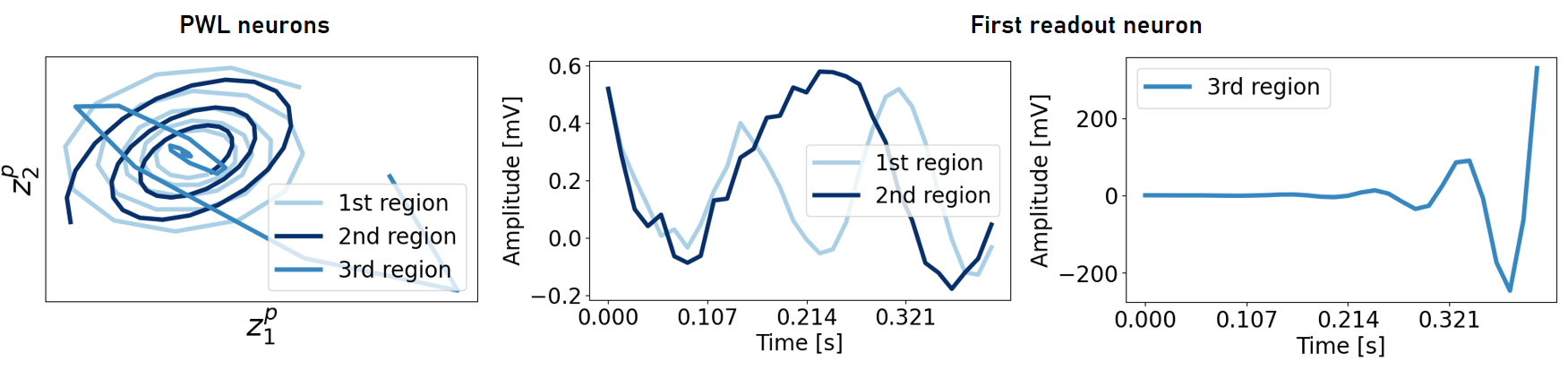}
	\caption{`Linearized' dynamics (i.e., considering the linear map from each subregion) within the three linear subregions of the AL-RNN trained on the ECG data from Fig. \ref{fig:ECG_minimal}. The first two subregions host weakly unstable spirals with shifted phase, corresponding to the excitatory/ inhibitory phases of the ECG. The strongly divergent activity in the third subregion induces the Q wave.}
	\label{fig:ecg_linear_activities}
\end{figure*}

\begin{figure*}[!htb]
    \centering
    \begin{subfigure}[b]{0.49\textwidth}
         \includegraphics[width=\textwidth]{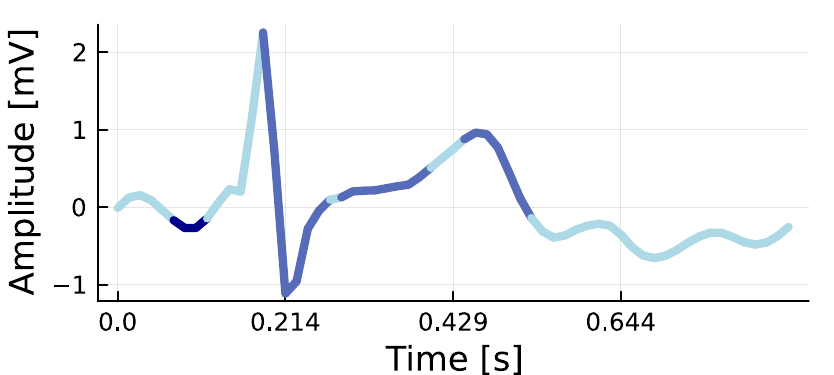}
    \end{subfigure}
    \hfill
    \begin{subfigure}[b]{0.49\textwidth}
         \includegraphics[width=\textwidth]{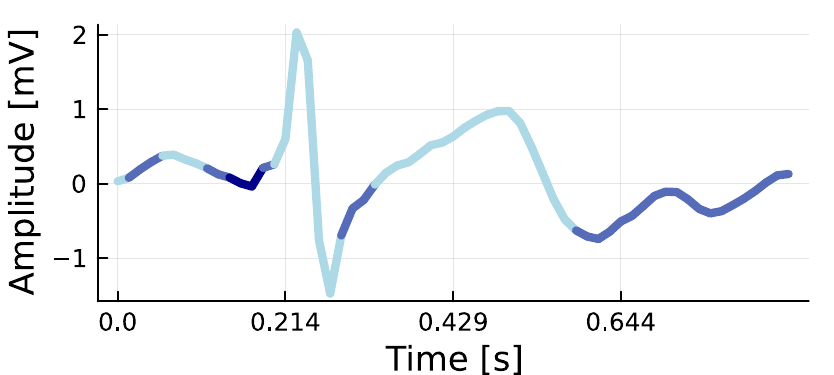}
    \end{subfigure}
    \begin{subfigure}[b]{0.49\textwidth}
         \includegraphics[width=\textwidth]{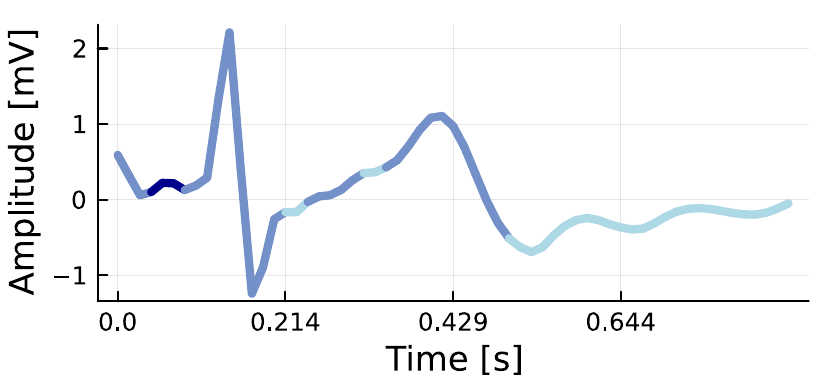}
    \end{subfigure}
    \hfill
    \begin{subfigure}[b]{0.49\textwidth}
         \includegraphics[width=\textwidth]{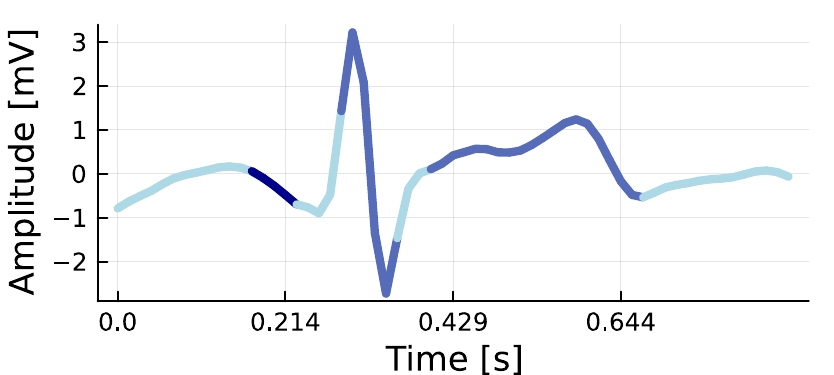}
    \end{subfigure}
    \caption{Freely generated ECG activity using an AL-RNN with 3 linear subregions (color-coded) shows consistent assignment of the Q wave to a distinct subregion across multiple successful reconstructions.}
    \label{fig:ECG_minimal_further}
\end{figure*}

\begin{figure*}[!htb]
    \centering
	\includegraphics[width=0.99\linewidth]{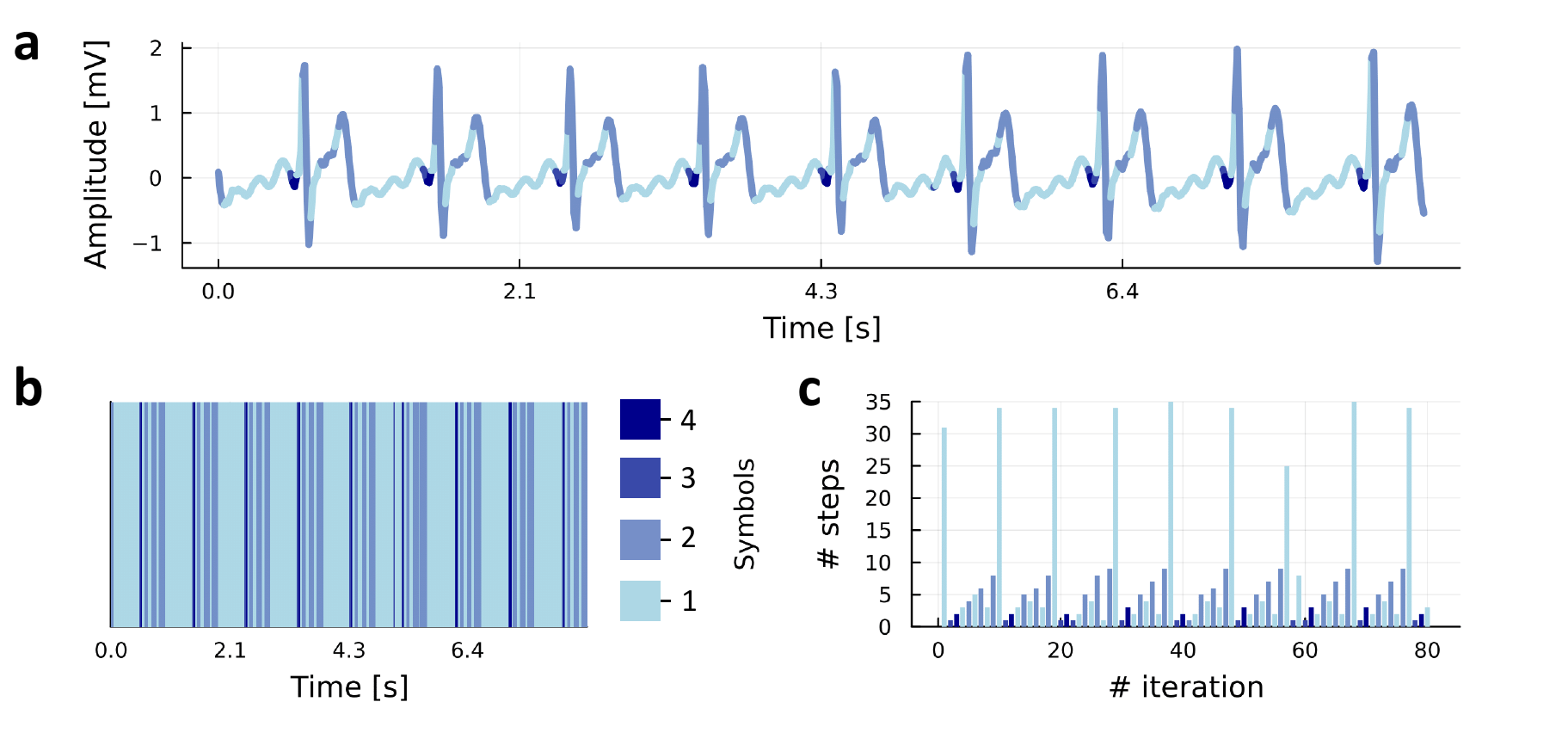}
	\caption{\textbf{a}: Freely generated ECG activity by the AL-RNN with $M=100$ total units and $P=2$ PWL units. \textbf{b}: Symbolic coding of the dynamics (with each shade of blue a different symbol/ linear subregion), reflecting the QRS phase with alternating excitation/inhibition (lighter shades of blue) following the short Q wave burst (dark blue). \textbf{c}: Time histogram of distinct symbols along the symbolic trajectory, exposing the mildly chaotic nature of the reconstructed ECG signal \cite{hess_generalized_2023}.}
	\label{fig:ECG_symbolic}
\end{figure*}

\begin{figure}[!htb]
    \centering
	\includegraphics[width=0.75\linewidth]{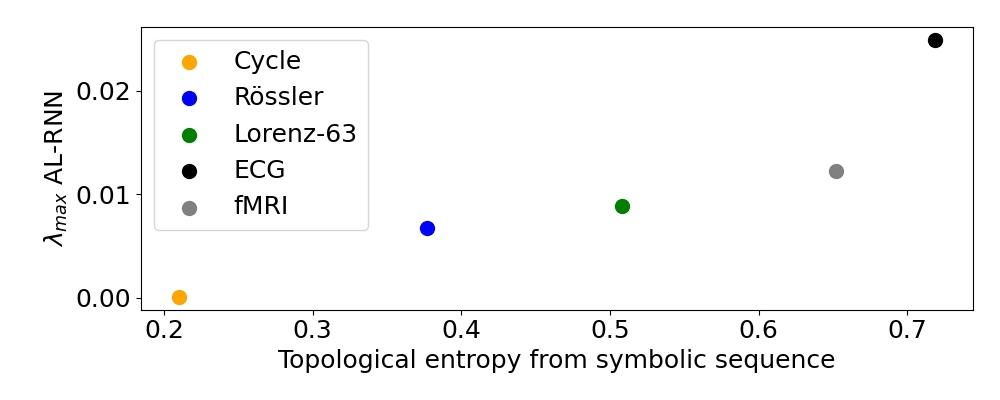}
	\caption{Topological entropy %approximated
 computed from symbolic sequences (Figs. \ref{fig:ECG_symbolic} $\&$ \ref{fig:fMRI_freely_generated}) versus $\lambda_{max}$, calculated from corresponding topologically minimal AL-RNNs (Figs. \ref{fig:minimal_reconstructions} $\&$ \ref{fig:ECG_minimal}).}
	\label{fig:entropy_lyaps}
\end{figure}

\begin{figure*}[!htb]
    \centering
	\includegraphics[width=0.8\linewidth]{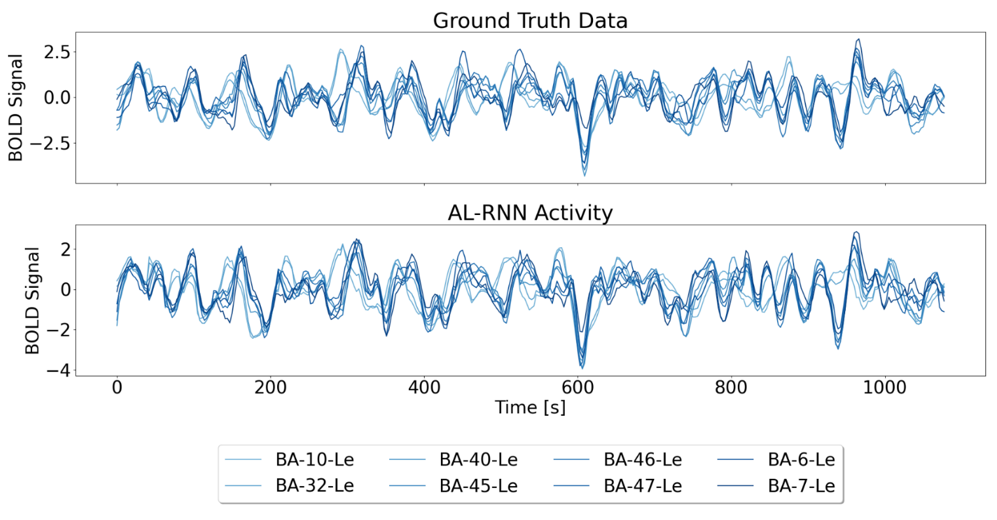}
	\caption{Generated fMRI activity using an AL-RNN with $M=100$ total units and $P=2$ PWL units, with the readout unit states replaced by observations every $7$ time steps.}%\textbf{a}: Freely generated activity of the two most frequented subregions for the model in Fig. \ref{fig:fMRI_freely_generated} ($M=100$, $P=3$). \textbf{b}:}
	\label{fig:fMRI_generated_linear_tf}
\end{figure*}

\begin{figure*}[!htb]
    \centering
	\includegraphics[width=0.99\linewidth]{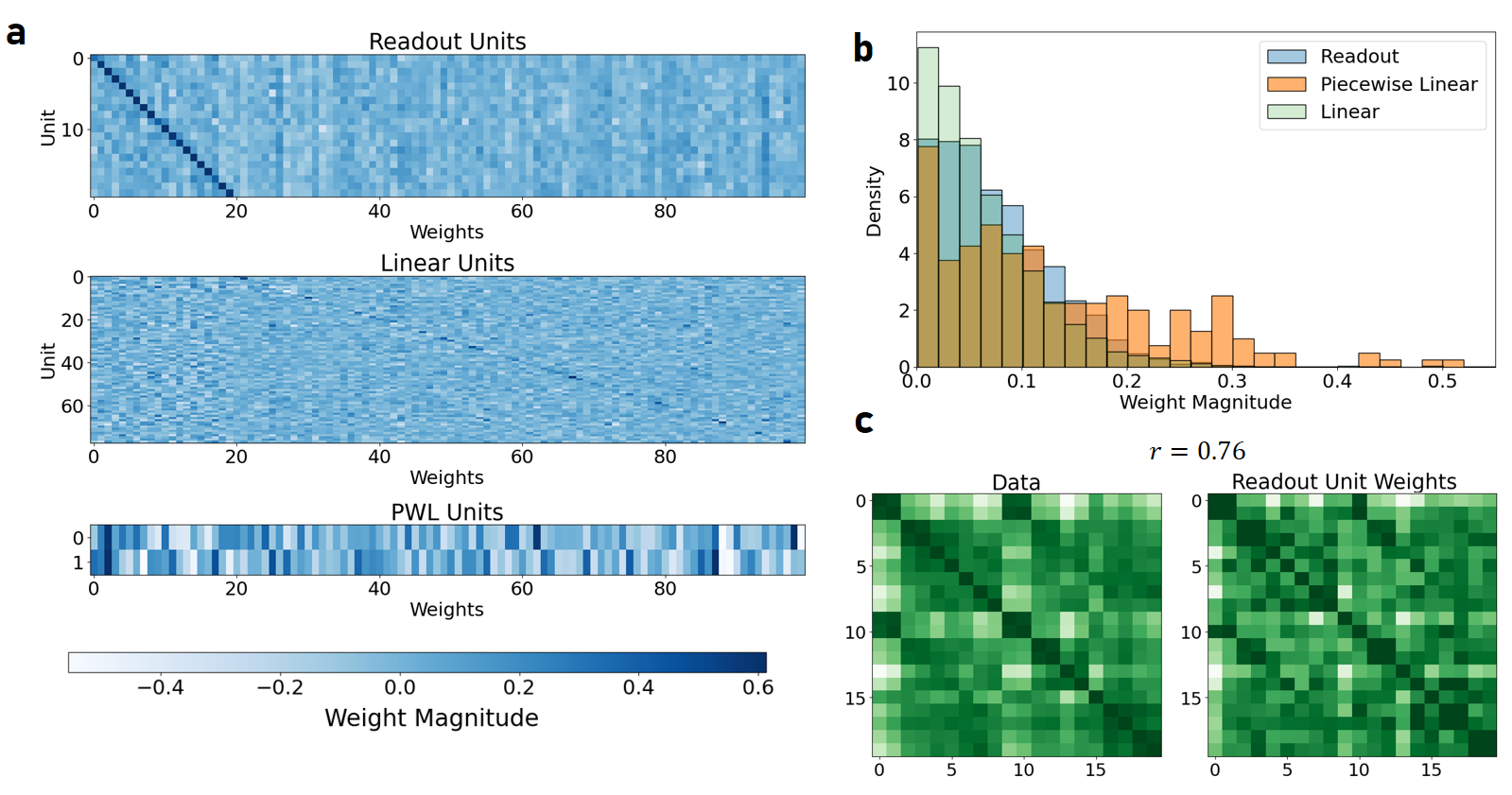}
	\caption{\textbf{a}: Weights of the reconstructed AL-RNN. \textbf{b}: Histogram of the absolute weight distributions for the different types of AL-RNN units. On average, weight magnitudes of the PWL units are much higher than those of the other unit types. \textbf{c}: The correlation structure among the weights of the $N=20$ readout units (rows in \textbf{a}, top) reflects the correlation structure within the observed time series variables (correlation between both matrices $r \approx 0.76$).}
	\label{fig:fMRI_weights}
\end{figure*}

\begin{figure*}[!htb]
    \centering
	\includegraphics[width=0.99\linewidth]{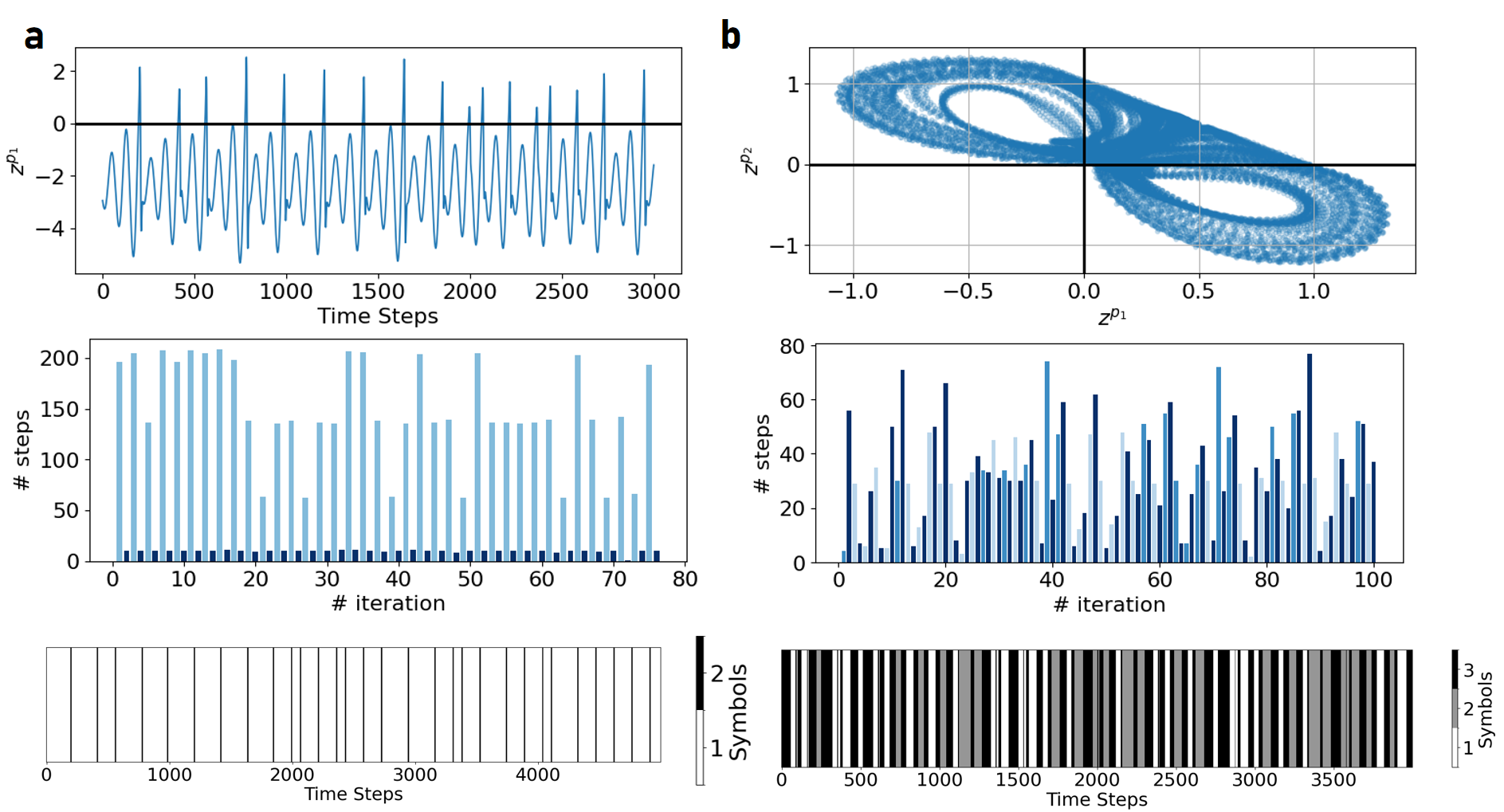}
	\caption{Top row: Activity of the PWL units for the topologically minimal representations of the Rössler (\textbf{a}) and Lorenz-63 attractor (\textbf{b}) from Fig. \ref{fig:minimal_reconstructions}. Center row: Time histogram of discrete symbols of the symbolic trajectory. Bottom row: Time series of the symbolic trajectory.}
	\label{fig:symbolic_roessler_lorenz}
\end{figure*}

%\clearpage

\begin{figure*}[!htb]
    \centering
	\includegraphics[width=0.99\linewidth]{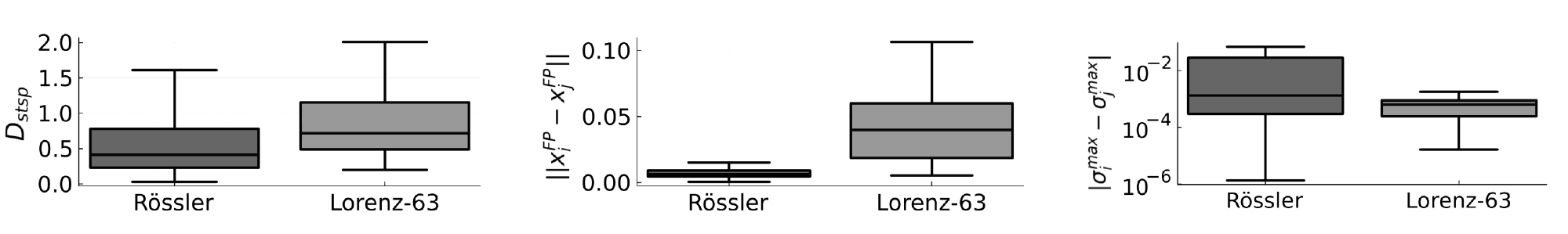}
	\caption{Same as Fig. \ref{fig:minimal_reconstructions}\textbf{d-f}, but showing absolute instead of relative deviations (for $D_{stsp}$, values $<1.0$ indicate tight agreement).}
 \label{fig:absolute_robustness}
\end{figure*}

\begin{figure}[!htb]
    \centering
    \scalebox{0.99}{
        \begin{minipage}[b]{0.33\columnwidth}
            \centering
            \includegraphics[width=\textwidth]{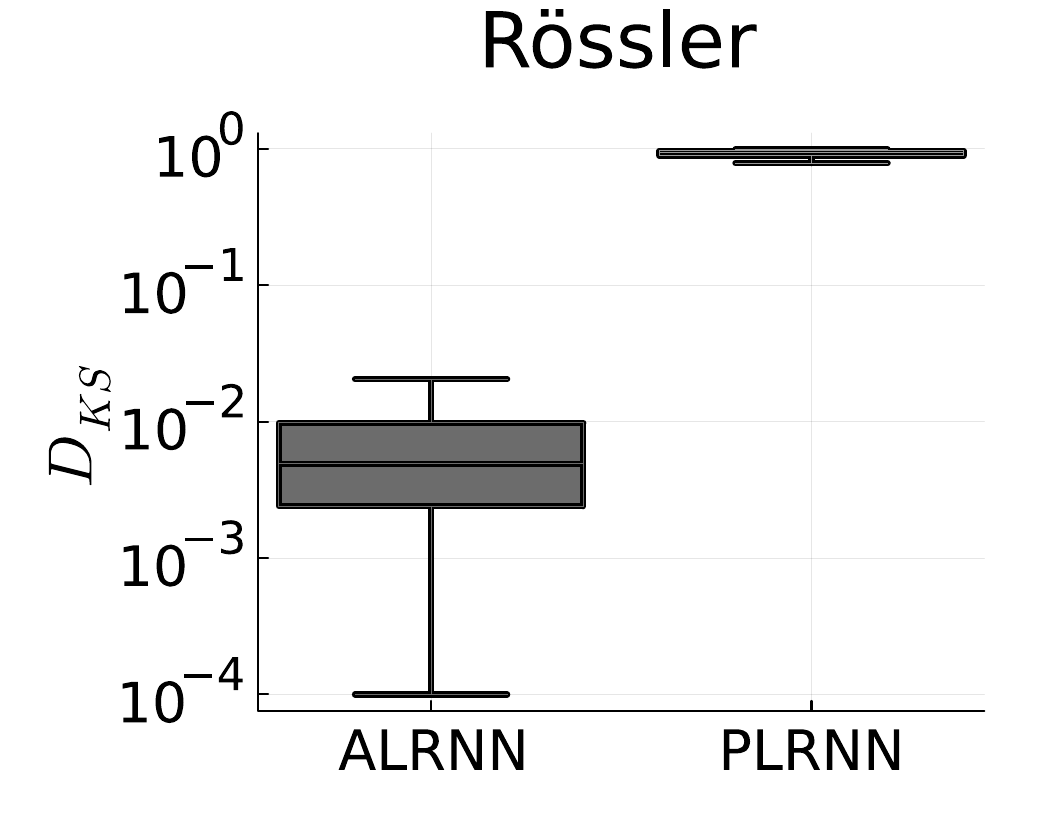}
        \end{minipage}
        \hfill
        \begin{minipage}[b]{0.33\columnwidth}
            \centering
            \includegraphics[width=\textwidth]{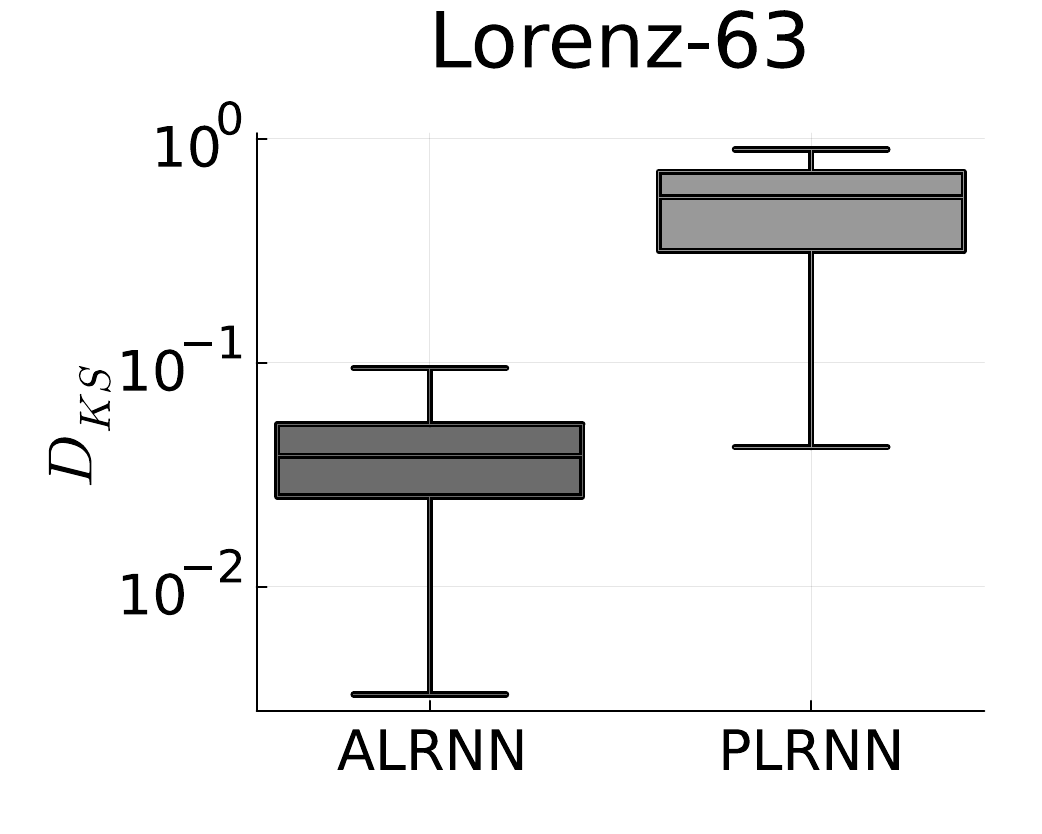}
        \end{minipage}
        \hfill
        \begin{minipage}[b]{0.33\columnwidth}
            \centering
            \includegraphics[width=\textwidth]{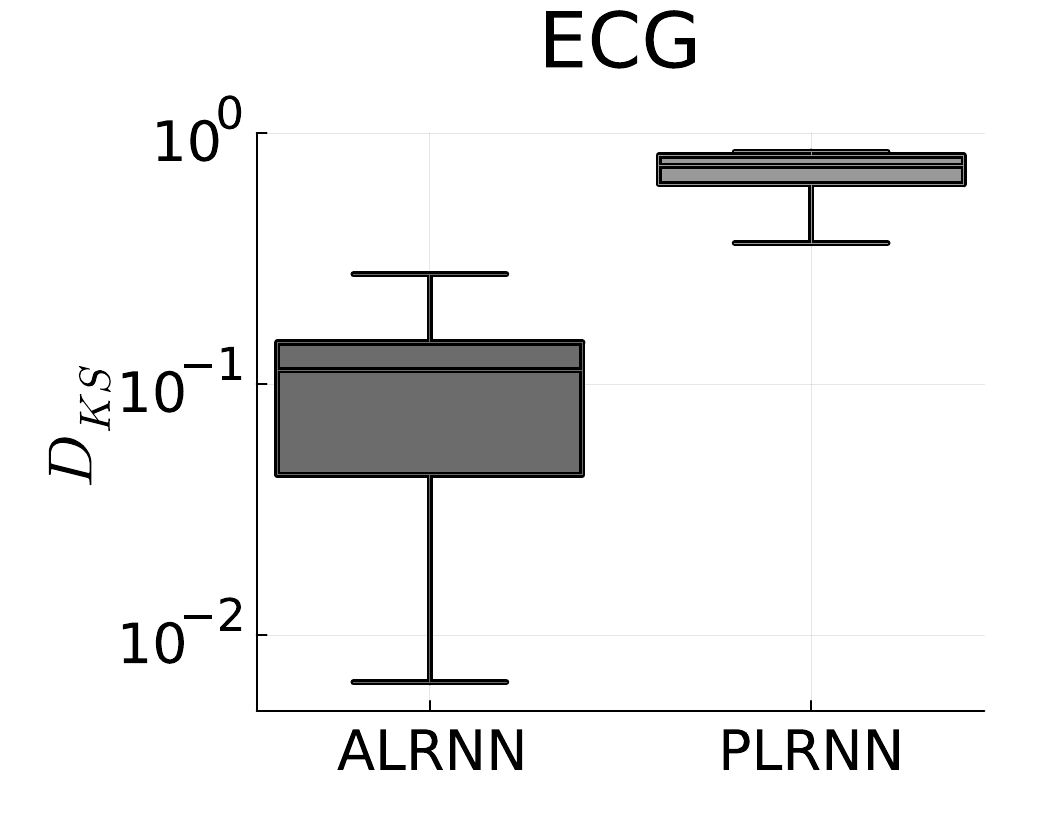}
        \end{minipage}
    }
    \caption{Pairwise differences in sorted (from lowest to highest probability) cumulative trajectory point distributions across linear subregions across all valid pairs from 20 training runs (quantified by the Kolmogorov–Smirnov distance, $D_{KS}$) for the AL-RNN vs. PLRNN, revealing much higher consistency for AL-RNN. Note the log-scale on the y-axis.}
    \label{fig:chi2_comparison}
\end{figure}

\begin{figure*}[!htb]
    \centering
	\includegraphics[width=0.8\linewidth]{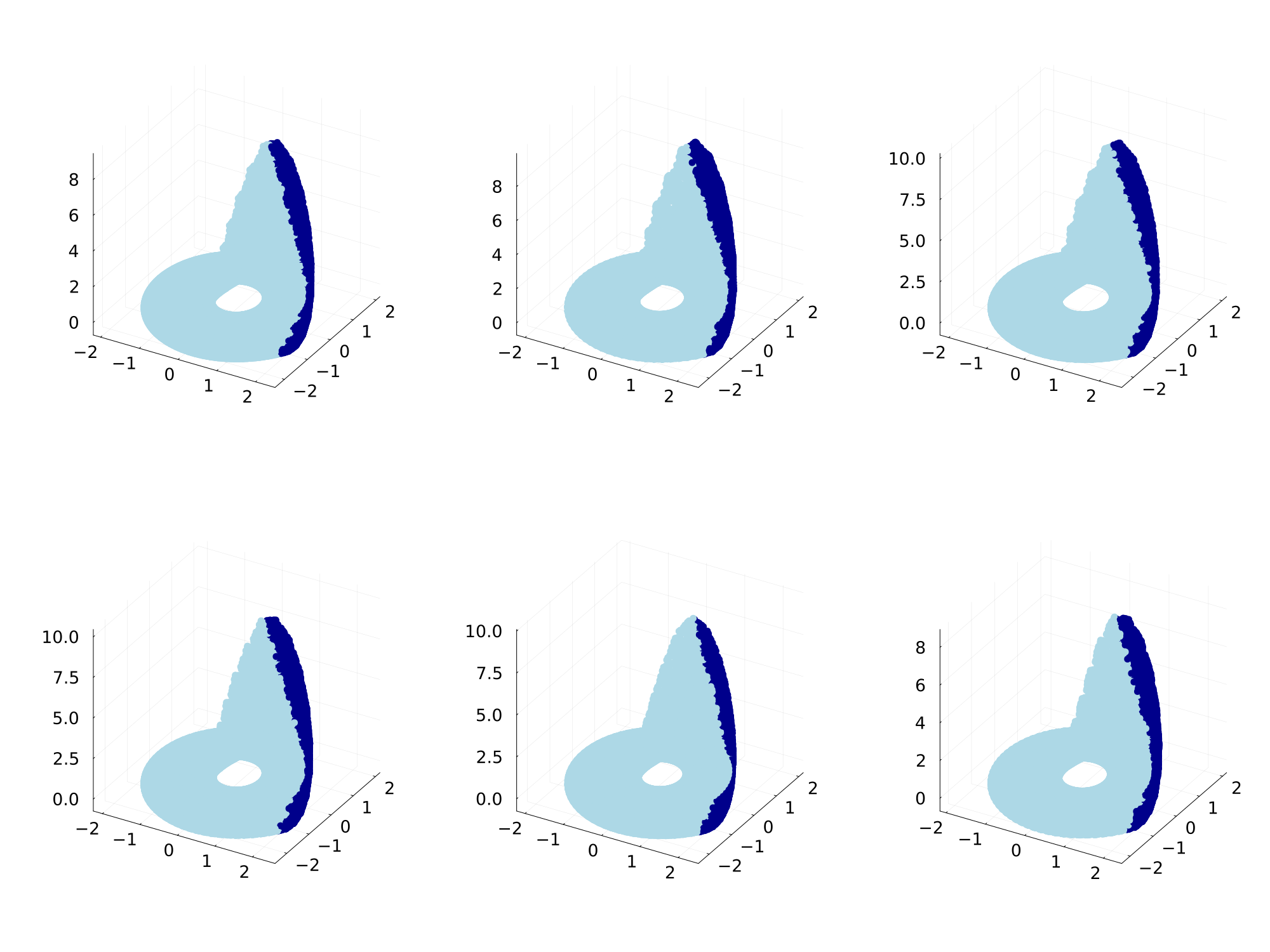}
	\caption{Linear subregions (color-coded) mapped to observation space of the Rössler system, showing a robust representation of the individual subregions across multiple training runs/ models.}
	\label{fig:minimal_subregions_Rossler}
\end{figure*}

\begin{figure*}[!htb]
    \centering
	\includegraphics[width=0.8\linewidth]{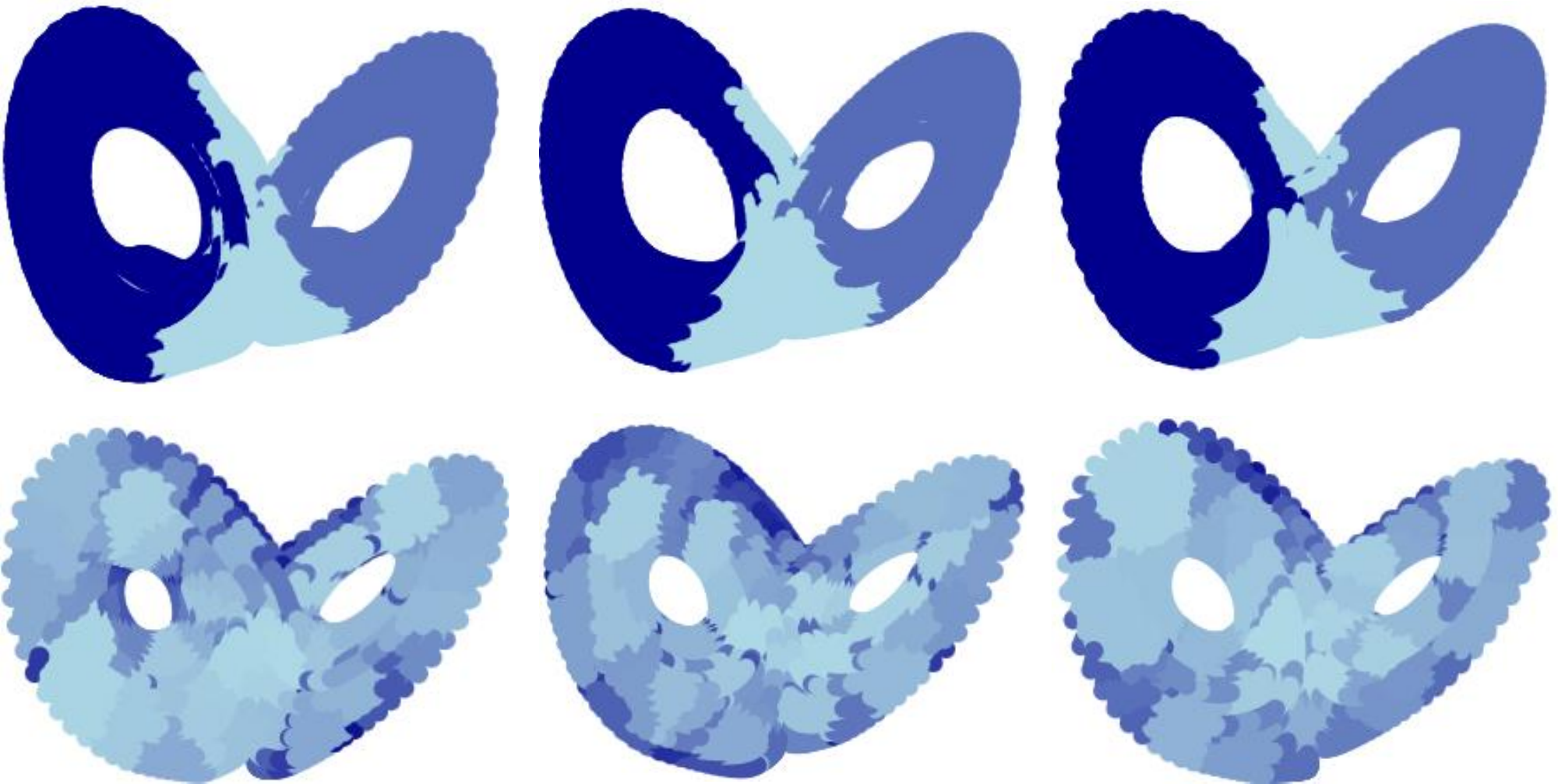}
	\caption{Top row: Robust placing of linear subregions (color coded) mapped to observation space across %20 
 training runs using the AL-RNN. Model recovery experiments further confirmed the robustness of the model solutions, with very similar overall performance measures across different experiments (original: $D_{stsp}=3.14$, $D_H=0.28$; recovered: $D_{stsp}=3.38\pm0.18$, $D_H=0.28\pm0.03$; 3 linear subregions in all cases). Bottom row: In contrast, for the PLRNN linear subregions are differently assigned (with varying boundaries) on each run.}
	\label{fig:minimal_subregions_Lorenz63}
\end{figure*}

\clearpage

\clearpage
\section{Proofs of Theorems} \label{proofs}

Define a shift space of finite type $(A_{\mathcal{F}},\sigma)$ such that each symbol of the alphabet of $\mathcal{A}$ is associated with exactly one set $U_e$ of the topological partition of $S$, i.e. we have a total of $|\mathcal{U}|$ symbols in $\mathcal{A}$. Further define for $\bm{a} \in A_{\mathcal{F}}$ the sets \cite{lind_introduction_1995}
\begin{align}
   D_l(\bm{a})= \bigcap_{k=-l}^{l} \phi^{-k}(U_{a_{k}}) \subseteq S,
\end{align}
where $a_k$ is the $k$th symbol in the sequence $\bm{a}$: Think of this as building the intersection of $U_{a_0}$ with $k$-times forward iterates of subsets associated with $a_{-k}$ and $k$-times backward iterates of subsets associated with $a_{k}$, such that in the limit $l \rightarrow \infty$ hopefully we end up with a single point corresponding to a unique trajectory in $S$, i.e. considering $D(\bm{a})= \bigcap_{l=0}^{\infty} \overline{D_l}(\bm{a})$ (see \citet{lind_introduction_1995} for details). We now define \cite{lind_introduction_1995}
\begin{definition}\label{def:symbrepr}
       A \textit{symbolic representation} of an invertible DS $(S,\phi)$ with topological partition $\mathcal{U}=\{U_0 \dots U_{n-1}\}$ is a shift space $(A_{\mathcal{F}},\sigma)$ with alphabet $\mathcal{A}=\{0 \dots n-1\}$, such that each symbol $a_k \in \mathcal{A}$ is associated with exactly one subset $U_k \in \mathcal{U}$, and $D(\bm{a})= \bigcap_{l=0}^{\infty} \overline{D_l}(\bm{a})$ contains exactly one point $\bm{x} \in S$ for each $\bm{a} \in A_{\mathcal{F}}$. If $A_{\mathcal{F}}$ is a finite shift, we call this a \textit{Markov partition}.
\end{definition} 
Ideally, we would like our symbolic coding of the DS to be a symbolic representation or Markov partition according to this definition, but in practice this may entail other unfavorable properties (e.g., too fine-grained) and we contend here with the properties given by Theorems \ref{FPsymbPLRNN} - \ref{ChaossymbPLRNN}.

%The following definitions are adapted from Alligood's book:\\
In the statement of our theorems, we further used the terms `eventually periodic' and `asymptotically periodic'. Let us now strictly define them (according to \citet{alligood_chaos_1996}):

\begin{definition}
  An orbit $\{\bm{z}_1, \ldots, \bm{z}_n, \ldots\}$ of the map %$F_{\bm{\theta}}$
  $\phi$ is said to be \textbf{asymptotically periodic} if it converges to a periodic orbit as $n \to \infty$. This implies that there is a periodic orbit $\Gamma_k \, = \, \{\bm{y}_1, \bm{y}_2, \ldots, \bm{y}_k, \bm{y}_1, \bm{y}_2, \ldots \}$ such that $\displaystyle{\lim_{n \to \infty} d(\vz_{n}, \Gamma_k}) \, = \, 0$.    
\end{definition}
%--------------------
For instance, any orbit that is attracted to a stable fixed point or to a saddle fixed point (evolving on its stable manifold) is asymptotically periodic (fixed).
%------------------------
\begin{definition}
 A point $\bm{z}$ is called \textbf{eventually periodic} with period $p$ for the map %$F_{\bm{\theta}}$
 $\phi$, if for some positive integer $N$, %$F_{\bm{\theta}}^{n+p} (\bm{z}) \, = \, F_{\bm{\theta}}^{n} (\bm{z})$
 $\phi^{n+p} (\bm{z}) \, = \, \phi^{n} (\bm{z})$ for all $n \geq N$, and $p$ is the smallest positive integer with this exact property. This means that the orbit of $\bm{z}$ eventually maps \textbf{exactly
onto a periodic orbit}. 
\end{definition}
%---------------------------
\textbf{Note}: The term `eventually periodic' describes the extreme case where an orbit coincides \textit{precisely} with a periodic orbit. Thus, any eventually periodic orbit is also asymptotically periodic, but the reverse is not always true: An asymptotically periodic orbit comes arbitrarily close to a periodic orbit, but may not land precisely on it.\\

With these definitions we are now ready to prove the theorems.

\paragraph{Proof of Theorem \ref{FPsymbPLRNN}}
\begin{proof}
 ``$\Rightarrow$'': Suppose that the orbit $\Omega_S=\{\bm{z}_1, \ldots, \bm{z}_n, \ldots\}$ is asymptotically fixed. Hence, there exists a fixed point $\bm{z}^* \in U_{a^*}$ of the AL-RNN $F_{\bm{\theta}}$ (i.e., $\bm{z}^*=F_{\theta}(\bm{z}^*)$) such that $\displaystyle{\lim_{n \to \infty} \bm{z}_n } \, = \, \bm{z}^*$. Let $\bm{a}= (a_{1} a_{2} a_{3} \ldots) \,$ be the corresponding symbolic sequence of the orbit $\Omega_S$ with $\bm{z}_n \in U_{a_n}, \forall n$. Since $\displaystyle{\lim_{n \to \infty} \bm{z}_n } \, = \, \bm{z}^*$, so there exists some $\,N\in \mathbb{N}\,$ such that for every $\,n \geq N\,$ the points $\bm{z}_n$ will remain arbitrarily close to $\bm{z}^*$. This means the points $\bm{z}_n$ eventually enter the same linear subregion that contains $\bm{z}^*$ and will remain there for all future iterations. Thus, $ \bm{a} \, = \, (a_{1} a_{2} a_{3}\dots a_{N-1})(a^*)^{\infty} \,$ is eventually fixed.
\\[1ex]
``$\Leftarrow$'': Assume $\bm{a} \, = \, (a_{1} a_{2} a_{3}\dots a_{N-1})(a^*)^{\infty} \,$ is an eventually fixed sequence of the symbolic encoding. This means for all the corresponding orbits $\Omega_S=\{\bm{z}_1, \dots, \bm{z}_n, \dots\}$ of $F_{\bm{\theta}}$, there exists an index $ N $ such that for every $\,n \geq N\,$ the orbit points $\, \bm{z}_n \,$ must remain in the same subregion, say $U_{a^*}$. Since by assumption the map $F_{\theta}$ is non-globally-diverging 
and hyperbolic, the system cannot be expanding in all directions in any of the subregions. Thus, there must be at least one contracting or stable direction in $U_{a^*}$. Consequently, there must be at least one corresponding orbit $\{\bm{z}_1, \dots, \bm{z}_n, \dots\}$ that converges toward some stable structure in $U_{a^*}$ as $n \to \infty$. Since $F_{\theta}$ is a linear hyperbolic map in each subregion, there cannot be any $k$-cycle, $k\geq2$, with all periodic points contained within a single subregion. Therefore, the corresponding orbit converges to a (saddle) fixed point $\bm{z}^* \in U_{a^*}$ in the stable manifold along stable directions. If \textbf{all} directions in $U_{a^*}$ are contracting or stable, then all corresponding orbits will converge to a stable fixed point within that subregion.
\end{proof}
%------------------------

\paragraph{Proof of Theorem \ref{CyclesymbPLRNN}}
%------------------
\begin{proof}
``$\Rightarrow$'': Let $\Omega_S=\{\bm{z}_1, \ldots, \bm{z}_n, \ldots\}$ be an asymptotically $p$-periodic orbit of $F_{\theta}$. Then there is a periodic
orbit $\Omega_{S,p}=\{\bm{z}_1^*, \dots, \bm{z}_p^*\}$ (i.e., such that all points $\bm{z}_k^* \in \Omega_{S,p}$ are distinct and $\bm{z}_{k+p}^*=F^p_{\bm{\theta}}(\bm{z}_k^*)=\bm{z}_{k}^*, k=1 \dots p$) and
%-----------------
\begin{align}\label{eq-limit}
 \displaystyle{\lim_{n \to \infty} d(\vz_{n}, \Omega_{S,p}}) \, = \, 0.  
\end{align}
%------------------------
Since $F_{\bm{\theta}}$ is a linear and hyperbolic map in each subregion, there cannot be any $p$-periodic orbit with $p \geq 2$ where all periodic points are contained within a single subregion. On the other hand, due to the definition of the symbolic coding for each trajectory, each point $\vz_t$ at time step $t$ is assigned its own symbol $a_t$, depending on its associated linear subregion. Hence, the corresponding symbolic sequence of the periodic orbit $\Omega_{S,p}$ is $(a_{1}^* a_{2}^* \ldots a_{p}^*)^{\infty}$ with $\bm{z}_k \in U_{a^*_k}, k=1 \dots p\,$. Moreover, due to eq. \ref{eq-limit}, for some large enough index $N \in \mathbb{N}$,
 the points $\bm{z}_n$ of the orbit $\Omega_S$ become arbitrarily close to the orbit $\Omega_{S,p}$. Thus, for $n \geq N$, the itinerary of $\Omega_S$ will follow the same repeating pattern as the periodic orbit’s itinerary and will revisit the same subregions as the periodic orbit points $\bm{z}_{k}^*$, $k=1 \dots p$. Therefore, the corresponding symbolic sequence of the orbit $\Omega_S$ is $\bm{a} \, = 
 \, (a_{1} a_{2} \ldots a_{N-1})(a^*_1 a^*_2 \ldots a^*_{p})^{\infty}$, which is an eventually $p$-periodic orbit of the shift map $\sigma$.
\\[1ex]
``$\Leftarrow$'': Assume that $\bm{a} \, = \, (a_{1} a_{2} \ldots a_{N-1})(a^*_1 a^*_2 \ldots a^*_{p})^{\infty} \in A_{\mathcal{U},F_{\bm{\theta}}}$ is an eventually $p$-periodic orbit of the shift map $\sigma$. Consequently, 
%-----------------
\begin{align}
\exists N \in \mathbb{N} \, \, \, \, \forall \, \, \, n \geq N\, : \hspace{.5cm} a_{n+p}= a_n,
\end{align}
%-------------------
which says that the symbolic sequence is repeating every $p$ steps. Therefore, for all the corresponding orbits $\Omega_S=\{\bm{z}_1, \dots, \bm{z}_n, \dots\}$ of $F_{\bm{\theta}}$, there exists an index $ N $ such that for every $\,n \geq N\,$ the orbit points $\, \bm{z}_n \,$ will stay in the same $\, p \,$ linear subregions, say $U_{a_1^*}, U_{a_2^*} \ldots , U_{a_{p}^*}$, and revisit each $U_{a_i^*} (i = 1, 2, \ldots, p)$ after exactly $p$ time steps: 
%-------------------
\begin{align}\label{sub_seq}
\exists N \in \mathbb{N} \, \, \, \, \forall \, \, \, n \geq N\, : \hspace{.5cm} U_{a_{n+p}^*} = U_{a_{n}^*}.    
\end{align}
%---------------------
Since by assumption the map $F_{\theta}$ is non-globally-diverging 
and hyperbolic, the system cannot be expanding in all directions in any of the subregions. Thus, there exists at least one contracting direction in each of the subregions $U_{a_1^*}, U_{a_2^*} \ldots , U_{a_{p}^*}$, and therefore at least one corresponding orbit $ \Omega_S = \{\bm{z}_1, \dots, \bm{z}_n, \dots\}$ that converges toward some stable structure within $U_{a_1^*}, U_{a_2^*} \ldots , U_{a_{p}^*}$ as $n \to \infty$. Furthermore, as $F_{\theta}$ is a linear and hyperbolic map in each subregion there cannot be any $k$-cycle, $k\geq2$, with all periodic points contained within only one of the subregions. Similarly, it cannot be chaotic or quasi-periodic within
just one subregion. Now, due to eq. \ref{sub_seq}, for the corresponding orbit $ \Omega_S$ 
%-------------------
\begin{align}\label{}
\forall \, n \geq N \, \, \, \forall \, \vz_n \in U_{a^*_i} (i=1 \dots p) \, \, \, :  \, \vz_{n+p} = F^{p}_{\bm{\theta}} (\vz_n) = \mW_{\va} \vz_n + \vh_{\va} \in U_{a^*_i},
\end{align}
%---------------------
with fixed parameters $\bm{W}_{\bm{a}}:=\bm{W}_{\Omega^{a^*_p}} \cdots \bm{W}_{\Omega^{a^*_2}} \bm{W}_{\Omega^{a^*_1}}$ and $\bm{h}_{\bm{a}}:=\sum_{i=2}^{p}{\prod_{j=2}^{i} \bm{W}_{\Omega^{a_{p-j+2}}}} \bm{h} + \bm{h}$. Since $F^{p}_{\bm{\theta}}$ is strictly affine, for $n \geq N $, any sub-sequence $\{ \vz_{n+ mp} \}^{\infty}_{m=1}$ of $\Omega_S$ cannot be convergent to an aperiodic (i.e., chaotic or quasi-periodic) orbit. Therefore, the corresponding orbit $\Omega_S$ converges to a (saddle) $p$-periodic orbit, with all periodic points within the sub-regions $U_{a_1^*}, U_{a_2^*} \ldots , U_{a_{\bar{p}}^*}$, in the stable manifold along stable directions. If all directions in $U_{a^*}$ are contracting or stable, then all corresponding orbits will converge to a stable $p$-periodic orbit with all periodic points within the subregions $U_{a_1^*}, U_{a_2^*} \ldots , U_{a_{\bar{p}}^*}$.
\end{proof}
%---------------------- 
%-----------------
\paragraph{Proof of Theorem \ref{ChaossymbPLRNN}}
\begin{proof}
``$\Rightarrow$'': Let $\Omega_S=\{\bm{z}_1, \bm{z}_1, \dots, \bm{z}_n, \dots \}$ of $F_{\bm{\theta}}$ be an asymptotically \textbf{aperiodic} orbit of $F_{\theta}$. Then, there is an aperiodic
orbit $\Omega=\{\bar{\vz}_1, \bar{\vz}_2 \dots\}$ (i.e., with $\bar{\vz}_k \neq F^p_{\bm{\theta}}(\bar{\vz}_k) \forall k, p > 0$) and
%-----------------
\begin{align}\label{eq-ap}
 \displaystyle{\lim_{n \to \infty} d(\vz_{n}, \Omega}) \, = \, 0.  
\end{align}
%------------------------
According to the proof of Theorem \ref{CyclesymbPLRNN} (second part), this orbit cannot have an eventually periodic symbolic representation $(a_{1} a_{2} \ldots a_{N-1})(a_1 a_2 \dots a_p)^{\infty}$, because if it had, it would need to be asymptotically periodic as well.
\\[1ex]
``$\Leftarrow$'': Assume an aperiodic symbolic sequence $\bm{a}=(a_1, \ldots, a_n, \ldots)$, where there is no $p>0$ such that $a_k=\sigma^p(a_k) \forall k$. This will correspond to an aperiodic succession $ U_{a_{1}} \ldots U_{a_{n}}\ldots$ of linear subregions $U_{a_k}$ visited, since each subregion has its unique symbol. However, from the proof of Theorem \ref{CyclesymbPLRNN} (first part) we know that any asymptotically periodic orbit of $F_{\bm{\theta}}$ must have an eventually periodic symbolic encoding, so the orbit $\Omega_S$ corresponding to $\bm{a}$ cannot be asymptotically periodic. Consequently, it cannot be (eventually) periodic either, which implies that it must be aperiodic.
\end{proof}
%--------------------------

\newpage
\section*{NeurIPS Paper Checklist}

\begin{enumerate}

\item {\bf Claims}
    \item[] Question: Do the main claims made in the abstract and introduction accurately reflect the paper's contributions and scope?
    \item[] Answer:  \answerYes{} % Replace by \answerYes{}, \answerNo{}, or \answerNA{}.
    \item[] Justification: The results illustrated in the text and figures properly address the presented claims, including formal proof for the theorems and hyperparameter settings and code for reproducing the results of the paper. %\justificationTODO{}

\item {\bf Limitations}
    \item[] Question: Does the paper discuss the limitations of the work performed by the authors?
    \item[] Answer:  \answerYes{} % Replace by \answerYes{}, \answerNo{}, or \answerNA{}.
    \item[] Justification: Limitations are addressed in the conclusion or, e.g. in the case of theoretical results, directly in the respective section.%\justificationTODO{}

\item {\bf Theory Assumptions and Proofs}
    \item[] Question: For each theoretical result, does the paper provide the full set of assumptions and a complete (and correct) proof?
    \item[] Answer:  \answerYes{} % Replace by \answerYes{}, \answerNo{}, or \answerNA{}.
    \item[] Justification: The paper includes formal proofs of all theorems Appx. \ref{proofs}. %\justificationTODO{}

    \item {\bf Experimental Result Reproducibility}
    \item[] Question: Does the paper fully disclose all the information needed to reproduce the main experimental results of the paper to the extent that it affects the main claims and/or conclusions of the paper (regardless of whether the code and data are provided or not)?
    \item[] Answer:  \answerYes{} % Replace by \answerYes{}, \answerNo{}, or \answerNA{}.
    \item[] Justification: Detailed hyperparameter settings for the results in the paper are given in Sect. \ref{appx:method_details}, which combined with the codebase allow for replication of the results. %\justificationTODO{}

\item {\bf Open access to data and code}
    \item[] Question: Does the paper provide open access to the data and code, with sufficient instructions to faithfully reproduce the main experimental results, as described in supplemental material?
    \item[] Answer: \answerYes{} % Replace by \answerYes{}, \answerNo{}, or \answerNA{}.
    \item[] Justification: A link to a github repository is provided that contains the implementation of the network architecture that was used for the main results in the paper (\url{https://github.com/DurstewitzLab/ALRNN-DSR}).%\justificationTODO{}

\item {\bf Experimental Setting/Details}
    \item[] Question: Does the paper specify all the training and test details (e.g., data splits, hyperparameters, how they were chosen, type of optimizer, etc.) necessary to understand the results?
    \item[] Answer: \answerYes{} % Replace by \answerYes{}, \answerNo{}, or \answerNA{}.
    \item[] Justification: All hyperparameter settings are provided in Sect. \ref{appx:method_details}.%\justificationTODO{}

\item {\bf Experiment Statistical Significance}
    \item[] Question: Does the paper report error bars suitably and correctly defined or other appropriate information about the statistical significance of the experiments?
    \item[] Answer: \answerYes{} % Replace by \answerYes{}, \answerNo{}, or \answerNA{}.
    \item[] Justification: Error bars are reported in figures and with numerical results, i.e. Figs. \ref{fig:performance_relu}, \ref{fig:subregions_covered} and \ref{fig:minimal_reconstructions}. The settings for the random initialization of the network architecture are specified in Sect. \ref{appx:method_details}. %\justificationTODO{}

\item {\bf Experiments Compute Resources}
    \item[] Question: For each experiment, does the paper provide sufficient information on the computer resources (type of compute workers, memory, time of execution) needed to reproduce the experiments?
    \item[] Answer: \answerYes{} % Replace by \answerYes{}, \answerNo{}, or \answerNA{}.
    \item[] Justification: Details on computational resources are provided in Sect. \ref{appx:method_details}.%\justificationTODO{}
    
\item {\bf Code Of Ethics}
    \item[] Question: Does the research conducted in the paper conform, in every respect, with the NeurIPS Code of Ethics \url{https://neurips.cc/public/EthicsGuidelines}?
    \item[] Answer: \answerYes{} % Replace by \answerYes{}, \answerNo{}, or \answerNA{}.
    \item[] Justification: All datasets considered here were publically available. The research is aimed at facilitating interpretability of machine learning models. We believe our approach has primarily positive ethical implications. Although we have not identified specific ethical concerns, the wide range of potential applications means that possible misuses cannot be entirely ruled out.% \justificationTODO{}

\item {\bf Broader Impacts}
    \item[] Question: Does the paper discuss both potential positive societal impacts and negative societal impacts of the work performed?
    \item[] Answer: \answerNA{} % Replace by \answerYes{}, \answerNo{}, or \answerNA{}.
    \item[] Justification: This work constitutes foundational research. As such, we do not foresee any direct negative societal impact. %\justificationTODO{}
    
\item {\bf Safeguards}
    \item[] Question: Does the paper describe safeguards that have been put in place for responsible release of data or models that have a high risk for misuse (e.g., pretrained language models, image generators, or scraped datasets)?
    \item[] Answer: \answerNA{} % Replace by \answerYes{}, \answerNo{}, or \answerNA{}.
    \item[] Justification: We do not foresee a high risk for misuse. %\justificationTODO{}

\item {\bf Licenses for existing assets}
    \item[] Question: Are the creators or original owners of assets (e.g., code, data, models), used in the paper, properly credited and are the license and terms of use explicitly mentioned and properly respected?
    \item[] Answer: \answerYes{} % Replace by \answerYes{}, \answerNo{}, or \answerNA{}.
    \item[] Justification: All code and results were created by us. Datasets were publically sourced and the respective authors cited. %\justificationTODO{}

\item {\bf New Assets}
    \item[] Question: Are new assets introduced in the paper well documented and is the documentation provided alongside the assets?
    \item[] Answer: \answerYes{} % Replace by \answerYes{}, \answerNo{}, or \answerNA{}.
    \item[] Justification: A fully documented version of the code is published on github.%A preliminary version of the code is provided with the submission, and a fully documented version will be published on github with acceptance. %\justificationNA{}

\item {\bf Crowdsourcing and Research with Human Subjects}
    \item[] Question: For crowdsourcing experiments and research with human subjects, does the paper include the full text of instructions given to participants and screenshots, if applicable, as well as details about compensation (if any)? 
    \item[] Answer: \answerNA{} % Replace by \answerYes{}, \answerNo{}, or \answerNA{}.
    \item[] Justification: The paper did not involved crowdsourcing or research with human subjects. %\justificationTODO{}

\item {\bf Institutional Review Board (IRB) Approvals or Equivalent for Research with Human Subjects}
    \item[] Question: Does the paper describe potential risks incurred by study participants, whether such risks were disclosed to the subjects, and whether Institutional Review Board (IRB) approvals (or an equivalent approval/review based on the requirements of your country or institution) were obtained?
    \item[] Answer: \answerNA{} % Replace by \answerYes{}, \answerNo{}, or \answerNA{}.
    \item[] Justification: The paper did not involved crowdsourcing or research with human subjects. %\justificationTODO{}

\end{enumerate}

\end{document}